\documentclass[5p]{elsarticle}

\usepackage{lineno,hyperref}

\usepackage{soul}
\soulregister\cite7 
\soulregister\citep7 
\soulregister\citet7 
\soulregister\ref7 
\soulregister\pageref7 
\usepackage{mathrsfs}
\usepackage[noend]{algpseudocode}

\usepackage{algorithmicx,algorithm}
\usepackage{bbm}
\usepackage{graphicx}
\usepackage{amsmath}
\usepackage{amssymb}
\usepackage{multirow}
\usepackage{helvet}
\usepackage{courier}
\usepackage{float}
\usepackage{graphicx}

\usepackage{amsfonts}
\usepackage{textcomp}
\usepackage{hyperref}

\usepackage{subfigure}
\usepackage{times}
\usepackage{epsfig}
\usepackage[usenames,dvipsnames]{color}

\usepackage{amsmath,amsfonts}

\usepackage{array}
\usepackage{textcomp}
\usepackage{stfloats}
\usepackage{url}
\usepackage{verbatim}
\usepackage{arydshln}
\usepackage{bbding}
\usepackage{wrapfig}
\usepackage{caption}
\usepackage{booktabs}
\usepackage{colortbl}
\usepackage{newfloat}
\usepackage{listings}
\usepackage{bm}
\usepackage{pifont}

\usepackage{tabularx}
\usepackage{booktabs}  
\usepackage{graphicx}  
\usepackage{adjustbox} 

\definecolor{ceiling}{RGB}{210,42,33}
\definecolor{floor}{RGB}{40,159,7}
\definecolor{wall}{RGB}{162,215,224}
\definecolor{windows}{RGB}{113,158,201}
\definecolor{chair}{RGB}{202,205,76}
\definecolor{bed}{RGB}{224,190,159}
\definecolor{sofa}{RGB}{153,98,192}
\definecolor{table}{RGB}{23,124,181}
\definecolor{tvs}{RGB}{160,187,34}
\definecolor{furniture}{RGB}{225,126,18}
\definecolor{objects}{RGB}{199,174,221}

\definecolor{road}{RGB}{255,1,252}
\definecolor{sidewalk}{RGB}{76,0,74}
\definecolor{parking}{RGB}{255,149,255}
\definecolor{other_grnd}{RGB}{178,4,75}
\definecolor{building}{RGB}{255,198,0}
\definecolor{car}{RGB}{98,152,240}
\definecolor{truck}{RGB}{82,28,183}
\definecolor{bicycle}{RGB}{101,229,248}
\definecolor{motorcycle}{RGB}{35,57,148}
\definecolor{other_veh}{RGB}{100,82,245}
\definecolor{vegetation}{RGB}{2,176,0}
\definecolor{trunk}{RGB}{131,62,6}
\definecolor{terrain}{RGB}{153,238,85}
\definecolor{person}{RGB}{255,26,29}
\definecolor{bicyclist}{RGB}{255,40,202}
\definecolor{motorcycl}{RGB}{150,29,99}
\definecolor{fence}{RGB}{148,125,42}
\definecolor{pole}{RGB}{255,239,143}
\definecolor{traf_sign}{RGB}{248,2,3}

\makeatletter
\renewcommand{\@makefntext}[1]{#1}
\makeatother

\journal{Information Fusion}

\begin{document}

\begin{sloppypar}
\begin{frontmatter}

\title{Multimodal Fusion and Vision-Language Models: A Survey for Robot Vision}

\author{Xiaofeng Han\textsuperscript{b,a,*}, 
        Shunpeng Chen\textsuperscript{c,*}, 
        Zenghuang Fu\textsuperscript{b,a}, 
        Zhe Feng\textsuperscript{b,a},
        Lue Fan\textsuperscript{a,b},
        Dong An\textsuperscript{a,b},
        Changwei Wang\textsuperscript{d,e},
        Li Guo\textsuperscript{c}, 
        Weiliang Meng\textsuperscript{a,b,\dag}, 
        Xiaopeng Zhang\textsuperscript{a,b},
        Rongtao Xu\textsuperscript{a,b,\dag},
        Shibiao Xu\textsuperscript{c}}

\address{%
\textsuperscript{a}The State Key Laboratory of Multimodal Artificial Intelligence Systems, Institute of Automation, Chinese Academy of Sciences, China\\[1ex]
\textsuperscript{b}School of Artificial Intelligence, University of Chinese Academy of Sciences, China\\[1ex]
\textsuperscript{c}School of Artificial Intelligence, Beijing University of Posts and Telecommunications, China\\[1ex]
\textsuperscript{d}Key Laboratory of Computing Power Network and Information Security, Ministry of Education; Shandong Computer Science Center, Qilu University of Technology (Shandong Academy of Sciences), China\\[1ex]
\textsuperscript{e} Shandong Provincial Key Laboratory of Computing Power Internet and Service Computing, Shandong Fundamental Research Center for Computer Science, China}

\fntext[myfootnote]{\textsuperscript{*}Equal contribution.\\
\textsuperscript{\dag}Corresponding authors.}

\date{} 

\begin{abstract}
Robot vision has greatly benefited from advancements in multimodal fusion techniques and vision-language models (VLMs). We adopt a task-oriented perspective to systematically review the applications and advancements of multimodal fusion methods and VLMs in the field of robot vision. For semantic scene understanding tasks, we categorize fusion approaches into encoder-decoder frameworks, attention-based architectures, and graph neural networks. Meanwhile, we also analyze the architectural characteristics and practical implementations of these fusion strategies in key tasks such as simultaneous localization and mapping (SLAM), 3D object detection, navigation, and manipulation. We compare the evolutionary paths and applicability of VLMs based on large language models (LLMs) with traditional multimodal fusion methods.Additionally, we conduct an in-depth analysis of commonly used datasets, evaluating their applicability and challenges in real-world robotic scenarios. Building on this analysis, we identify key challenges in current research, including cross-modal alignment, efficient fusion, real-time deployment, and domain adaptation. We propose future directions such as self-supervised learning for robust multimodal representations, structured spatial memory and environment modeling to enhance spatial intelligence, and the integration of adversarial robustness and human feedback mechanisms to enable ethically aligned system deployment. Through a comprehensive review, comparative analysis, and forward-looking discussion, we provide a valuable reference for advancing multimodal perception and interaction in robotic vision. A comprehensive list of studies in this survey is available at \href{https://github.com/Xiaofeng-Han-Res/MF-RV}{https://github.com/Xiaofeng-Han-Res/MF-RV}.

\end{abstract}

\end{frontmatter}


\section{Introduction}
\label{sec:introduction}
\begin{figure*}[ht]
    \centering
    \includegraphics[width=\linewidth]{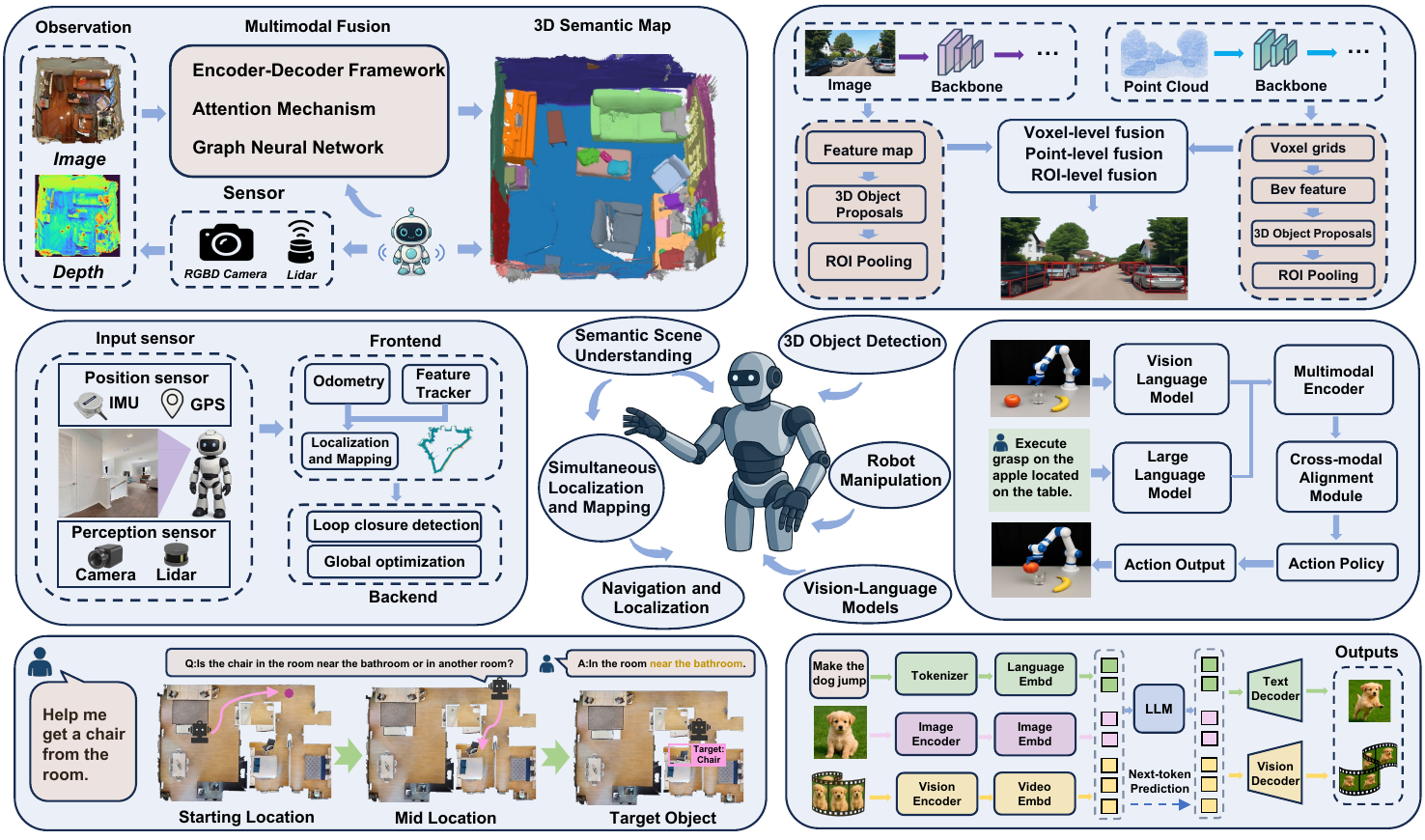}
    \caption{The overview figure illustrates the overall framework of multimodal fusion and VLMs for robot vision. Various sensory inputs (e.g., RGB, Depth, LiDAR, GPS, IMU) are first processed through multimodal fusion strategies, including encoder-decoder frameworks, attention mechanisms, and graph neural networks, to enhance perception. The resulting fused features support core robotic vision tasks such as 3D semantic scene understanding, SLAM, 3D object detection, navigation and localization, and robot manipulation. Vision-language models further bridge perception and reasoning by aligning visual and linguistic information, enabling semantic understanding and action generation. The diagram highlights the integration of traditional fusion pipelines with large vision-language models for complex task execution in robotic systems.}
    \label{fig:overview_figure}
\end{figure*}
With the rapid development of artificial intelligence and machine learning, multimodal fusion and vision-language models (VLMs) have become important tools driving the advancement of robot vision technologies. Traditional unimodal approaches (such as relying solely on RGB images) often encounter perceptual limitations when dealing with complex tasks in real-world environments—such as occlusions, lighting variations, sparse textures, and insufficient semantic information\cite{xu2020zoomnet,shi2021geometry,wang2021depth}. In contrast, multimodal fusion integrates complementary information from various sensor inputs (e.g., vision, language, depth, LiDAR, and tactile data), thereby enhancing the perception, reasoning, and decision-making capabilities of robot vision systems~\cite{xu2024mrftrans,chen2017multi,li2024rd,babadian2023fusion}. The integration of diverse modalities enables robots to achieve more robust scene understanding, stronger task generalization, and more natural human-robot interaction in complex environments.

Building on this foundation, the rapid rise of vision-language models in recent years has further advanced the paradigm of multimodal fusion. Large-scale pretrained VLMs not only possess cross-modal alignment and generalization abilities, but also demonstrate strong potential in tasks such as zero-shot understanding~\cite{radford2019language}, instruction following~\cite{long2024instructnav,zhou2024navgpt}, and visual question answering~\cite{team2023gemini}. This evolution marks a shift in robotic vision systems from passive perception to proactive intelligent systems capable of semantic understanding and natural language interaction.

Despite the promising prospects of multimodal fusion, several practical challenges remain. First, efficiently integrating heterogeneous data from different modalities remains a core issue, involving problems such as modality alignment, unified feature representation, and spatiotemporal synchronization. Second, robotic systems impose stringent requirements on real-time performance and resource efficiency, necessitating a balance between computational cost and model accuracy when designing fusion architectures. Third, although pretrained VLMs have demonstrated strong general capabilities, their adaptability to specific robotic tasks is still limited—particularly in scenarios with limited annotated data or dynamic environments. To address these challenges, researchers have proposed a variety of fusion strategies and solutions. For example, encoder-decoder architectures and Transformer-based structures are widely used to model dependencies between modalities~\cite{Xu_2023_CVPR,bai2022transfusion}; contrastive learning and cross-modal attention mechanisms are applied to improve alignment performance~\cite{xu2024deffusion}; and graph neural networks are used to enhance the modeling of relational structures within scenes~\cite{Aflalo_2023_ICCV}. At the same time, foundation model-driven multimodal pretraining paradigms are gradually being integrated into robot vision systems, showing great potential for scalability and generalization.

This survey provides a systematic review of research progress and key technologies in multimodal fusion and vision-language models for robot vision, as illustrated in Figure\ref{fig:overview_figure}. It covers a wide range of representative tasks, including semantic scene understanding~\cite{wolters2024unleashing}, 3D object detection~\cite{chen2017multi}, embodied navigation~\cite{shah2021ving}, simultaneous localization and mapping (SLAM)~\cite{zuo2019lic}, robotic manipulation~\cite{kim2024openvla}, and visual localization~\cite{mohanty2016deepvo}. Based on the analysis of model architectures and training paradigms, we further explore the connections and differences between traditional multimodal methods and emerging VLMs~\cite{dai2024deepseekmoe,achiam2023gpt}, revealing their complementarity and integration potential. In addition, we analyze widely used multimodal datasets~\cite{caesar2020nuscenes,behley2019semantickitti,khazatsky2024droid,chang2017matterport3d}, highlighting their applicability and limitations in robotic scenarios.

Finally, this paper summarizes the major bottlenecks currently facing multimodal fusion in robot vision and proposes future research directions, including lightweight fusion architectures, efficient pretraining mechanisms, cross-modal self-supervised learning methods, and system optimization strategies for real-world deployment. These directions aim to contribute to the development of more intelligent and generalizable robot vision systems. The main contributions of this paper are as follows:
\begin{itemize}
    \item We systematically integrate traditional multimodal fusion strategies with emerging vision-language models, and conduct a comparative analysis in terms of architectural design, functional characteristics, and applicable tasks, revealing their connections, complementary strengths, and integration potential.
    \item Distinguishing from previous reviews that mainly focus on basic tasks such as semantic segmentation and target detection, this paper extends the scope of the analysis by focusing on emerging application scenarios such as multimodal SLAM, robot manipulation, and embodied navigation, demonstrating the potential of multimodal fusion and VLMs for complex reasoning and long-duration task decision making.
    \item We summarize the key advantages of multimodal systems over unimodal approaches, including enhanced perceptual robustness, semantic expressiveness, cross-modal alignment, and high-level reasoning, highlighting their practical value in dynamic, ambiguous, or partially observable environments.
    \item We provide an in-depth analysis of the mainstream multimodal datasets currently used for robotics tasks, covering their modal combinations, covered tasks, applicable scenarios and limitations, to provide a reference basis for future benchmarking and model evaluation.
    \item We identify the key challenges in multimodal fusion, such as cross-modal alignment techniques, efficient training strategies, and real-time performance optimization. Based on these challenges, we propose future research directions to advance the field.
\end{itemize}
\begin{table}[ht]
\centering
\caption{Comparison of existing surveys and ours in terms of task scope and technical focus. \textbf{Tasks}: Number of covered tasks, including semantic scene understanding, SLAM, 3D object detection, navigation and localization, and robot manipulation. \textbf{Arch.}: Vision-Language Model architecture analysis; \textbf{CM-SSL}: Cross-modal self-supervised learning ;\textbf{LightFusion}: Lightweight fusion methods.}
\label{tab:survey_comparison_simplified}
\scalebox{0.8}{
\begin{tabular}{lcccc}
\hline
\textbf{Survey} & \textbf{\#Tasks} & \textbf{Arch.} & \textbf{CM-SSL} & \textbf{LightFusion} \\
\hline
Zhang et al.~\cite{zhang2021deep} & 2 & \ding{55} & \ding{55} & \ding{55} \\
\hline
Alaba et al.~\cite{alaba2024emerging} & 1 & \ding{55} & \ding{55} & \ding{55} \\
\hline
Li et al.~\cite{li2025benchmark} & 1 & \ding{51} & \ding{55} & \ding{55} \\
\hline
Wu et al.~\cite{wu2024embodied} & 1 & \ding{55} & \ding{55} & \ding{55} \\
\hline
Bordes et al.~\cite{bordes2024introduction} & 1 & \ding{51} & \ding{51} & \ding{55} \\
\hline
Lin et al.~\cite{lin2023advances} & 1 & \ding{51} & \ding{55} & \ding{55} \\
\hline
Chen et al.~\cite{chen2025review} & 1 & \ding{55} & \ding{55} & \ding{55} \\
\hline
Abdulmaksoud et al.~\cite{abdulmaksoud2025transformer} & 2 & \ding{55} & \ding{55} & \ding{55} \\
\hline
\textbf{Ours} & \textbf{5} & \ding{51} & \ding{51} & \ding{51} \\
\hline
\end{tabular}
}
\end{table}

Although recent years have seen several surveys on multimodal fusion and vision-language models, significant limitations remain in terms of task coverage and modeling depth (Table~\ref{tab:survey_comparison_simplified}). On one hand, some surveys, such as Zhang et al.\cite{zhang2021deep} and Alaba et al.\cite{alaba2024emerging}, focus on specific traditional tasks (e.g., semantic segmentation, 3D detection), lacking systematic coverage of critical robotic tasks such as SLAM, embodied navigation, multimodal 3D perception, and manipulation. On the other hand, studies like Li et al.\cite{li2025benchmark} and Bordes et al.\cite{bordes2024introduction} explore the architectural evolution and evaluation paradigms of VLMs, but their analyses are primarily centered on language generation tasks, with insufficient discussion on the practical deployment of vision-language models in real robotic systems.Moreover, although Wu et al.\cite{wu2024embodied} and Lin et al.\cite{lin2023advances} focus on the integration of perception and decision-making in embodied navigation, they do not incorporate tasks such as SLAM and manipulation into a unified framework. Meanwhile, Abdulmaksoud et al.\cite{abdulmaksoud2025transformer} and Chen et al.\cite{chen2025review} provide systematic reviews of sensor fusion methods and LiDAR-Vision SLAM, respectively, yet they do not explore the potential integration of VLMs with self-supervised alignment strategies. In contrast, this survey focuses on the systematic integration and comparison of multimodal fusion methods and vision-language models (VLMs) in the context of robotic vision tasks, covering key applications ranging from semantic scene understanding\cite{lin2025curriflowcurriculumguideddepthfusion} to embodied navigation, SLAM, and robotic manipulation. Moreover, we not only systematically analyze and compare traditional multimodal fusion methods and VLMs, but also highlight often-overlooked research directions such as cross-modal self-supervised learning and lightweight fusion architectures. Our goal is to provide forward-looking insights for building more intelligent, generalizable, and deployable robotic vision systems.
\begin{figure*}[ht]
    \centering
    \includegraphics[width=\linewidth]{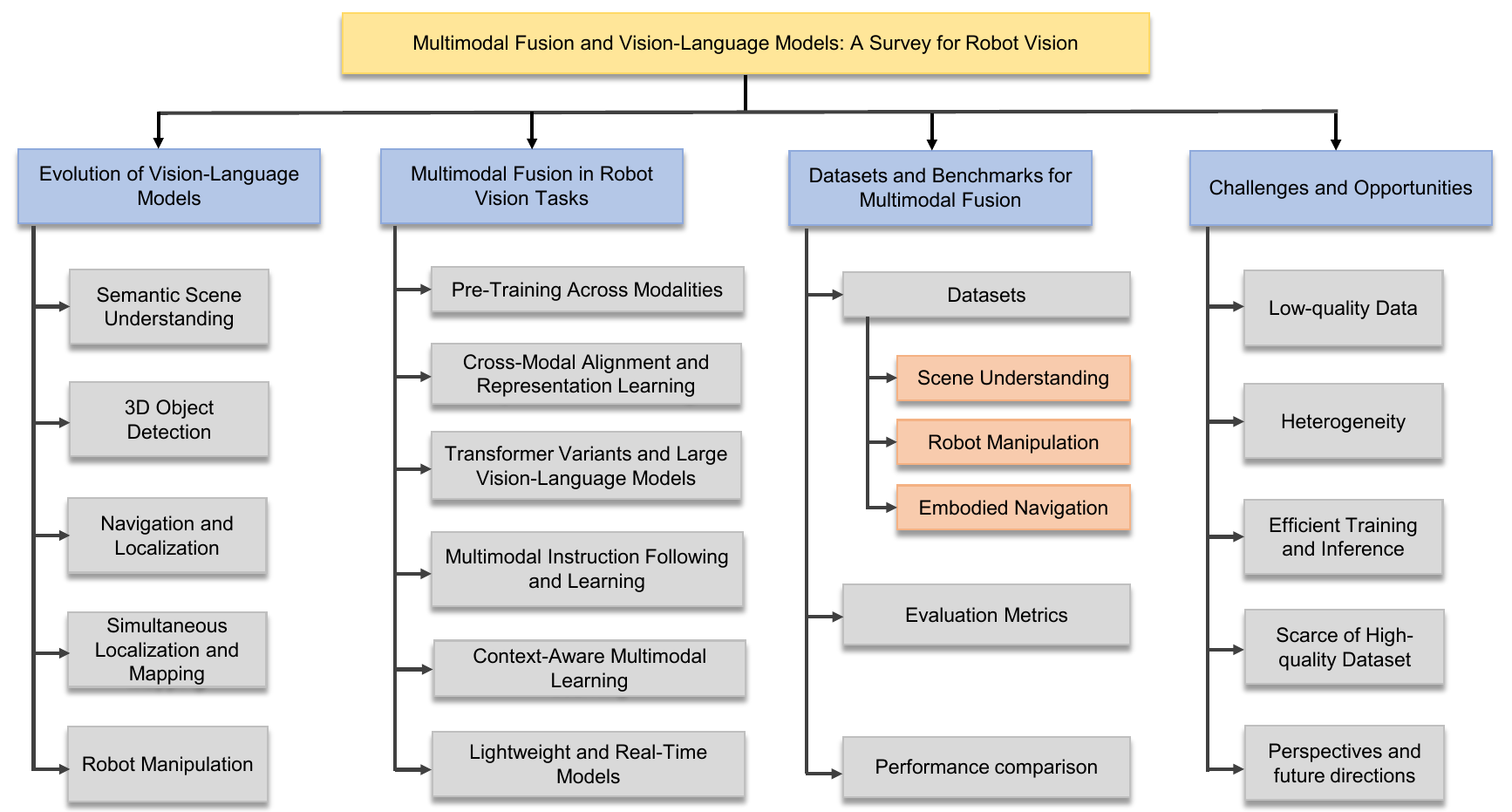}
    \caption{The overall structure of the survey on multimodal fusion and vision-language model in robot vision.}
    \label{fig:Structure}
\end{figure*}
The remaining structure of this paper is shown in Figure \ref{fig:Structure}. First, we systematically review multimodal fusion techniques in various robotic vision tasks (Section \ref{sec:multimodal_fusion_robot_vision}) and introduce the technological evolution of vision-language models (Section \ref{sec:technological_evolution}). Then, we summarize datasets related to robotic vision tasks (Section \ref{sec:Datasets_for_Multimodal_Fusion}), present relevant evaluation metrics, and conduct performance comparisons and analyses (Section \ref{sec:performance_evaluation}). Finally, we discuss future challenges and research directions (Section \ref{sec:challenges_future_directions}) and conclude the survey (Section \ref{sec:conclusions}).
\section{Multimodal Fusion in Robot Vision Tasks}
\label{sec:multimodal_fusion_robot_vision}

\subsection{Semantic Scene Understanding}

Scene semantic understanding is a core task in computer vision, aimed at conducting high-level semantic analysis of scenes in images or videos, including object recognition, segmentation, and modeling of object relationships.Relying on traditional single-modal approaches (e.g., solely using RGB images) can encounter various limitations in complex scenarios, such as occlusion, lighting variations, and multi-object ambiguities~\cite{mikolajczyk2005performance,everingham2010pascal,geiger2012kitti}. Multimodal fusion is a technique developed to overcome the limitations of single-modality approaches. It combines various sources of information, such as vision, language, depth data, lidar, and speech, to enhance the capability of scene semantic understanding~\cite{Zhang2020DeepMF,Rizzoli2022MultimodalSS,Zhao2024DeepMD}.
\begin{figure}[ht]
    \centering
    \includegraphics[width=\linewidth]{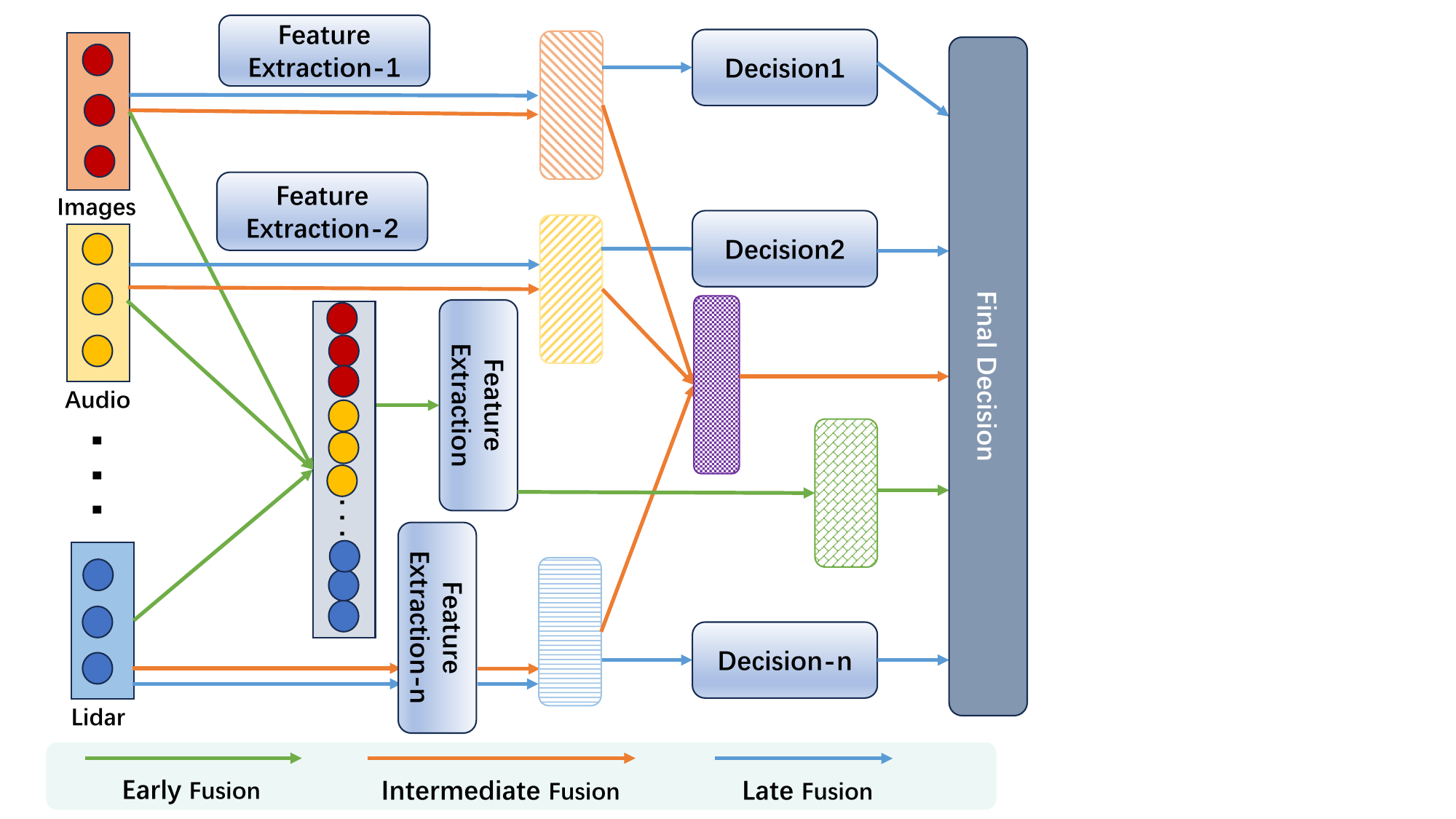}
    \caption{The above diagram intuitively illustrates the basic processes of three main strategies for multimodal fusion: early fusion, mid fusion, and late fusion. In the figure, each modality data (such as images, audio, Lidar, text, etc.) is independently processed by a feature extractor. Early fusion directly fuses data from different modalities before feature extraction; Mid term fusion combines modal features through specific mechanisms such as feature concatenation or weighting after extracting them; Late stage fusion is achieved by integrating the decision results of each modality after independent decision-making is completed. The diagram clearly reflects the key differences and roles of the three fusion methods in the multimodal processing flow.}
    \label{fig:picture1}
\end{figure}
Multimodal fusion plays a crucial role in semantic scene understanding. This fusion reduces sensor limitations (e.g., occlusions, noise) and enhances scene interpretation robustness. As illustrated in Figure \ref{fig:picture1}, multimodal fusion strategies can be categorized into early fusion, mid fusion, and late fusion based on the fusion strategy~\cite{Gaw2021MultimodalDF,Gandhi2022MultimodalSA}. Early fusion directly integrates multimodal inputs at the data level, such as concatenating RGB~\cite{Xiao2019MultimodalEA} images and depth maps before feeding them into the model. Its advantage lies in its simple structure and the ability to directly utilize the raw data characteristics, while its disadvantage is that the features between modalities may not match, making it difficult to handle high-dimensional data.Mid term fusion occurring in the middle layer is a common method for balancing modal independence and capturing semantic relationships. This method typically employs attention mechanisms, feature weighting, connectivity, or GNNs to combine modality specific features~\cite{Mai2019ModalityTM,9190483}. A common attention mechanism similar to transformer structure has been proposed to improve the applicability of different modal data and capture local feature correlations~\cite{Zhang2024Multimodal}. In addition, adversarial representation learning has been used to create modality invariant embedding spaces, reduce modal gaps, and improve cross modal fusion. These methods demonstrate high flexibility, adaptability, and robustness in complex scenarios, effectively enhancing semantic understanding of the scene~\cite{Valada2018SelfSupervisedMA}.Post fusion is a key method in multimodal analysis, which combines the results of decision level independent processing of modalities ~\cite{Atrey2010MultimodalFF}. Common techniques include weighted averaging, voting mechanisms, and logical rules . This method has multiple advantages, such as strong modal independence, ease of individual optimization, and scalability of multimodal systems~\cite{Wu2004OptimalMF} . Roitberg et al.~\cite{Roitberg2022ACA} compared and analyzed seven decision-level fusion strategies for driver behavior understanding, offering valuable insights for selecting appropriate fusion schemes.

Traditional multimodal fusion methods offer a basic theoretical framework but struggle with complex data. With deep neural networks, feature extraction, modality interaction, and decision-making have become deeply integrated, making fusion stages less distinct. This has driven a shift from explicit to implicit fusion~\cite{Zhao2024DeepMD}, where network design inherently captures modality relationships, enhancing deep semantic extraction.
This chapter categorizes multimodal fusion approaches in semantic scene understanding into three main directions: encoder-decoder frameworks~\ref{subsubsec:Encoder_Decoder_Framework}, attention-based fusion~\ref{subsubsec:Attention_Mechanism}, and graph neural network (GNN)-based modeling~\ref{subsubsec:Graph_Neural_Network}. The encoder-decoder method~\cite{Wang2024EfficientND} efficiently represents scene semantics through encoding, interaction, and decoding. Attention-based fusion~\cite{Liu2022FocusYA,Kuga2017MultitaskLU,Wu2019MultimodalSA} aligns key features and dynamically assigns weights between modalities for deeper semantic capture. GNN-based methods~\cite{Misraa2020MultiModalRU,Singh2022MultimodalVU} model multimodal data through graph structures, enhancing global and robust semantic understanding. These approaches collectively advance multimodal fusion for semantic scene understanding.


\subsubsection{Encoder-Decoder Framework}
\label{subsubsec:Encoder_Decoder_Framework}

In multimodal fusion, the encoder-decoder framework first extracts features from different modalities (e.g., image, text, etc.), with the encoder independently processing each modality to generate high-dimensional representations. The decoder then fuses these features to produce the final output. During this process, cross-modal feature fusion is typically achieved through early or deep fusion, often optimized by incorporating attention mechanisms to enhance the flow of information and focus on key features. This allows the encoder-decoder framework to effectively handle complex multimodal tasks.

DeepLabv3+~\cite{Chen_2018_ECCV} is a classic encoder-decoder framework for semantic segmentation tasks. The encoder uses a deep convolutional neural network (such as ResNet) with dilated convolutions, while the decoder restores spatial details through upsampling and skip connections.PointRend~\cite{Kirillov_2020_CVPR} proposes an image segmentation method based on an encoder-decoder framework, treating the segmentation task as a rendering problem and adaptively selecting key points in the decoder for fine-grained segmentation. HRNet~\cite{9052469} achieves precise semantic segmentation by preserving high-resolution feature maps. It uses parallel multi-resolution branches and fuses features in the decoder to maintain spatial details. This design excels in segmentation tasks without relying on complex attention mechanisms.
DDRNet~\cite{hong2021deep} proposes a dual-resolution network architecture for real-time semantic segmentation. Its encoder extracts features through dual-resolution branches, and the decoder generates segmentation results by feature fusion.MAMT~\cite{Cheng_2022_CVPR} is a general image segmentation method based on an encoder-decoder framework. Its encoder extracts image features, and the decoder generates segmentation results through mask prediction.PIDNet~\cite{Xu_2023_CVPR} proposes a parallel decoder architecture for real-time semantic segmentation. Its encoder extracts multi-level features, and the decoder processes features of different resolutions through parallel branches, ultimately fusing them to generate segmentation results.

Early multimodal fusion methods primarily relied on the encoder-decoder framework to extract and combine information from different modalities. Nevertheless, as research has progressed, contemporary methodologies have incorporated a range of attention mechanisms alongside the encoder-decoder framework, thereby augmenting the flexibility and precision of information fusion. These attention mechanisms effectively guide the model to focus on key features across modalities, thereby improving the performance of multimodal learning, especially in handling complex tasks.Recent innovations such as elastic deep multi-view autoencoders with diversity embedding~\cite{daneshfar2025elastic} further enhance the diversity and robustness of multimodal representations, complementing attention-based fusion strategies.

\subsubsection{Attention Mechanism}
\label{subsubsec:Attention_Mechanism}
In multimodal fusion for semantic scene understanding, the attention mechanism, with its powerful adaptive weighting ability, effectively captures long-range dependencies between cross-modal features, enhancing the association between different modal information and thereby improving the model's expressive power and generalization ability. Its parallel computing capability also accelerates the processing of large-scale datasets, making it particularly suitable for real-time scene understanding tasks. By utilizing a unified model architecture, the attention mechanism reduces the complexity of data preprocessing and strengthens the fusion of multimodal data, becoming the mainstream method in the field of multimodal fusion.

\begin{figure}[ht]
    \centering
    \includegraphics[width=1\linewidth]{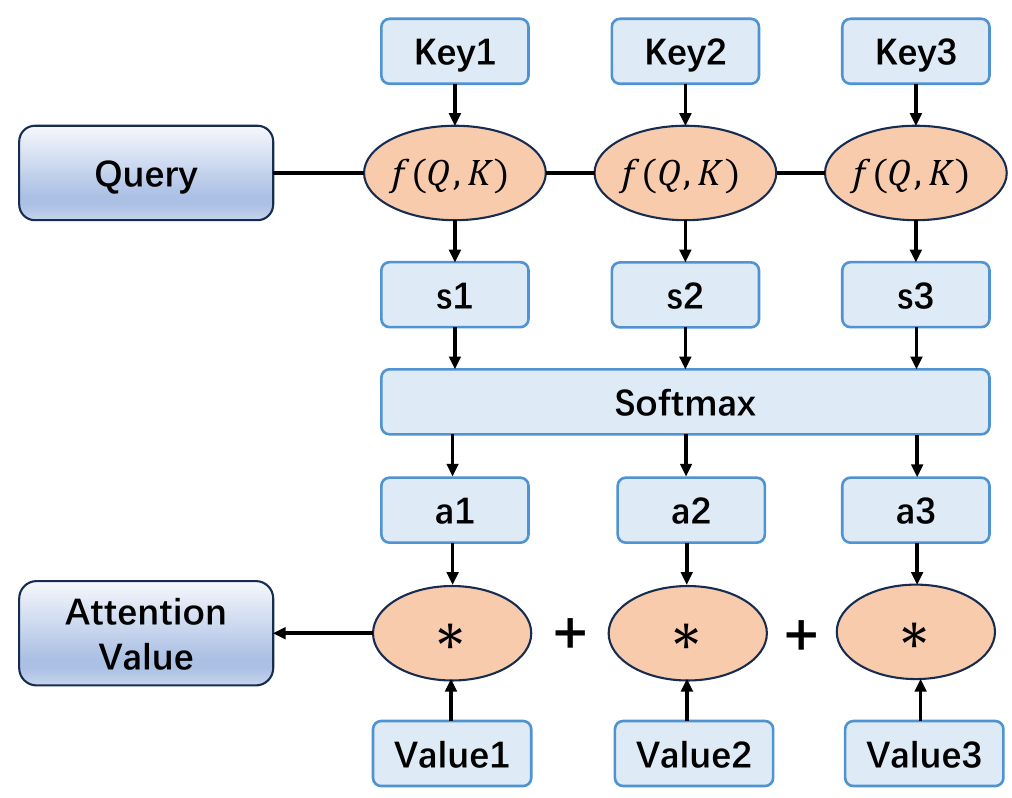}
    \caption{Illustration of the standard self-attention mechanism. The query vector is compared against all key vectors using a compatibility function \( f(Q, K) \), typically dot product. The resulting scores \( s_1, s_2, s_3 \) are normalized via the Softmax function to obtain attention weights \( a_1, a_2, a_3 \), which are then used to compute a weighted sum over the value vectors \( V_1, V_2, V_3 \), producing the final attention output.}
    \label{fig:attention}
\end{figure}

Several recent studies have exemplified the effectiveness of attention-based frameworks in multimodal scene understanding, particularly in 3D scene completion, segmentation, and object detection. For instance,MRFTrans~\cite{xu2024mrftrans}, a transformer-based framework, fuses semantic, geometric, and depth information to address the challenge of monocular 3D scene completion, achieving state-of-the-art performance on SemanticKITTI and NYUv2 datasets by effectively modeling long-range dependencies and refining depth estimates. DefFusion~\cite{xu2024deffusion} introduces a deformable transformer mechanism that dynamically adjusts to multimodal data, mitigating over-attention and noise, and significantly improving 3D semantic segmentation, particularly on large-scale benchmarks like NuScenes.On the other hand ,DMA~\cite{li2024dense} focuses on open-vocabulary 3D scene understanding, aligning dense correspondences between 3D points, 2D images, and text, utilizing vision-language models to generalize across diverse indoor and outdoor scenarios. Lastly, Multi-Sem Fusion~\cite{xu2024multi} integrates LiDAR and camera data for 3D object detection, employing adaptive attention and deep feature fusion to enhance detection accuracy, although its reliance on offline processing limits real-time applicability.
The CLFT~\cite{gu2024clft} model achieves the fusion of camera and LiDAR data by projecting LiDAR point clouds onto the camera plane, upsampling them into 2D feature maps, and then inputting them together with camera images into a Transformer-based network. The model utilizes a bidirectional network and a cross-fusion strategy to integrate data from both sensors at the Transformer decoder layer, enhancing the accuracy and robustness of semantic segmentation in autonomous driving.
However, it is important to note that although the attention mechanism performs well in multimodal alignment, the standard self-attention operation is computationally expensive, with its time and memory complexity growing quadratically with input length,as show in the
standard self-attention operation is defined as:
\begin{equation}
\text{Attention}(Q, K, V) = \text{softmax}\left( \frac{QK^\top}{\sqrt{d_k}} \right)V,
\end{equation}
as illustrated in Figure~\ref{fig:attention}, where $Q, K, V \in \mathbb{R}^{L \times d_k}$ are the query, key, and value matrices, $L$ is the input sequence length, and $d_k$ is the feature dimension. This operation requires pairwise interactions between all tokens in the sequence.
Its computational complexity is:
\begin{equation}
\mathcal{O}(L^2 \cdot d_k),
\end{equation}
which becomes a bottleneck for long sequences or high-resolution visual inputs in resource-constrained environments.This challenge is further intensified in multimodal settings where high-resolution images and long text sequences must be processed jointly. Although approaches such as sparse attention, linear attention, and low-rank approximations have been proposed to alleviate this burden, a trade-off still exists between efficiency and alignment accuracy, which must be carefully balanced in time-critical applications.

Occupancy Prediction is vital for multimodal scene understanding, predicting semantic labels for occupied spatial regions, which is crucial for autonomous navigation, obstacle avoidance, and decision-making. However, many occupancy prediction methods rely mainly on visual data, leading to performance drops in adverse conditions like rain or fog. To address this, researchers integrate additional sensors like LiDAR and Radar to improve perception in complex environments. For example, the CONet approach within the OccNet framework~\cite{wang2023openoccupancy} combines vision with LiDAR and Radar to overcome environmental challenges. Methods like OccuFusion~\cite{wang2023openoccupancy} use SENet modules~\cite{hu2018squeeze} to adaptively weight modality features, improving generalization and reliability. Additionally, HyDRa~\cite{wolters2024unleashing} models uncertainty and enforces depth consistency constraints, enhancing cross-modal depth estimation. Overall, attention mechanisms, especially cross-modal and self-attention, help emphasize useful features while reducing redundancy, improving occupancy prediction model robustness and generalization.

\subsubsection{Graph Neural Network}
\label{subsubsec:Graph_Neural_Network}

\begin{figure}[ht]
    \centering
    \includegraphics[width=1\linewidth]{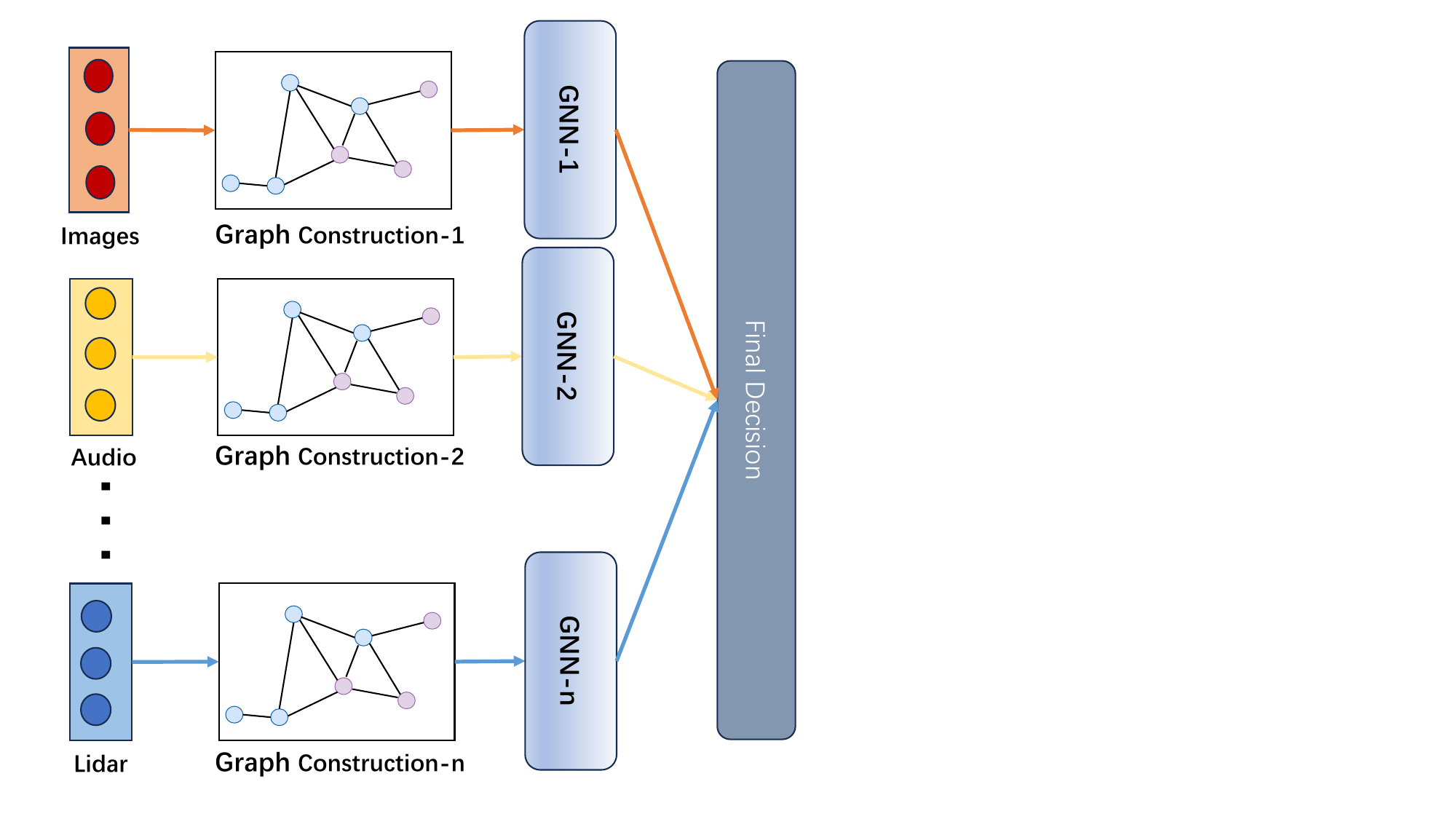}
    \caption{The above diagram intuitively illustrates the workflow of multimodal fusion using Graph Neural Networks (GNN). In the figure, each modality data (such as images, audio, Lidar, text, etc.) is first transformed into a graph structure through a graph construction process. These graphs are then processed by separate GNNs to extract high-level representations. The final decision stage integrates the outputs from different GNNs, enabling effective multimodal reasoning. This diagram clearly demonstrates the role of GNNs in learning structured relationships within each modality and highlights the fusion process in multimodal understanding.}
    \label{fig:gnn}
\end{figure}
In multimodal semantic scene understanding tasks, Graph Neural Networks (GNNs) initially construct a graph structure by mapping multimodal data (such as images and text) onto nodes and edges of a graph. Through message passing and feature aggregation, GNNs then extract and fuse high-level semantic representations from different modalities. As illustrated in Figure \ref{fig:gnn}, data from each modality (e.g., modality data 1, modality data 2) is first transformed into graph structures independently, then processed through separate GNN networks to extract high-level semantic features. The final decision-making stage integrates outputs from these separate GNNs, enabling effective multimodal reasoning. This approach leverages the powerful relational modeling capabilities of GNNs, their flexible multimodal fusion strategies, high scalability, and generalizability across tasks, making them particularly suitable for complex scene understanding tasks, while also maintaining good interpretability.

\begin{table}[h!]
\centering
\caption{Representative multimodal fusion methods categorized by fusion strategy and publication year.}
\label{tab:multimodal_fusion_methods}
\scalebox{0.85}{
\begin{tabular}{llc}
\hline
\textbf{Multimodal fusion Type} & \textbf{Representative Methods} & \textbf{Year} \\ 
\hline

\multirow{6}{*}{\textbf{Encoder-Decoder Framework}} 
& PIDNet~\cite{Xu_2023_CVPR} & 2023 \\
& MAMT~\cite{Cheng_2022_CVPR} & 2022 \\
& DDRNet~\cite{hong2021deep} & 2021 \\
& PointRend~\cite{Kirillov_2020_CVPR} & 2020 \\
& HRNet~\cite{9052469} & 2019 \\
& DeepLabv3+~\cite{Chen_2018_ECCV} & 2018 \\
\hline

\multirow{9}{*}{\textbf{Attention Mechanism}} 
& MRFTrans~\cite{xu2024mrftrans} & 2024 \\
& DefFusion~\cite{xu2024deffusion} & 2024 \\
& DMA~\cite{li2024dense} & 2024 \\
& Multi-Sem Fusion~\cite{xu2024multi} & 2024 \\
& CLFT~\cite{gu2024clft} & 2024 \\
& OccFusion~\cite{ming2024occfusion} & 2024 \\
& HyDRa~\cite{wolters2024unleashing} & 2024 \\
& CONet~\cite{wang2023openoccupancy} & 2023 \\
& SENET~\cite{hu2018squeeze} & 2018 \\
\hline

\multirow{6}{*}{\textbf{Graph Neural Network}} 
& MISSIONGNN~\cite{yun2024missiongnn} & 2024 \\
& VQA-GNN~\cite{wang2023vqa} & 2023 \\
& MuSe-GNN~\cite{liu2023muse} & 2023 \\
& CCA-GNN~\cite{passos2023multimodal} & 2023 \\
& MGA~\cite{dutta2023performance} & 2023 \\
& MGNNS~\cite{yang2021multimodal} & 2021 \\
\hline

\end{tabular}
}
\end{table}

Building on this paradigm, numerous multimodal GNN-based approaches have emerged in recent years, which aim to enhance representation learning and semantic reasoning by unifying modality-specific structures and enabling information propagation through graph-based mechanisms. One line of research focuses on constructing separate graphs for each modality and performing alignment and fusion at a higher semantic level. For instance, MISSIONGNN~\cite{yun2024missiongnn} employs lightweight hierarchical GNNs built upon task-specific knowledge graphs automatically generated by large language models, enabling efficient frame-level semantic reasoning. Another class of methods favors the construction of unified semantic graphs, where entities from different modalities are jointly encoded into a shared multimodal structure. VQA-GNN~\cite{wang2023vqa} exemplifies this approach by introducing a shared context node to bridge visual and conceptual subgraphs, facilitating bidirectional message passing to mitigate modality-level representational gaps. To enhance temporal consistency and modality alignment, CCA-GNN~\cite{passos2023multimodal} proposes a temporal positional encoding scheme for graph nodes and aligns audio-visual representations through canonical correlation analysis. MuSe-GNN~\cite{liu2023muse} further integrates modality-specific entities into a unified semantic space by leveraging shared GNN encoders and jointly optimizing similarity- and contrastive-based objectives. In addition, models such as MGA~\cite{dutta2023performance} and MGNNS~\cite{yang2021multimodal} demonstrate that heterogeneous GNNs and multi-channel architectures can effectively bridge structural and semantic modality gaps by integrating multimodal graph representations. Collectively, these approaches reflect a common design principle: modeling both intra- and inter-modal relationships through graph structures, and advancing multimodal fusion through flexible and interpretable graph-based reasoning frameworks.

\subsection{3D Object Detection}

\subsubsection{Lidar+Camera}
In the multimodal fusion of robotic vision systems, the effectiveness and adaptability of fusion strategies are primarily determined by three key design considerations: when to fuse, what to fuse, and how to fuse. These factors play a crucial role in determining the performance of multimodal perception systems, especially in complex applications such as autonomous driving.
\begin{figure}[ht]
    \centering
    \includegraphics[width=\linewidth]{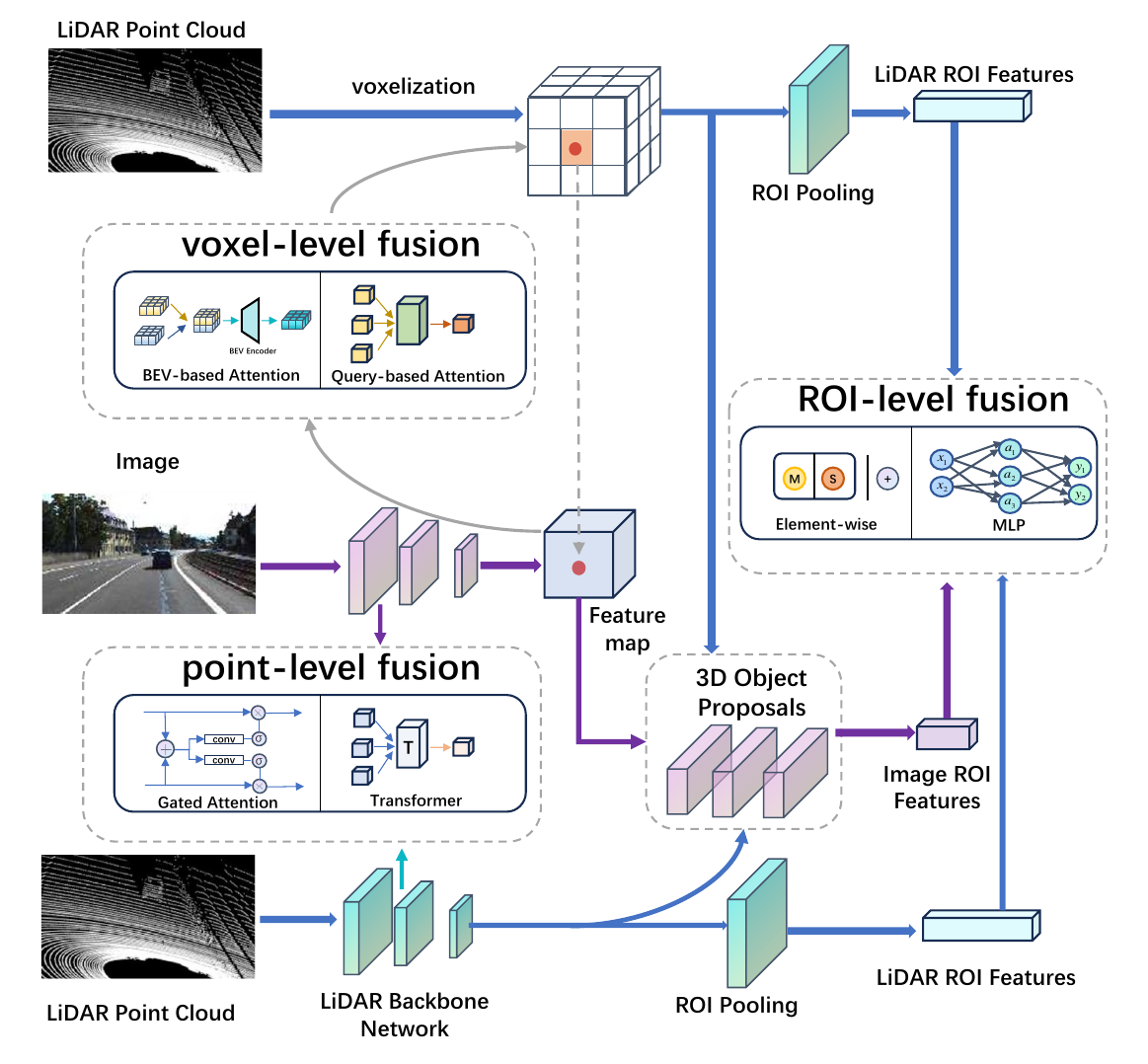}
    \caption{The multimodal fusion framework for 3D object detection includes voxel-level, point-level, and ROI-level fusion to integrate LiDAR and image features and improve detection accuracy.}
    \label{fig:3D_object_detection}
\end{figure}
\begin{itemize}
    \item \textbf{When to Fuse:} The timing of fusion can be categorized into early fusion, intermediate fusion, and late fusion. Early fusion occurs at the raw data stage, integrating sensor inputs. Intermediate fusion takes place after feature extraction, where features from different modalities are combined. Late fusion happens after each modality makes independent decisions, and these decisions are subsequently merged.
    \item \textbf{What to Fuse:} For camera inputs, common data elements include feature maps, attention maps, and pseudo-LiDAR point clouds. For LiDAR, typical inputs consist of raw point clouds, voxelized representations, and multi-view projections such as BEV (Bird's Eye View), FV (Front View), and RV (Rear View).
    \item \textbf{How to Fuse:} As shown in Figure \ref{fig:3D_object_detection}, in LiDAR-Camera systems, the granularity of fusion can be divided into ROI-wise, voxel-wise, and point-wise approaches. Fusion techniques can be further classified as attention-agnostic or attention-based. Early fusion methods, such as those used in MV3D~\cite{chen2017multi} and AVOD~\cite{ku2018joint}, are generally attention-agnostic, relying on direct concatenation or feature pooling to combine multimodal data. With the rise of attention mechanisms and transformer architectures, attention-based fusion techniques have become increasingly important in 3D object detection, enabling more dynamic and context-aware integration of features.
\end{itemize}
\label{subsec:autonomous_driving}

3D object detection is a crucial perception task in autonomous driving systems. Its primary goal is to acquire three-dimensional environmental information through various sensors, identify and localize targets (e.g., vehicles, pedestrians, obstacles), and provide support for path planning and decision-making.In autonomous driving, common sensors include Cameras, LiDARs, Radars, and ultrasonic sensors, etc. A single sensor often cannot meet the requirements for safe and reliable perception. Multi-modal fusion combines the advantages of different types of sensors to make up for their respective shortcomings, thus providing more comprehensive and accurate environmental perception capabilities.

Camera-based single-modal 3D object detection \cite{xu2020zoomnet,park2021pseudo,shi2021geometry,wang2021depth} infers 3D information from 2D images. While RGB images provide rich visual data, the lack of direct depth perception requires complex estimation techniques, making detection vulnerable to occlusions, poor angles, and adverse weather conditions. These factors degrade accuracy and robustness. In contrast, LiDAR-based detection \cite{yang2018ipod,ye2020hvnet,deng2021voxel,mao2021voxel} processes 3D point cloud data, offering precise depth estimation and spatial structure representation. However, LiDAR data is often sparse, especially at long distances or on low-reflectivity surfaces, making it challenging to detect small or distant objects. Additionally, the absence of color and texture limits its ability to distinguish visually similar but categorically different objects.
Therefore, although single-modal 3D object detection methods each have their own advantages, they all have certain limitations. To solve this problem, more and more multi-modal fusion strategies are adopted in autonomous driving systems, among which the fusion of Cameras and LiDAR point clouds is a main research direction.

\begin{table}[h!]
\caption{3D Object Detection Methods. LiDAR-Camera fusion combines depth accuracy and semantic richness, while other sensor fusion methods enhance robustness. Recent advances leverage Transformers and adaptive fusion for improved performance.}

\label{tab:3d_Object_Detection_Methods}
\scalebox{0.65}{
\begin{tabular}{cll}
\hline
Type  &\multicolumn{1}{c}{2017-2021}                  &\multicolumn{1}{c}{2022-2024}                             \\ \hline Lidar+Camera       & \begin{tabular}[c]{@{}l@{}} MV3D~\cite{chen2017multi}$_{CVPR’2017}$\\
AVODc$_{IROS’2018}$\\
Pointfusion~\cite{xu2018pointfusion}$_{CVPR’2018}$\\
MVX-Net~\cite{sindagi2019mvx}$_{ICRA’2019}$\\
PointPainting~\cite{vora2020pointpainting}$_{CVPR’2020}$\\
3D-CVF~\cite{yoo20203d}$_{ECCV’2020}$\\
EPNet~\cite{huang2020epnet}$_{ECCV’2020}$\\
FusionPainting~\cite{xu2021fusionpainting}$_{ITSC’2021}$\\
\end{tabular}                                  & \begin{tabular}[c]{@{}l@{}}
TransFusion~\cite{bai2022transfusion}$_{CVPR’2022}$\\
UVTR~\cite{li2022unifying}$_{NeurIPS’2022}$\\
DeepFusion~\cite{li2022deepfusion}$_{CVPR’2022}$\\
DeepInteraction~\cite{yang2022deepinteraction}$_{NeurIPS’2022}$\\
CMT~\cite{yan2023cross}$_{ICCV’2023}$\\

BEVFusion~\cite{liu2023bevfusion}$_{CVPR’2023}$\\ 
GAFusion~\cite{li2024gafusion}$_{CVPR’2024}$\\
IS-Fusion~\cite{yin2024fusion}$_{CVPR’2024}$\\
\end{tabular}             \\ \hline
Other Sensor Fusion      
& \begin{tabular}[c]{@{}l@{}}
RRPN~\cite{nabati2019rrpn}$_{ICIP’2019}$\\ 
FusionNet~\cite{lim2019radar}$_{NeurIPS’2019}$\\ 
BIRANet~\cite{yadav2020radar+}$_{ICIP’2020}$\\ 
CenterFusion~\cite{nabati2021centerfusion}$_{WACV’2021}$\\ 
AssociationNet~\cite{dong2021radar}$_{CVPR’2021}$\\
RISFNet~\cite{cheng2021robust}$_{ICCV’2021}$\\
MVDNet~\cite{qian2021robust}$_{CVPR’2021}$\\
\end{tabular}                                      & \begin{tabular}[c]{@{}l@{}}
InterFusion~\cite{wang2022interfusion}$_{IROS’2022}$\\
Simple-BEV~\cite{harley2023simple}$_{ICRA’2023}$\\
CRAFT~\cite{kim2023craft}$_{AAAI’2023}$\\ 
CRN~\cite{kim2023crn}$_{ICCV’2023}$\\ 
RCBEVDet~\cite{lin2024rcbevdet}$_{CVPR’2024}$\\  
RCM-Fusion~\cite{kim2024rcm}$_{ICRA’2024}$\\
CRKD~\cite{zhao2024crkd}$_{CVPR’2024}$\\
\end{tabular}    \\ \hline 

\end{tabular}}
\end{table}
Early LiDAR and camera feature fusion methods were non-attention-based, such as element-wise mean fusion operations, weighted fusion, MLP, etc. MV3D~\cite{chen2017multi} is a representative work that utilizes multi-view (BV, FV, RGB) feature fusion of LiDAR and cameras. It first maps point cloud data into a bird's eye view to generate highly accurate 3D candidate boxes, then projects these candidates into multiple views to extract features from different perspectives. It then employs a deep fusion approach where intermediate layers from different views interact with each other, and feature fusion is performed using an element-wise mean method. The features from each layer are used as inputs to the subsequent layers of the sub-networks. MV3D~\cite{chen2017multi} has lower detection accuracy for small objects. To address this issue, AVOD~\cite{ku2018joint} designed a high-resolution feature extractor based on an encoder-decoder structure. Unlike the ROI-level fusion of MV3D~\cite{chen2017multi} and AVOD~\cite{ku2018joint}, the VoxelFusion fusion strategy proposed in MVX-Net~\cite{sindagi2019mvx} is voxel-level. VoxelFusion can fuse at the voxel feature encoding layer, aggregating image features into corresponding voxels, and concatenating them with voxel point cloud features. Point-level fusion usually occurs at an earlier stage, facilitating ample interaction between point cloud features and image features. PointFusion~\cite{xu2018pointfusion} proposed a dense fusion network that fuses point cloud features extracted by PointNet~\cite{qi2017pointnet} and image features extracted by ResNet~\cite{he2016deep}. The feature vectors are concatenated and then passed through multiple fully connected layers; the fused features are used to predict the offsets of the eight corners of the target bounding box to the input points. PointPainting~\cite{vora2020pointpainting} projects LiDAR point clouds onto the output of the image semantic segmentation network and concatenates the channel activation values with the LiDAR point cloud intensity values, effectively utilizing image semantic information. EPNet~\cite{huang2020epnet} introduces a Li-Fusion module that adaptively fuses LiDAR and image features at the point level. It first maps the two features to the same channel and adds them, then generates a weight map to estimate the importance of the image features. The image features are multiplied element-wise with the weight map and concatenated with the point cloud features.

Attention-based methods enhance object detection by dynamically focusing on key multimodal information. 3D-CVF~\cite{yoo20203d} employs adaptive gating fusion with attention maps for feature weighting. FusionPainting~\cite{xu2021fusionpainting} integrates 2D and 3D segmentation to address boundary blurring. DeepFusion~\cite{li2022deepfusion} ensures precise alignment of LiDAR and image features using InverseAug and cross-attention mechanisms. Unlike hard association methods~\cite{chen2017multi, ku2018joint, yoo20203d}, TransFusion~\cite{bai2022transfusion} introduces a soft association fusion framework using a Transformer decoder for high-precision 3D detection. The first layer generates initial 3D boxes with object queries, while the second layer refines them using cross-attention. Additionally, an image-guided query strategy improves detection in sparse point clouds. UVTR~\cite{li2022unifying} converts images into voxel representations, enabling cross-modal learning. DeepInteraction~\cite{yang2022deepinteraction} addresses the limitation of unilateral fusion by introducing a multimodal interaction encoder and decoder. CMT~\cite{yan2023cross} enhances object localization by encoding 3D coordinates and using a Transformer decoder for multimodal interaction. 
BEVFusion\cite{liu2023bevfusion} aligns image and point cloud features in a unified BEV space, achieving high accuracy and efficiency in multimodal 3D perception tasks, particularly in large-scale scenarios such as autonomous driving. However, its image branch lacks spatial geometric modeling capabilities, making it prone to spatial reasoning errors in scenarios involving occlusions, viewpoint variations, or complex structures. In robotic vision tasks, although BEV representations provide a comprehensive global view, they are susceptible to degraded robustness in unstructured environments due to sparse point clouds, image blur, and depth estimation noise. Furthermore, the method relies on multi-modal projection and fusion operations that are computationally expensive, posing challenges for real-time deployment on resource-constrained platforms. The follow-up work GAFusion\cite{li2024gafusion} addresses the spatial awareness limitations by introducing sparse depth guidance and laser occupancy maps, and incorporates a laser-guided adaptive fusion transformer along with a temporal fusion module to enhance inter-modal and inter-frame consistency. However, these enhancements further increase the model complexity and inference cost, limiting its practicality on embedded robotic systems, especially in scenarios requiring both high frame rates and low power consumption.IS-Fusion~\cite{yin2024fusion} employs instance-guided fusion and hierarchical scene modeling to jointly capture fine-grained semantics and global contextual information, demonstrating strong performance in object detection and 3D scene understanding. While the method excels in structured and object-dense autonomous driving environments, it faces limitations in robotic applications involving dynamic, cluttered, or partially observable scenes, where its reliance on high-quality instance features can degrade performance. For example, in conditions with occlusions, low lighting, or incomplete observations, instance masks may be difficult to obtain accurately, reducing the effectiveness of the fusion process. Moreover, the method's multi-level fusion architecture introduces substantial computational overhead and lower inference efficiency, posing challenges for deployment in control-critical robotic systems. Additionally, as its experiments are primarily based on autonomous driving datasets, the method's generalization to common indoor tasks such as navigation and manipulation in service robots remains to be further validated.
\begin{figure*}[ht]
    \centering
    \includegraphics[width=\linewidth]{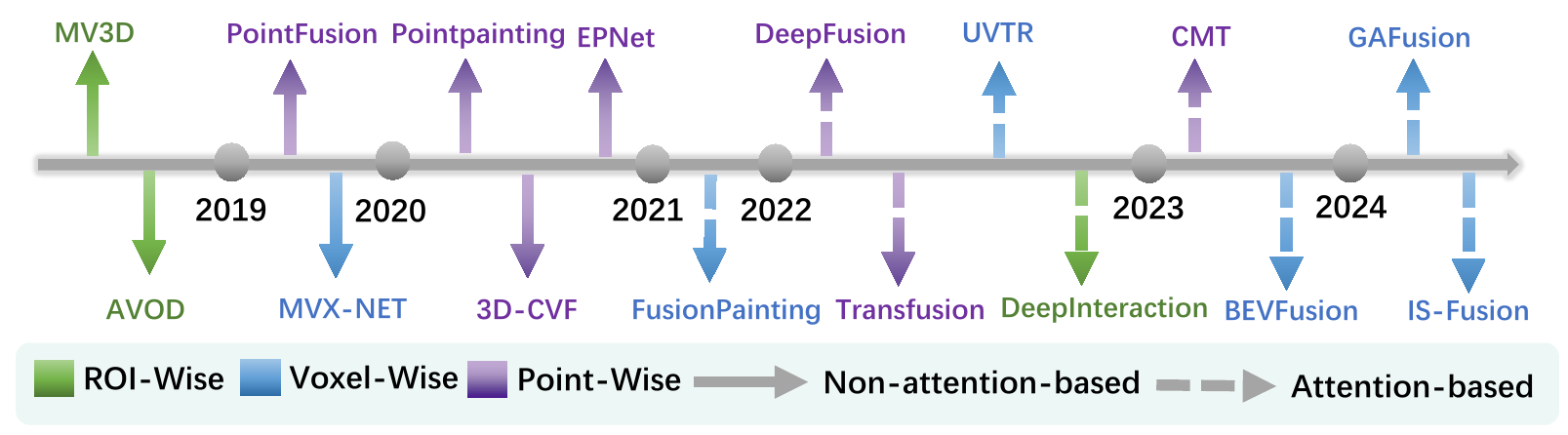}
    \caption{Multimodal fusion for 3D object detection methods.The methods are classified based on fusion granularity and fusion methods, with fusion granularity divided into ROI-wise, Voxel-wise, and Point-wise, and fusion methods divided into attention-based and non-attention-based.In 2017, MV3D~\cite{chen2017multi} pioneered the fusion of Lidar's projections in BEV (Bird's Eye View) and FV (Front View) with RGB features, followed by AVOD~\cite{ku2018joint}, which uses FPN to extract high-resolution features for feature fusion; methods such as Pointfusion~\cite{xu2018pointfusion}, MVX-Net~\cite{sindagi2019mvx}, and Pointpainting~\cite{vora2020pointpainting} have further explored the benefits of different fusion granularity feature fusion advantages. In recent years, models such as TransFusion~\cite{bai2022transfusion}, UVTR~\cite{li2022unifying}, and DeepFusion~\cite{li2022deepfusion} use the Transformer architecture to enhance cross-modal information exchange, and IS-Fusion~\cite{yin2024fusion} improves the understanding of complex environments through the instance-scene collaborative fusion mechanism.}
    \label{fig:methods}
\end{figure*}
As shown in Figure \ref{fig:methods}, in the field of LiDAR-Camera fusion, we have witnessed the evolution from early simple feature level fusion to today's complex attention mechanisms and Transformer architectures. As shown in Table \ref{tab:3d_Object_Detection_Methods}, Transformer architectures have gradually become mainstream, giving rise to a number of new models such as TransFusion, BEVFusion, and MV2DFusion. By leveraging multi-head attention mechanisms and modality-specific structures, these models enable more flexible and efficient feature fusion, enhancing the model's ability to understand complex scenes and multimodal semantics.These advanced fusion strategies have not only improved the accuracy of 3D object detection but also enhanced the model's adaptability and robustness to complex environments. With the advancement of deep learning and sensor technology, future fusion methods are expected to achieve higher precision and lower computational costs, while also better handling challenges in dynamic scenes, such as occlusions and rapidly moving objects.

\subsubsection{Other Sensor Fusion}
In addition to Lidar-Camera fusion, there are studies based on Radar-Camera and Lidar-Radar. RRPN~\cite{nabati2019rrpn} maps radar detections to the image coordinate system, generating anchor boxes for quick ROI suggestions. CenterFusion~\cite{nabati2021centerfusion} uses a staccato-based method to correlate radar data with camera centroids, enhancing 3D object detection accuracy, especially in dynamic environments by integrating velocity and depth information. BIRANet~\cite{yadav2020radar+} improves target detection using Attentive FPN with spatial and channel scSE modules. AssociationNet~\cite{dong2021radar} introduces a deep association learning module to tackle data association challenges and uses loss sampling and ordinal loss to refine spatial relationships. The CRAFT~\cite{kim2023craft} model adapts radar-camera correlation in polar coordinates, utilizing cross-attention for robust fusion. FusionNet~\cite{lim2019radar} uses dense 2D radar maps and spatial transformation to align sensor feature maps. RISFNet~\cite{cheng2021robust} integrates multi-frame radar data with RGB images using temporal encoding and self-attention. Simple-BEV~\cite{harley2023simple} combines camera and radar features through rasterised radar information and bilinear sampling for 3D bird’s-eye view representation. CRN~\cite{kim2023crn} uses Radar-Assisted View Transformation (RVT) to align camera and radar features in BEV and applies multimodal deformable cross-attention (MDCA) for precise fusion. RCBEVDet~\cite{lin2024rcbevdet} dynamically aligns radar and camera BEV features with the Cross Attention Multi-Layer Fusion Module (CAMF). RCM-Fusion~\cite{kim2024rcm} combines radar and camera data at both the feature and instance layers, refining 3D schemes using radar points. CRKD~\cite{zhao2024crkd} proposes a cross-modal knowledge distillation framework for improving camera-radar fusion through BEV representations.

For Radar-Lidar sensor fusion, MVDNet~\cite{qian2021robust} introduces a two-stage deep fusion detector. It first generates proposals from the BEV space of Lidar and Radar sensors, then fuses regional features using attention mechanisms and temporal fusion. With advancements in 4D radar, its advantages over 3D radar have gained attention. 4D radar not only provides 3D positional data but also includes velocity measurements, improving dynamic object tracking and identification. Its robustness in adverse weather and high spatial resolution make it promising for autonomous driving and advanced driver assistance systems (ADAS). InterFusion~\cite{wang2022interfusion} employs self-attention to fuse Lidar point clouds with 4D radar features. Chae et al.~\cite{chae2024towards} introduced a weather-adaptive radar flow gating network that fuses 4D radar and Lidar features in the 3D domain.
Beyond LiDAR-Camera fusion, other sensor combinations like Radar-Camera and Radar-Lidar are also gaining traction. These approaches enhance environmental perception by leveraging complementary strengths, such as using radar's velocity data for improved dynamic object tracking and 4D radar's stability in harsh conditions. Innovative fusion techniques, including Sparse Depth Guidance (SDG), Laser Occupancy Guidance (LOG), and self-attention mechanisms, continue to drive progress in this field.

\subsection{Navigation and Localization}
\label{subsec:vision_navigation_localization}

\subsubsection{Embodied Navigation}

\begin{figure*}[ht]
    \centering
    \includegraphics[width=\linewidth]{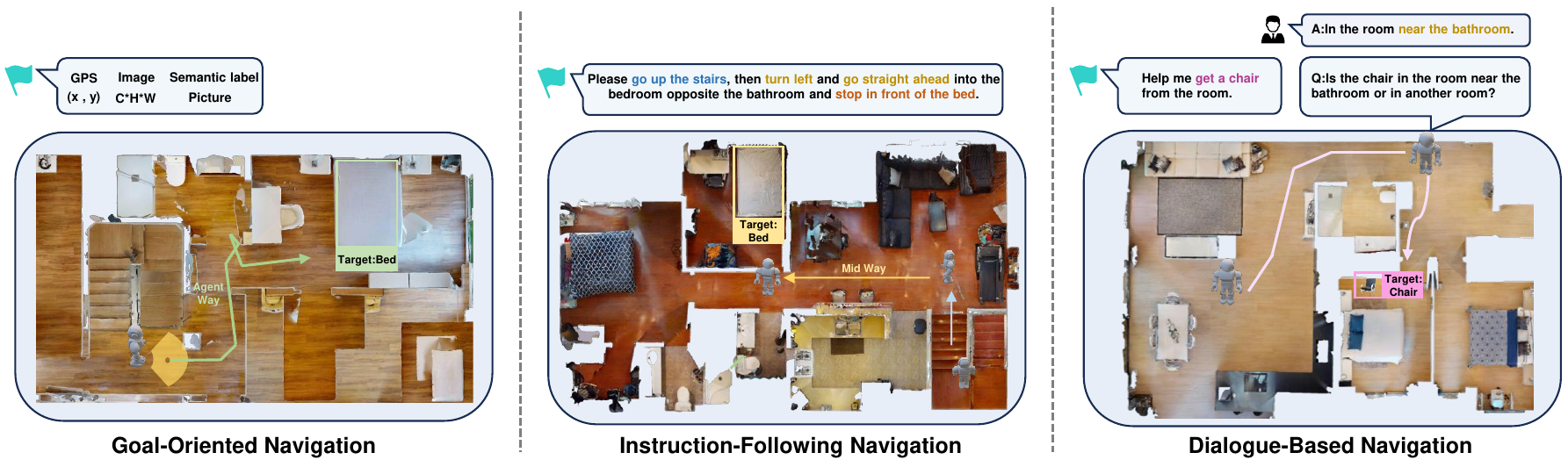}
    \caption{\textbf{Illustrations of goal-oriented, instruction-following, and dialogue-based navigation}.In goal-directed navigation, an agent utilizes multimodal inputs, such as visual images, semantic segmentation maps, and positional coordinates, to perceive the environment and locate the target object, thus enabling autonomous path planning without external instructions. Command-following navigation introduces natural language commands as guidance, requiring the agent to perform cross-modal semantic alignment for effective navigation. Dialogue-based navigation further incorporates interactive communication, allowing the agent to ask the user for disambiguation or instructions during execution. These examples reflect the evolution from perception-driven to language-guided and interaction-aware navigation, emphasizing the importance of multimodal fusion for making robust decisions in complex environments.}
    \label{fig:illustration_navigation}
\end{figure*}

Embodied Navigation relies on multimodal information (e.g., visual, tactile, and auditory inputs) to guide intelligences to act efficiently in dynamic and unstructured environments. As shown in Figure \ref{fig:illustration_navigation}, in recent years, the research in this field mainly focuses on the three directions of Goal-oriented Navigation, Instruction-following Navigation and Dialogue-Based Navigation. Goal-oriented Navigation, Instruction-following Navigation and Dialogue-Based Navigation. Goal-oriented Navigation combines visual semantics and spatial a priori to enable robots to autonomously navigate based on information about the objects in the environment, without relying on additional guidance from external advisors or collaborators. This type of navigation is the most basic requirement for AI intelligences, where the intelligence receives a goal location or object and plans a path by actively exploring the environment to efficiently reach the goal location. This capability has been extensively validated in Habitat's ObjectNav task ~\cite{savva2019habitat, zhu2017target, shah2021ving,ren2024infiniteworld}, demonstrating the effectiveness of achieving goal localisation in unknown environments. Instruction-following Navigation  relies on the development of Room-to-Room (R2R) and REVERIE tasks, which enhances natural language interaction and environment-specific reasoning, and further highlights the role of multimodal perception in complex scenarios ~\cite{ huang2019transferable, wu2024embodied,chen2024constraint,yan2024instrugen}.

In recent years, instruction-following navigation has made significant advancements. InstructNav\cite{long2024instructnav} employs a Dynamic Chain-of-Navigation (DCoN) framework, combined with Multi-sourced Value Maps, to achieve efficient zero-shot cross-task planning, enabling robots to directly generate executable trajectories from language instructions. It has demonstrated outstanding performance in R2R-CE and Habitat ObjectNav tasks. Similarly, NaVid\cite{zhang2024navid} leverages a video-driven Vision-Language Model (VLM), allowing robots to navigate using RGB video and human-issued instructions. It has achieved state-of-the-art (SOTA) performance on R2R and R2R-RxR tasks while exhibiting remarkable Sim2Real transferability. Meanwhile, ViNG\cite{shah2021ving} enables visual goal navigation by constructing a topological graph, allowing robots to operate without pre-built maps or explicit spatial priors. Additionally, DRL\cite{zhu2017target} applies reinforcement learning (RL) for goal-driven navigation, enabling robots to autonomously learn generalized navigation strategies and adapt quickly to diverse environments.

\begin{figure}[ht]
    \centering
    \includegraphics[width=\linewidth]{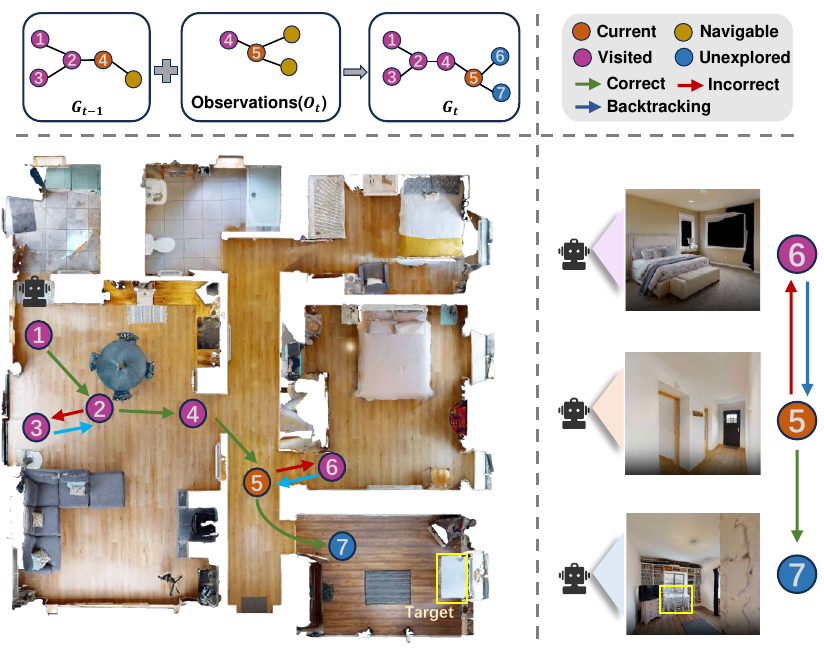}
    \caption{This figure illustrates the adaptive navigation process based on a topological map, where the agent continuously updates its environment model by integrating existing topological information with current observations, gradually expanding navigable areas. The construction of the topological map enables the agent to comprehend spatial structures and efficiently plan paths during exploration. The visual observations at different locations correspond to key nodes in the topological path, assisting the agent in environment understanding and goal matching. If the agent relies solely on a local action space, it may wander aimlessly, especially after a navigation error has occurred, making it difficult to effectively adjust the route. The topological map allows the agent to optimize navigation strategies globally, ensuring efficient exploration and dynamic path correction, thereby enhancing its autonomous adaptation in complex environments.}
    \label{fig:navigation}
\end{figure}

The construction of topological maps plays a crucial role in enhancing robotic navigation capabilities. As shown in Figure \ref{fig:navigation}, navigation agents can dynamically update environmental models, integrating current observations with existing topological information to progressively expand navigable regions. Compared to relying solely on local decision-making, incorporating topological maps allows robots to optimize navigation paths based on global information, effectively mitigating random exploration-induced disorientation. Furthermore, when navigation errors occur, robots can trace back historical trajectories and dynamically adjust paths. For instance, DUET~\cite{chen2022think} employs a dual-scale graph transformer, integrating local visual features with global spatial information during topological path construction to achieve cross-modal alignment and optimize navigation strategies. BEVBert~\cite{an2022bevbert} further enhances pre-training on hybrid topo-metric maps, enabling robots to better comprehend spatial structures and improve cross-modal reasoning capabilities in complex environments. Chasing Ghosts~\cite{anderson2019chasing} introduces Bayesian state tracking, integrating semantic spatial maps to enhance trajectory inference and goal localization, thereby improving navigation robustness in uncertain environments.Additionally, long-horizon planning is essential for embodied navigation. PREVALENT~\cite{hao2020towards} and PRET~\cite{lu2024pret} leverage Vision-Language Navigation (VLN) pre-training and directed fidelity trajectories to enhance alignment between visual and linguistic information, thereby optimizing path search efficiency. HAMT~\cite{chen2021history} introduces a History-Aware Multimodal Transformer (HAMT), explicitly encoding comprehensive navigation history during cross-modal decision-making, reinforcing navigation capabilities in long-horizon tasks. Meanwhile, TD-STP~\cite{zhao2022target} employs a Target-Driven Structured Transformer Planner (TD-STP), integrating topological maps to predict long-term navigation goals and modeling environmental topologies, ensuring efficient adaptation and optimized action planning for robotic navigation. 
PRET~\cite{lu2024pret} proposes a directionally faithful trajectory planning strategy that enhances planning efficiency and significantly reduces computational overhead by evaluating the semantic alignment between language instructions and candidate paths to select the most suitable goal node. It employs a direction-aware graph structure to avoid frequent updates to the entire map, making it particularly suitable for structured indoor environments. However, its deployment on real-world robotic platforms still faces multiple challenges. The dynamic nature of real environments, noisy sensory inputs, and reduced accuracy in trajectory-instruction alignment can all negatively affect navigation performance. Moreover, PRET requires continuous maintenance of global path information, imposing high demands on real-time perception synchronization and computational resources.
NavGPT-2~\cite{zhou2024navgpt}, on the other hand, integrates large-scale vision-language models (e.g., InstructBLIP) with a topological path planning module, adopting a decoupled design that separates language reasoning from navigation policy learning. This allows for both semantic interpretability and efficient decision-making. By freezing the large model and only training a lightweight policy network, NavGPT-2 achieves strong performance in instruction comprehension, path visualization, and human-robot interaction, demonstrating good data efficiency and generalization capability. However, it still relies on the pretrained model's visual-semantic alignment ability, which may degrade under real-world conditions such as low lighting, occlusion, or sensory noise. Furthermore, the model's long reasoning pipeline and complex prompt design present efficiency bottlenecks when deployed on resource-constrained embedded platforms.

Dialogue-based navigation enables intelligent agents to actively interact with an oracle or a human during navigation to acquire additional environmental information, clarify ambiguous instructions, or correct navigation errors, thereby enhancing adaptability and generalization. Unlike traditional static instruction-following navigation, this approach allows agents to dynamically adjust their paths while executing tasks rather than strictly following predefined instructions. For example, VLNA~\cite{nguyen2019vision} and HANNA~\cite{nguyen2019help} utilize fixed-format help signals, enabling agents to request sub-goal information from advisors within the environment, while CVDN~\cite{thomason2020vision} and DialFRED~\cite{gao2022dialfred} further incorporate open-ended natural language dialogues, allowing agents to proactively ask questions and receive more fine-grained navigation guidance. Through these methods, dialogue-based navigation enhances an agent’s ability to comprehend instructions, correct errors, and adapt to complex tasks in dynamic environments, driving embodied navigation toward more natural human-agent interaction.
These methods, such as InstructNav~\cite{long2024instructnav}, NaVid~\cite{zhang2024navid}, BEVBert~\cite{an2022bevbert}, and CVDN~\cite{thomason2020vision}, demonstrate the transformative potential of multimodal, fusion in embodied navigation. By integrating visual input, language processing, and spatial reasoning, robots can now follow natural language instructions, navigate toward visual goals, or perform goal-directed navigation tasks with greater flexibility and adaptability~\cite{zhang2024navid,liang2025structured,lin2023development,wang2024data,lin2025evolvenav}. The fusion of different sensory modalities such as vision, language, and spatial cues enables robots to navigate more robustly in diverse and dynamic environments, pushing the boundaries of autonomous navigation systems.

\subsubsection{Visual Localization}
Visual Localization plays a pivotal role in robot vision, enabling autonomous agents to accurately determine their position in complex and dynamic environments. Recent advancements in multimodal fusion, particularly deep-learning-based approaches, have significantly enhanced localization accuracy and robustness~\cite{chen2023deep}. These methods leverage complementary sensory inputs such as RGB images, depth data, and inertial measurements to address challenges like sensor noise, dynamic scenes, and varying lighting conditions~\cite{wang2023triple,wang2023attention,wang2022mtldesc}. Model,s like DeepVO~\cite{mohanty2016deepvo} and D3VO~\cite{yang2020d3vo} exemplify the integration of convolutional and recurrent neural networks, enabling end-to-end estimation of ego-motion and depth. Supervised methods excel in constrained settings by utilizing labeled datasets, while self-supervised approaches leverage geometric and photometric consistency to adapt effectively to unseen environments~\cite{zhou2017unsupervised}.

Hybrid frameworks combining classical geometric models with deep neural networks offer a promising solution to challenges like scale ambiguity and drift. By integrating learned depth and optical flow predictions~\cite{shi2023videoflow}, these systems achieve consistent pose estimations, improving monocular system accuracy. For example, D3VO incorporates uncertainty estimation into the visual odometry pipeline for robust performance in dynamic and occluded environments. This highlights the value of merging data-driven and model-based methods for reliable localization~\cite{xu2024local,wang2025focus}.
Recent advances in scene coordinate regression and neural implicit representations are transforming the field. Scene coordinate regression techniques eliminate the need for explicit 3D maps by learning direct transformations from image pixels to 3D coordinates, simplifying localization. Neural implicit representations, like NeRF~\cite{mildenhall2021nerf} and Semantic-NeRF~\cite{zhi2021place}, encode scene geometry and semantics into compact neural fields, enabling robust localization in new environments and showcasing the potential of multimodal fusion for bridging perception and mapping.
Despite progress, challenges remain, such as the need for large-scale datasets, high computational costs, and limited generalization to real-world conditions. Future research will likely focus on improving scalability, real-time capabilities, and developing unified frameworks that integrate localization with high-level perception and decision-making tasks~\cite{chen2025sage}.

\subsection{Simultaneous Localization and Mapping}

Simultaneous Localization and Mapping (SLAM) is a fundamental task in robot vision, enabling autonomous agents to accurately estimate their own pose while exploring unknown environments. Traditional SLAM systems include LiDAR-based\cite{article5,shan2018lego} and vision-based~\cite{Davison2007MonoSLAM,Engel2014LSDSLAMLD} methods. While these single-modal approaches have achieved significant progress, they are inherently limited by the constraints of their respective sensors. For instance, pure visual SLAM struggles in low-light or textureless environments, whereas LiDAR-based methods lack semantic understanding and suffer from data sparsity in long-range or low-reflectivity areas.

\begin{figure}[htbp]
    \centering

    \includegraphics[width=0.5\textwidth]{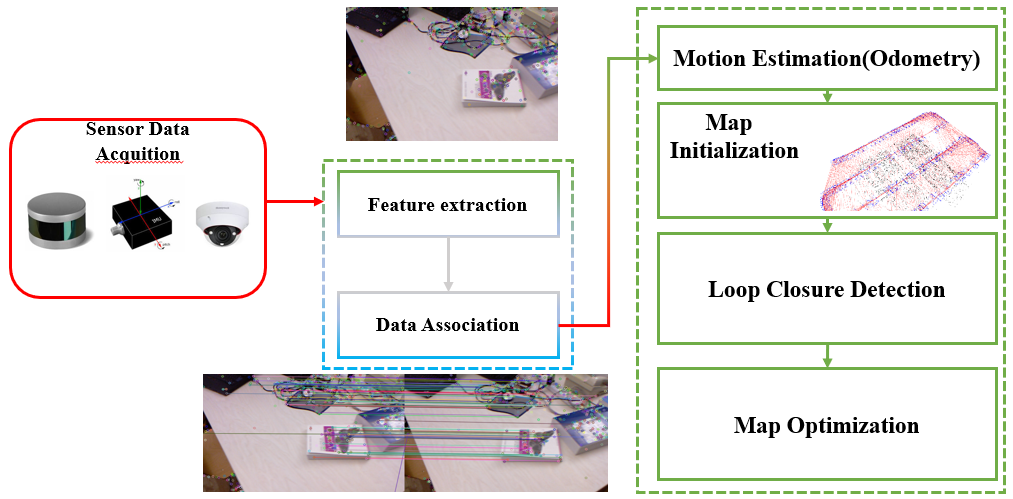}
    \caption{%
        A block diagram illustrating the SLAM pipeline: 
        from sensor data acquisition and feature extraction, 
        to motion estimation (odometry), loop closure detection, 
        and map optimization.
    }
    \label{fig:slam-pipeline}
\end{figure}

To overcome these limitations, multimodal fusion-based SLAM has emerged as a research hotspot. These methods integrate heterogeneous sensor data—such as cameras, LiDAR, inertial measurement units (IMU), GPS, and radar—to achieve enhanced environmental perception. LiDAR provides precise spatial structural information, vision offers rich semantic details, and IMU compensates for short-term motion estimation uncertainties. By combining these modalities, SLAM systems have significantly improved robustness, accuracy, and environmental adaptability.A generalized SLAM pipeline involves sensor data acquisition, feature extraction, data association, motion estimation (odometry), map initialization, loop closure detection, and map optimization, as illustrated in Figure~\ref{fig:slam-pipeline}. This structured pipeline ensures accurate localization and mapping, facilitating robust robot navigation in diverse environments.

Early multimodal fusion approaches, such as V-LOAM~\cite{zhang2015visual}, integrated visual odometry with LiDAR-based mapping, effectively leveraging the complementary nature of geometric and texture-based features. LIMO~\cite{graeter2018limo} further incorporated IMU data, utilizing sparse visual features to enhance system stability in dynamic environments. More advanced systems, such as LIC-Fusion~\cite{zuo2019lic} and its successor LIC-Fusion 2.0~\cite{zuo2020lic}, adopted tightly coupled optimization frameworks to deeply integrate LiDAR, IMU, and visual data, achieving a balance between high accuracy and real-time performance in large-scale environments.A comparison of recent multimodal SLAM methods and their corresponding sensor fusion strategies is summarized in Table~\ref{tab:multimodal_slam}. These methods integrate diverse modalities to enhance performance across various robotic applications.

In recent years, with the development of transformer-based deep learning techniques, some methods have introduced attention mechanisms to model temporal and cross-modal relationships. For example, UVIO~\cite{delama2023uvio} employs a transformer backbone to jointly learn from visual and IMU inputs, significantly improving pose estimation in dynamic environments. RD-VIO~\cite{li2024rd} takes this further by incorporating radar data, leveraging radar’s robustness under adverse weather conditions to enhance object tracking and localization. Additionally, MM3DGS SLAM~\cite{sun2024mm3dgs} integrates vision, depth, and IMU data, utilizing a 3D Gaussian mapping approach to improve real-time mapping quality and pose estimation accuracy. Meanwhile, ConceptFusion~\cite{jatavallabhula2023conceptfusion} introduces open-set multimodal 3D mapping, combining vision, language, and audio data to enable multimodal querying of the environment, thereby enhancing semantic understanding. Wang et al.~\cite{wang2023co} adopted hybrid neural representations and multi-resolution hash grids to achieve high-fidelity scene reconstruction, significantly improving accuracy and robustness in RGB-D SLAM tasks.

Furthermore, LVI-SAM~\cite{shan2021lvi} represents a landmark multimodal SLAM framework, utilizing factor graph optimization to deeply couple visual and LiDAR data, particularly excelling in low-texture and dynamic environments. In terms of vision-IMU-GPS fusion, VINS-Fusion has become a mainstream solution for large-scale, high-precision localization and is widely used in UAVs and mobile robots.

In summary, multimodal SLAM systems have evolved from loosely coupled architectures to deeply integrated frameworks, achieving stronger cross-modal alignment and environmental understanding. These approaches not only enhance system adaptability in complex environments but also pave the way for semantic SLAM, enabling robots to perceive and understand their surroundings at a higher cognitive level. Future research is expected to explore foundation model-driven SLAM, cross-modal self-supervised learning, and lightweight real-time fusion networks suitable for edge devices.

\begin{table}[h]
    \centering
    \caption{Representative multimodal SLAM methods and their modality configurations.}
    \label{tab:multimodal_slam}
    \begin{tabular}{p{3.5cm}p{4.5cm}}
        \hline
        \textbf{Method} & \textbf{Modality} \\
        \hline
        V-LOAM~\cite{zhang2015visual} & LiDAR + RGB \\
        LVI-SAM~\cite{shan2021lvi} & LiDAR + RGB \\
        LIC-Fusion~\cite{zuo2019lic} & LiDAR + RGB + IMU \\
        LIC-Fusion 2.0~\cite{zuo2020lic} & LiDAR + RGB + IMU \\
        VINS-Fusion~\cite{qin2018vins} & RGB + IMU + GPS \\
        UVIO~\cite{delama2023uvio} & RGB + IMU \\
        RD-VIO~\cite{li2024rd} & Radar + RGB + IMU \\
        MM3DGS SLAM~\cite{sun2024mm3dgs} & RGB + IMU + Depth \\
        ConceptFusion~\cite{jatavallabhula2023conceptfusion} & RGB + Audio \\
        Co-SLAM~\cite{wang2023co} & RGB + Depth \\
        \hline
    \end{tabular}
\end{table}

\subsection{Robot Manipulation}
\subsubsection{Vision-Language-Action Models}
Multimodal feature fusion is the core technology enabling the seamless transition from perception to action in robot manipulation tasks. Vision-Language-Action (VLA) models~\cite{xu2025a0}, which integrate visual perception, language understanding, and action planning, offer an efficient framework for addressing the challenges of complex manipulation tasks. The key innovation in VLA models lies in their ability to extract complementary features from different modalities and integrate them into a unified representation, enabling precise action planning and execution.

The visual modality is crucial for manipulation tasks, providing information about object position, shape, and layout. However, relying only on vision limits the robot's task understanding. To address this, VLA models integrate language to process natural language instructions, bridging gaps in visual perception. For example, the RT-2~\cite{brohan2023rt} model aligns visual and language representations via pretraining, allowing the robot to generate control commands from linguistic input using a Transformer architecture to map high-level semantics to actions. Additionally, the RoboMamba~\cite{liu2024robomamba} model incorporates action dynamics into the multimodal fusion framework, optimizing vision, language, and action for task planning. The 3D-VLA~\cite{zhen20243d} model enhances visual representation by generating 3D point clouds and depth maps, improving task precision and robustness with language-guided action generation in dynamic environments. OpenVLA~\cite{kim2024openvla} employs Low-Rank Adaptation (LoRA) to efficiently fine-tune VLMs, achieving cross-modal alignment with minimal trainable parameters. This design enables effective instruction following and generalization, particularly suited for deployment on resource-constrained robotic platforms. However, its focus on benchmark instruction tasks overlooks the continuous visual perception and fine-grained spatial reasoning needed in real-world robotics. Additionally, while LoRA ensures parameter efficiency, it may limit the model's expressiveness in complex multimodal scenarios, such as indoor manipulation or dynamic path planning, where detailed spatial understanding is critical. VoxPoser~\cite{huang2023voxposer} introduces a language-driven affordance reasoning framework that generates 3D value maps to guide complex manipulation. It excels in zero-shot generalization to novel objects and environments by linking natural language to 3D spatial actions. Yet, its reliance on voxel-based representations leads to high memory and compute costs, especially in large or high-resolution scenes. Moreover, the use of offline language models restricts real-time interaction and replanning, posing challenges for closed-loop control in dynamic robotic tasks.

In summary, multimodal feature fusion methods based on VLA models leverage the complementary strengths of visual, language, and action modalities to provide robots with robust perception and decision-making capabilities for manipulation tasks. These approaches not only significantly enhance the model’s ability to understand task semantics and plan actions but also open up new possibilities for deploying robots in dynamic and complex environments.
\subsubsection{Tactile-based Manipulation}

Vision-touch multimodal fusion is crucial for precision and stability in robotic grasping tasks. Vision provides global information on object position, posture, and geometry, while touch offers local feedback on force, surface properties, and slip detection. By combining these modalities, robots can overcome single modality limitations and adapt to dynamic environments. Recent research focuses on optimizing grasping stages such as generation, planning, stability assessment, and dynamic adjustment. During grasp generation, vision determines position and orientation, while tactile sensing provides surface details and force feedback for optimal grasp points. FusionNet-A~\cite{babadian2023fusion} combines tactile and visual features using 3D pooling and global average pooling for classification, improving object recognition and planning. Similarly, VITO-Transformer~\cite{li2023vito}, based on Vision Transformer (ViT), integrates tactile inputs through self-attention, enabling precise grasp point estimation and dynamic adaptability. In grasp planning, tactile feedback is crucial for path planning and posture adjustment. RotateIt~\cite{qi2023general} integrates visual, tactile, and proprioceptive data to optimize force closure in multi-axis rotation tasks. Sparsh~\cite{higuera2024sparsh} uses a universal tactile representation method pre-trained on a large tactile dataset, enhancing slip detection, grasp stability, and planning accuracy in dynamic environments.
In grasp stability assessment, multimodal fusion is crucial. Vision provides object position and posture, while tactile sensing offers real-time feedback on slip and force changes. LI et al.~\cite{li2024grasp} introduced a model using spatial and temporal attention, where spatial attention focuses on key object regions and temporal attention tracks dynamic grasp changes. This approach adjusts the weight of multimodal features based on sensor signal quality, improving grasp robustness and prediction accuracy in dynamic environments. Tactile feedback is also vital for grasp adjustments. For example, MimicTouch~\cite{yu2024mimictouch} uses a strategy learning framework that mimics human tactile-guided behaviors, collecting multimodal tactile data and combining self-supervised representation learning with reinforcement learning to optimize grasp stability. Octopi~\cite{yu2024octopi} uses a GelSight sensor to collect surface data, aligning tactile and visual information with the CLIP model to enhance grasping strategies by learning object properties. TAVI~\cite{guzey2024see} combines visual signals for spatial reasoning and tactile feedback for multi-finger tasks like flipping objects, enabling more precise robot operations.

In summary, the application of vision-touch multimodal fusion in robotic grasping tasks compensates for the shortcomings of single-modality perception and provides robust perception and decision-making capabilities for complex tasks. From improving the precision of grasp path planning to dynamically adjusting grasping forces, these studies highlight the immense potential of combining vision and touch. Future research will continue to focus on expanding the applicability of fusion models and improving their performance across diverse robotic platforms and complex scenarios.
\section{Evolution of Vision-Language Models}

The previous section provided a detailed overview of multimodal fusion strategies as applied to core robot vision tasks, such as semantic scene understanding, 3D object detection, SLAM, navigation, and manipulation. These fusion methods, grounded in traditional deep learning paradigms, lay the foundation for robust and efficient perception. However, recent advances in vision-language models (VLMs) introduce a new paradigm that enables robots to comprehend, reason, and act through natural language interfaces. In this section, we shift our focus to the evolution of vision-language models, examining their pretraining strategies, cross-modal alignment mechanisms, and emerging architectures that bridge visual and linguistic modalities for more generalizable robotic intelligence.

\label{sec:technological_evolution}

\subsection{Pre-Training Across Modalities}

Pre-training plays a crucial role in multimodal fusion. By pre-training on large-scale multimodal data, the model learns deep correlations between vision and language, significantly enhancing its ability to understand and represent information across different modalities~\cite{Dou2022Coarse-to-Fine,Du2022A}.This not only improves the model's performance on vision-language tasks (such as visual question answering and image captioning) but also enables knowledge transfer across modalities, reducing the reliance on large amounts of labeled data~\cite{Gan2022Vision-Language}.Additionally, pre-training facilitates joint learning across multiple modalities, strengthening the model's generalization ability on new tasks and data~\cite{Sun2023Fusion}.Therefore, pre-training is a key step in multimodal fusion technologies and plays an essential role in improving the performance of vision-language models.

Cross-modal pre-training is crucial for applications such as vision-language models because it effectively integrates and understands information from different modalities~\cite{Bakkali2022VLCDoC:}.Vision and language are fundamentally distinct forms of information, and cross-modal pre-training enables models to learn the deep correlations between them, making them more accurate in performing vision-language tasks~\cite{Zhang2023Multi-Task}.

The fundamental principle of cross-modal pre-training is to jointly train information from different modalities, enabling the model to learn the associations and complementarities between them. This is typically achieved by creating a shared embedding space where data from various modalities (such as images, text, audio, video, etc.) are projected into the same space, allowing the model to learn the relationships between these modalities~\cite{Xu2020Joint,Agrawal2020Tie}. During training, the model employs specialized encoders (e.g., convolutional neural networks for visual data, transformer architectures for text and audio data)~\cite{Zhang2024Dynamic,Zhang2023Bimodal} to extract features from different modalities, and then interacts and integrates these features through cross-modal alignment mechanisms~\cite{Alatkar2022CMOT,Qian2022Integrating}. By utilizing methods such as contrastive learning and self-supervised learning, the model can optimize the alignment between modalities, thereby enhancing the flow and interaction of information across them~\cite{Berg2024Multimodal}. This ultimately improves the performance of multimodal tasks. Such pre-training not only enables better handling of multimodal data but also enhances the model's overall performance in complex applications~\cite{Li2023Joint}.

\subsection{Cross-Modal Alignment and Representation Learning}

Cross-modal alignment and representation learning are crucial for vision-language models, enabling different modalities (e.g., vision, text, audio) to interact in a shared embedding space. Alignment helps models map relationships between modalities, such as images to text or images to audio. Representation learning extracts useful features from each modality for various tasks. This combination improves performance in tasks like image captioning, visual question answering, and speech recognition, advancing multimodal capabilities. Cross-modal alignment maps information from modalities, such as images and text, into a shared space to understand their relationships~\cite{kim2021vilt,xu20253d}. However, challenges arise from differences in structure, representation, and data features between modalities. Images are high-dimensional, while text consists of discrete symbols, making alignment difficult. Other challenges include scale differences and noise in large-scale unlabeled data, which can affect alignment accuracy~\cite{ni2021m3p}.

Common methods for representation learning include contrastive learning, self-supervised learning, and cross-modal generative models, as well as emerging frameworks such as diffusion-based captioning models that provide new perspectives on vision-language generation~\cite{daneshfar2024image}. Contrastive learning aims to minimize the distance between positive samples (same-category) and maximize it for negative samples (different-category). For cross-modal tasks, positive samples are visual and language pairs from the same context, while negative samples come from different contexts. The model is trained using a contrastive loss function to bring related modality pairs, like images and descriptions, closer in a shared embedding space while distancing unrelated pairs~\cite{Ferraro2021Enriched,Hu2022Unsupervised}. The popular method CLIP~\cite{radford2021learning} trains both vision and language models using contrastive learning, aligning images and text in a shared space for consistency. ALIGN~\cite{jia2021scaling} follows a similar approach, training cross-modal image and text data to enhance matching performance.

Self-supervised learning is a type of unsupervised learning where the model is trained by generating labels from the data itself. In cross-modal self-supervised learning, tasks such as image inpainting or text mask prediction are typically designed.~\cite{Alwassel2019Self-Supervised}The model learns representations by generating and inferring missing modality information. Self-supervised learning does not rely on large amounts of manually labeled data and can be trained using unlabeled data~\cite{Sayed2018Cross}.A recent mainstream approach is DINO~\cite{caron2021emerging}, a self-supervised method based on contrastive learning that does not rely on negative samples. Instead, it learns visual representations by maximizing the similarity between an image and its transformed versions. UniT~\cite{hu2021unit}, on the other hand, is a self-supervised learning framework for cross-modal tasks. It jointly pre-trains different modalities (e.g., vision and text) through self-supervised tasks, enabling the model to learn universal features.
The goal of cross-modal generative models is to generate one modality (e.g., an image) based on another modality (e.g., text). This process uses generative models (such as Variational Autoencoders or Generative Adversarial Networks) to map information from one modality into the space of another modality~\cite{Kang2024A}.During training,the model learns the joint distribution between the two modalities and optimizes the model through methods such as adversarial training or maximizing likelihood, enabling it to generate the target modality accurately~\cite{Zhang2020Multi-Pathway}.Recent mainstream methods include T2F (Text-to-Face), which uses Generative Adversarial Networks (GANs) to convert text descriptions into facial images, successfully combining text and image generation tasks. VQ-VAE-2~\cite{razavi2019generating} improves upon traditional VAE methods and excels in multimodal generation tasks, capable of generating high-quality images or other modality content. BLIP~\cite{li2022blip}, through a generative model, jointly trains images and text, enabling mutual generation and understanding between the two modalities, thus enhancing the performance of vision-language generation tasks.

In cross modal representation learning, methods such as contrastive learning, self-supervised learning, and cross-modal generative models each play an important role. Through different training mechanisms and strategies, they help models learn richer and deeper representations from multimodal data. These methods not only improve the alignment and interaction between different modalities but also enhance the overall performance of multimodal tasks, providing strong support for more accurate vision-language understanding and generation. As technology continues to evolve, new methods are emerging, driving the field of multimodal learning towards more efficient and complex directions.

\subsection{Transformer Variants and Large Vision-Language Models}
Since its introduction by Vaswani et al.~\cite{vaswani2017attention} in 2017, Transformer has become a core architecture in deep learning, achieving major advancements in natural language processing (NLP), computer vision (CV), and multimodal learning. Its key innovation, the self-attention mechanism, efficiently models long-range dependencies and improves training speed through parallel computation. Unlike traditional recurrent neural networks (RNNs), Transformer removes sequential computation constraints, enhancing performance in large-scale data tasks. With more computational power, Transformer has been scaled and optimized, becoming the foundation for large-scale AI systems.

Transformer was initially applied in NLP tasks, with BERT~\cite{kenton2019bert} and GPT~\cite{radford2018improving} as key models showcasing its advantages in natural language understanding (NLU) and natural language generation (NLG), respectively. Proposed by Google in 2018, BERT uses a Transformer encoder-only architecture and pretrains with masked language modeling (MLM) and next sentence prediction (NSP), improving text understanding tasks. In contrast, OpenAI’s GPT employs a decoder-only autoregressive approach, laying the foundation for generative AI. These early models helped establish large-scale pre-trained language models as a standard for NLP tasks.

As computational power increased, the scale of Transformer models rapidly expanded to enhance their generalization ability across multiple tasks and complex reasoning. GPT-2~\cite{radford2019language} increased the parameter count to 1.5 billion, showcasing zero-shot and few-shot learning capabilities and establishing a foundation for general-purpose text generation. GPT-3~\cite{brown2020language} further scaled up to 175 billion parameters and introduced in-context learning (ICL), allowing models to adapt to new tasks with minimal fine-tuning. However, as Transformer models continued to grow in size, computational complexity became a pressing issue. Due to the quadratic complexity $O(N^2)$ of the self-attention mechanism, processing long-sequence inputs led to an exponential increase in computational cost. To address this, researchers introduced a series of Efficient Transformer variants to optimize resource usage. Reformer~\cite{kitaev2020reformer} employed locality-sensitive hashing (LSH Attention) to reduce memory requirements for attention computation, while Longformer~\cite{beltagy2020longformer} and Linformer~\cite{wang2020linformer} utilized sparse attention and low-rank projections, respectively, to decrease computational complexity, making Transformer more efficient for handling long-text sequences.

Following its success in NLP, Transformer was rapidly adopted in computer vision, challenging the dominance of convolutional neural networks (CNNs) in image representation learning. Vision Transformer (ViT)~\cite{dosovitskiy2020image} demonstrated that Transformer architectures could be directly applied to image classification tasks, achieving performance surpassing CNNs on large-scale datasets. ViT~\cite{dosovitskiy2020image} introduced patch embedding, dividing images into small patches and converting them into tokens for input into the Transformer, enabling effective feature extraction through self-attention. Subsequently, Swin Transformer~\cite{liu2021swin} introduced shifted window attention, significantly improving computational efficiency and performance in dense prediction tasks such as object detection and semantic segmentation. Additionally, DETR~\cite{carion2020end} leveraged an end-to-end Transformer-based approach to replace traditional region proposal networks (RPNs) in object detection, highlighting Transformer’s potential in vision tasks. These studies demonstrated that Transformer could serve as a general-purpose deep learning architecture applicable to both text and visual modalities.

As models scaled, the Mixture of Experts (MoE) architecture emerged to optimize computational resource usage. MoE dynamically activates a subset of expert networks, allowing models to scale parameters without significantly increasing computational cost, improving efficiency while maintaining strong generalization. The core idea is to activate only a fraction of the model parameters during inference, preventing linear growth in computational cost. The Switch Transformer~\cite{fedus2022switch} was one of the first MoE-based Transformers, activating experts selectively in each layer, reducing computational overhead while outperforming dense Transformers in NLP tasks. GLaM~\cite{du2022glam} refined MoE by introducing adaptive expert selection, dynamically routing inputs to the most relevant experts, balancing efficiency and performance. MoE has been integrated into ultra-large-scale models like GPT-4~\cite{achiam2023gpt} and DeepSeek-MoE~\cite{dai2024deepseekmoe}, advancing multi-task and multimodal applications. GPT-4~\cite{achiam2023gpt} incorporates MoE-style routing, optimizing efficiency and reducing inference costs while supporting multimodal inputs. DeepSeek-MoE~\cite{dai2024deepseekmoe}, using 64 experts and dynamic activation, drastically reduces computational redundancy while maintaining high performance, and improves expert scheduling for better generalization in multilingual and multi-task settings. MoE enables Transformers to balance parameter expansion with computational efficiency, addressing the need for better generalization and complex reasoning without increasing computational burden. MoE is expected to play a pivotal role in future Transformer research, driving efficient, large-scale, general-purpose AI development.

\begin{figure*}[ht]
    \centering
    \includegraphics[width=\linewidth]{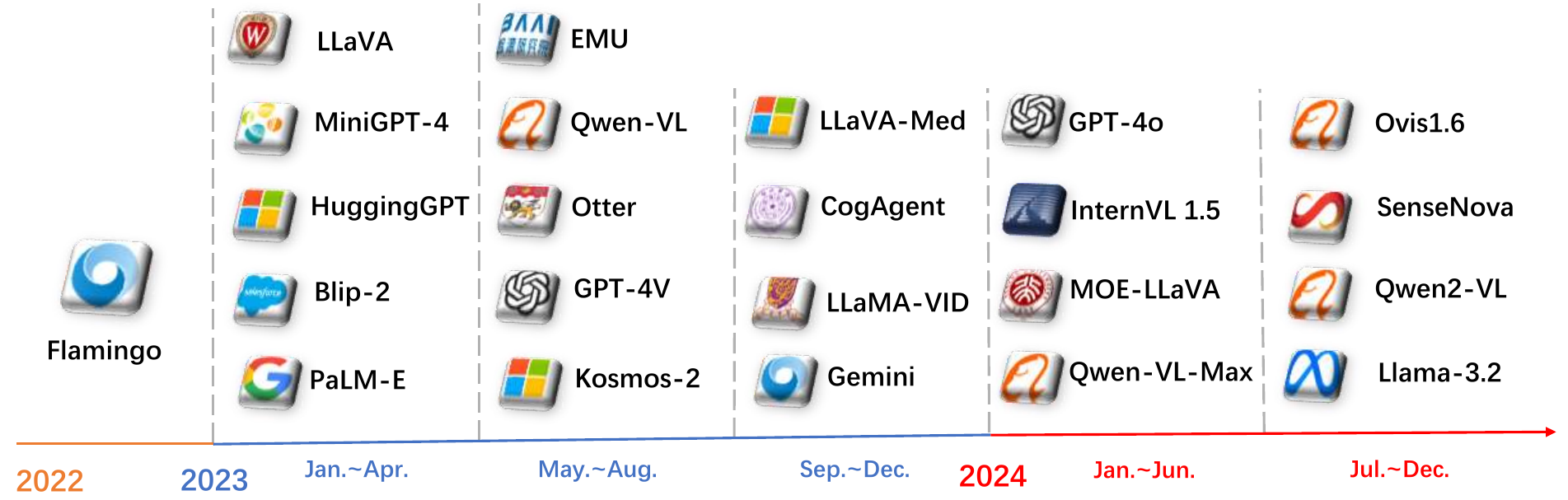}
    \caption{The timeline illustrates the development of various multimodal vision-language models from 2022 to 2024. The models are categorized according to their release dates: early 2022 (e.g., Flamingo~\cite{alayrac2022flamingo}, PaLM-E~\cite{driess2023palm}), 2023 (e.g., MiniGPT-4~\cite{zhu2023minigpt}, Qwen-VL~\cite{bai2023qwen}, GPT-4V~\cite{yang2023dawn}, Gemini), and 2024 (e.g., MoE-LLaVA~\cite{lin2024moe}, Llama-3.2~\cite{dubey2024llama}). These models represent the evolution of vision-language systems, with advancements in visual processing, decision-making, and multimodal integration across time. The diagram also highlights the ongoing trend of integrating larger and more sophisticated models for multimodal tasks.}
    \label{fig:VLMs}
\end{figure*}
With the widespread application of visual-language models (VLMs) in multimodal tasks, these models have played a crucial role in advancing the deep integration of computer vision and natural language processing. Figure \ref{fig:VLMs} illustrates the development process and technical evolution of different VLMs in visual and language understanding tasks since 2022. The figure showcases the release progress of various models, reflecting the continuous advancements in reasoning, generation, and multimodal understanding from the early models such as Flamingo~\cite{alayrac2022flamingo} and PaLM-E~\cite{driess2023palm} to later models like Qwen-VL~\cite{bai2023qwen} and Llama-3.2~\cite{dubey2024llama}.

Visual-language pretraining is the foundational technique for VLMs, allowing models to learn deep relationships between images and text through the joint training of large-scale image-text pairs, which improves the model's performance on cross-modal tasks. Flamingo~\cite{alayrac2022flamingo} is a classic example, combining few-shot learning and cross-modal reasoning to quickly adapt to new tasks with minimal examples, especially excelling in image captioning and visual question answering (VQA). Gemini~\cite{team2023gemini} further extends this approach by jointly training large-scale data from images, text, and audio, enhancing the model's cross-modal reasoning ability, particularly excelling in visual reasoning tasks. Kosmos-2~\cite{peng2023kosmos} strengthens the joint modeling of images and language through self-supervised learning, allowing it to handle more complex tasks, such as deep reasoning and generation of image and text. Additionally, PaLM-E~\cite{driess2023palm} combines improvements in multimodal learning and reasoning capabilities, enabling it to process multimodal data such as images, text, and audio, performing well in cross-modal reasoning and generation tasks.

The core innovation of these models lies in joint training, enhancing the synergy between images and text, making them more widely applicable in multimodal tasks. MiniGPT-4~\cite{zhu2023minigpt} and LLaVA~\cite{liu2024visual} further refine this by aligning visual-language representations, improving the accuracy and flexibility of image-to-text generation. LLaVA~\cite{liu2024visual} (Language and Vision Assistant) is an innovative visual-language model that employs instruction tuning, giving the model a high degree of flexibility in handling various tasks. LLaVA~\cite{liu2024visual}'s ability to adjust to different tasks using natural language instructions enables it to not only respond rapidly to traditional visual-language tasks like VQA and image captioning, but also to adjust its execution strategy dynamically for new tasks. Through instruction tuning, LLaVA~\cite{liu2024visual} can generate precise image descriptions or answer visual-related questions based on different task demands. This flexible task-switching capability makes LLaVA~\cite{liu2024visual} particularly advantageous in complex environments, excelling in tasks such as cross-modal and visual reasoning.

Llama~\cite{dubey2024llama}, developed by Meta, is a large-scale language model series designed to enhance language understanding and generation, with applications in visual-language tasks, particularly cross-modal reasoning. Llama~\cite{dubey2024llama} uses large-scale pretraining to boost text understanding and generation, improving performance in tasks like visual reasoning and image-to-text generation when combined with visual data. Its strong reasoning abilities, especially in commonsense reasoning and text generation, make it effective for complex cross-modal tasks. As task complexity grows, reasoning and computational efficiency are vital for VLM development. To improve computational efficiency and real-time reasoning on complex tasks, LLaVA~\cite{liu2024visual} introduces instruction tuning, enhancing the model's adaptability across various tasks like VQA and image captioning. Instruction tuning allows LLaVA~\cite{liu2024visual} to adjust its reasoning strategy based on natural language instructions, enabling quick adaptation to new multimodal tasks and boosting reasoning efficiency.

MoE-LLaVA~\cite{lin2024moe} uses a Mixture of Experts (MoE) architecture, activating only the relevant expert networks to reduce computational burden and improve reasoning efficiency, especially on large datasets. Qwen-VL~\cite{bai2023qwen} enhances computational efficiency with Multimodal Rotary Positional Embedding (M-RoPE) and dynamic resolution, improving reasoning speed and accuracy in high-resolution image processing by better understanding spatial relationships. InternVL~\cite{chen2024internvl} further optimizes efficiency through expert scheduling and distributed computing, ensuring dynamic resource allocation for large-scale multimodal data. Ovis1.6~\cite{lu2024ovis} balances computational efficiency and reasoning accuracy, making it ideal for real-time visual-language tasks. These advancements allow VLMs to handle complex scenarios in fields like autonomous driving, medical image analysis, and robotics. By improving both reasoning and computational efficiency, VLMs tackle complex cross-modal tasks, enhancing decision-making across various domains. As VLMs evolve, they will play a crucial role in more complex multimodal tasks, providing efficient and accurate solutions for intelligent decision systems.

In recent years, a new generation of VLMs has emerged, such as GPT-4V~\cite{yang2023dawn}, BLIP-2~\cite{li2023blip}, and LLaVA-3D~\cite{zhu2024llava}, demonstrating exceptional capabilities in open-vocabulary understanding, multimodal reasoning, and instruction following. However, their applicability in robot vision tasks remains constrained by several practical challenges. First, while these models excel in static image-text tasks, their limited generalization ability on robot-specific perceptual inputs (e.g., depth maps, 3D point clouds, and temporal data) stems from insufficient training during the pre-training phase. This limitation undermines their performance in tasks such as SLAM, dynamic scene understanding, and precise manipulation. Second, these models are typically large in size and rely on dense self-attention mechanisms, leading to high inference latency and memory consumption, making them difficult to deploy on resource-constrained platforms commonly used in robotics, such as embedded processors and edge devices. Additionally, these models often lack mechanisms for continuous adaptation in dynamic environments, a capability that is critical for long-term tasks such as autonomous navigation and interactive operations. To address these challenges, progress is needed in areas such as multimodal pre-training (covering 3D and temporal modalities), model compression (e.g., pruning, quantization, and adapter-based fine-tuning), and task-specific online learning strategies to achieve real-time, robust robot perception and decision-making capabilities.

\subsection{Multimodal Instruction Following and Learning}
Instruction tuning serves as a crucial technique for improving the generalization ability of large-scale pretrained models. Its core idea is to guide models in learning different task modes through explicit task instructions, enabling efficient transfer and generalization across multiple tasks. In multimodal tasks, the implementation of instruction tuning typically involves key steps such as instruction dataset construction, supervised fine-tuning (SFT), reinforcement learning with human feedback (RLHF), and multimodal adaptation.

Instruction Dataset Construction forms the foundation of instruction tuning. Researchers construct datasets comprising diverse task instructions to help models learn how to comprehend and execute different task directives. These datasets include task inputs, expected outputs, and additional task descriptions to enhance task understanding. For instance, the FLAN dataset~\cite{wei2021finetuned} covers thousands of task instructions across multiple domains, allowing models to generalize effectively. Similarly, T0 ~\cite{sanh2021multitask} trains language models using a large-scale instruction dataset, endowing them with zero-shot and few-shot learning capabilities. In multimodal tasks, datasets like PaLI-X ~\cite{chen2023pali} integrate multilingual and multimodal instructions, improving transfer learning in vision-language applications.

Supervised Fine-Tuning (SFT) is the core step of instruction tuning, leveraging the constructed instruction dataset to fine-tune models for precise task execution. Typically, the pretrain-finetune paradigm is adopted, where large-scale pretrained models are fine-tuned using supervised tasks. Models such as GPT-4~\cite{achiam2023gpt}, PaLM~\cite{chowdhery2023palm}, and BLIP-2 ~\cite{li2023blip} employ extensive supervised fine-tuning strategies to enhance their ability to understand and execute various task instructions. In multimodal tasks, this process involves joint optimization of vision encoders and text decoders. For example, BLIP-2 employs the Query Transformer to optimize vision-language information extraction and adapts to instructions for image captioning and VQA.
Reinforcement Learning with Human Feedback (RLHF) improves instruction tuning by aligning models with human preferences for complex real-world tasks. RLHF involves collecting human preference data, training a reward model, and optimizing language models with reinforcement learning. Models like GPT-4V use RLHF to improve controllability and reliability in VQA tasks, better aligning with user intent. LLaVA~\cite{liu2024visual} also uses similar strategies to generate more natural, user-friendly responses.

Multimodal Adaptation extends instruction tuning to tasks involving multiple modalities such as images, text, audio, and video. PaLI-X~\cite{chen2023pali} uses large-scale multimodal instruction training to improve vision-language task generalization, while Flamingo~\cite{alayrac2022flamingo} integrates few-shot learning for rapid adaptation with minimal examples.
Instruction tuning enhances model controllability and adaptability across various multimodal tasks. In image captioning, models adjust style and detail based on the task, as seen in BLIP-2, which aids in medical imaging and social media captioning. PaLI-X~\cite{chen2023pali} improves cross-language image description, helping in document processing and translation. In VQA, models like GPT-4V provide accurate responses in diverse domains, and LLaVA~\cite{liu2024visual} enhances visual understanding for detailed answers. Flamingo’s few-shot learning allows AI assistants to generate personalized responses.

In visual reasoning, instruction tuning strengthens complex logical reasoning capabilities, including causal, spatial, and attribute reasoning. Models must infer relationships based on image content. PaLM-E~\cite{driess2023palm} integrates language, vision, and robotic control signals, optimizing reasoning performance in physical-world tasks. In robotic navigation, PaLM-E~\cite{driess2023palm} predicts optimal paths, avoids obstacles, and provides action recommendations. DeepMind’s Flamingo further enhances causal reasoning, answering inference chain questions like “Why did this object fall?” or “What could have caused the lights to go out?”

Instruction tuning plays a pivotal role in optimizing large-scale multimodal models, significantly improving their adaptability and generalization across multiple tasks and domains. Through high-quality instruction datasets, supervised fine-tuning, RLHF, and multimodal adaptation, instruction tuning enhances model performance in image captioning, VQA, and visual reasoning. As multimodal AI models continue to evolve, instruction tuning will further drive improvements in task controllability, cross-modal comprehension, and human-AI interaction, laying a robust foundation for the widespread application of multimodal AI.
\subsection{Context-Aware Multimodal Learning}

Context-aware modeling is a cutting-edge approach in multimodal learning, designed to boost the representational power and prediction accuracy of models by merging data from various modalities. The key idea behind this method is to include contextual information during the training process, which allows the model to better grasp the relationships between data points and their underlying context. By being context-aware, the model becomes more adaptable to diverse data distributions and real-world scenarios, enhancing its ability to generalize.

Context-aware modeling generally consists of several stages, including feature extraction from input data, defining and integrating contextual information, and making decisions based on context. From a technical perspective, this involves constructing multimodal feature representations using deep learning models and designing dynamic weight adjustment mechanisms that can adapt to changes in context. The challenge of context-awareness lies in effectively capturing and leveraging contextual information, which often demands intricate model architectures and sophisticated algorithmic approaches.

Context-aware modeling is widely used across various domains, including natural language processing~\cite{Chen2024CaMMLCM}, computer vision~\cite{Prakash2023AdaptFuseNetCM}, and recommendation systems~\cite{Malikireddy2024RevolutionizingPR}, where it can substantially boost both system performance and user experience. For instance, in text classification tasks, context-awareness helps the model capture the underlying meaning of sentences~\cite{Lan2019ALBERTAL}, while in image recognition, it improves the model’s ability to identify and understand complex scenes. The versatility of these applications demands that context-aware modeling approaches be highly scalable and adaptable.

Context-aware modeling faces several challenges, such as handling dynamically changing contexts~\cite{Akbik2018ContextualSE}, balancing the significance of different modality data~\cite{Tsai2019MultimodalTF}, and preventing overfitting. Solutions to these challenges involve using adaptive model architectures, incorporating attention mechanisms, and leveraging techniques like transfer learning to enhance the robustness and generalization of the model. Furthermore, extensive data labeling and model validation can help refine the performance of context-aware modeling systems.

At present, context-aware modeling research is becoming more refined, focusing on better extracting and utilizing context information, such as using GNNs to model complex relational networks~\cite{Dar2024ASC}. Researchers are also investigating how to integrate context-aware modeling with techniques like reinforcement learning~\cite{Ahn2024TuningLM} and unsupervised learning to create smarter and more adaptive decision-making processes~\cite{Vacareanu2020AnUM}. With advancements in computational capabilities and the growth in data, context-aware modeling is likely to make significant breakthroughs in various fields.
\subsection{Lightweight and Real-Time Models}

In recent years, the rise of the Internet of Everything and the growth of wireless networks have led to a surge in edge devices and data. IDC predicts that by 2025, global data volume will exceed 180 zettabytes (ZB), with over 70\% of IoT data processed at the network edge~\cite{Zwolenski}. This growth has made traditional cloud computing models inadequate for handling edge-generated data. Conventional cloud computing faces issues like poor real-time capabilities, bandwidth limitations, high energy consumption, and data security concerns. To address these, edge computing has emerged to process data directly at the network edge~\cite{Shi2016EdgeCV}, handling both downstream cloud data and upstream IoT data.
The increasing need for real-time inference in resource-constrained environments has driven the development of lightweight multimodal models. These models, balancing efficiency and accuracy, are ideal for mobile devices, embedded systems, and edge computing. By combining various data types like images, text, and audio, multimodal models have shown great potential in fields like VQA, image captioning, and speech recognition. However, designing and deploying these models efficiently on limited hardware remains a challenge.

In the context of multimodal learning, efficient vision models are crucial. EfficientNet\cite{Tan2019EfficientNetRM} has emerged as one of the most promising solutions for reducing the computational burden of vision models. By proposing a new method for scaling the depth, width, and resolution of CNNs, EfficientNet achieves state-of-the-art performance with fewer parameters and operations compared to traditional networks. This approach, which can be seamlessly integrated into multimodal tasks, ensures that vision-based components in a multimodal system can run efficiently even on resource-constrained devices.

Recent works have proposed several strategies to achieve this goal. For instance, MobileViT~\cite{mehta2021mobilevit} introduces a novel hybrid architecture that leverages the strengths of both convolutional neural networks (CNNs) and vision transformers (ViTs), effectively reducing computational costs while maintaining high accuracy. On the other hand, lightweight natural language processing (NLP) models such as TinyBERT\cite{Jiao2019TinyBERTDB} focus on reducing the size of transformer-based models via knowledge distillation, making them suitable for mobile deployment.

Autonomous vehicles are another example of edge computing applications, with ADAS enabling autonomous driving. Traditional cloud models capture general driving behaviors but fail to consider individual driver differences or environmental factors. To address this, Zhang et al. proposed a cloud-edge collaborative system~\cite{Zhang123}, which uses data like vehicle position, speed, and acceleration to model driving behavior and detect abnormal actions like speeding or drunk driving.
Connected vehicles enhance edge computing further. Lu et al. introduced CLONE~\cite{Lu2019CollaborativeLO}, a framework for cooperative learning across edge devices, using real-world data from an electric vehicle company to focus on battery and component failures.
Smart homes also illustrate edge intelligence, where real-time detection of appliance states is crucial for understanding user behavior and optimizing energy use. Zhang et al. proposed a non-intrusive appliance state detection system~\cite{Zhang2017IEHouseAN} that uses a single sensor to monitor multiple appliances via a cloud-edge-device collaborative framework.

In edge AI for Real-time fusion scenarios, development is predominantly built on existing cloud computing frameworks. However, these frameworks often fall short in edge-specific scenarios and struggle to accommodate the diverse and heterogeneous nature of cloud-edge-device systems. This lack of standardization hampers development efficiency and limits the potential benefits of collaboration. As a result, there is a pressing need for a universal and cross-platform computational framework that can address these challenges.
\section{Datasets and Benchmarks for Multimodal Fusion}
\label{sec:Datasets_for_Multimodal_Fusion}
The availability and diversity of multimodal datasets have played a foundational role in advancing robot vision across a broad spectrum of tasks, including scene understanding, navigation, and manipulation. This section provides a comprehensive overview of benchmark datasets that support multimodal fusion and vision-language modeling in robotic scenarios. These datasets encompass a range of sensor modalities (e.g., RGB, depth, LiDAR, radar, audio, text, tactile) and cover both indoor and outdoor environments.

As shown in Table \ref{tab:scene_understanding_datasets}, scene understanding datasets such as ScanQA, NuScenes, and SemanticKITTI are designed to evaluate spatial perception and semantic segmentation under different modality conditions. For manipulation tasks, Table \ref{tab:manipulation_datasets} summarizes representative datasets integrating visual, linguistic, and tactile modalities to support grasping, planning, and fine-grained control in human-object interaction contexts.

These datasets not only serve as training resources but also form the basis for the performance benchmarks discussed in Section \ref{sec:performance_evaluation}. The modality combinations and task types presented here directly impact model performance and generalization, which are further analyzed in the following section.

\subsection{Scene Understanding Datasets}
Multimodal datasets play a critical role in the field of semantic scene understanding. Table \ref{tab:scene_understanding_datasets} presents representative datasets that have been widely used in recent years. These datasets cover a wide range of scene types, from indoor environments to urban streetscapes, and integrate multimodal data such as RGB images, depth information, LiDAR, radar, and audio. They provide essential support for multi-task learning and cross-domain research.

In the domain of autonomous driving, nuScenes\cite{caesar2020nuscenes} and the Waymo Open Dataset\cite{sun2020scalability} are two iconic datasets. nuScenes~\cite{caesar2020nuscenes} offers data from six cameras, five radars, and one LiDAR, covering 360$^\circ$ views across 1,000 driving scenes, with more than 1.4 million frames annotated for 3D detection and tracking tasks across 23 categories. It captures diverse weather and lighting conditions in urban environments, but its limited geographic scope and relatively low vehicle density can affect the generalization performance in more complex or global traffic scenarios. The Waymo~\cite{sun2020scalability} dataset leads in terms of geographic coverage and data scale, containing 12 million 3D bounding box annotations and supporting cross-domain generalization and multimodal sensor fusion research. With synchronized high-resolution data from five LiDARs and five cameras, it supports comprehensive research in cross-domain generalization and multimodal sensor fusion. Its large volume, diverse locations, and rich annotations make it particularly suitable for real-world deployment and performance evaluation under varied driving conditions. Additionally, SemanticKITTI~\cite{behley2019semantickitti} focuses on semantic segmentation of LiDAR point clouds and provides an important benchmark for sequence-level semantic scene completion, with 43,000 densely annotated LiDAR scans. While it is valuable for understanding temporal consistency in point clouds, its reliance on a single sensor modality and limited environmental variability somewhat constrain its generalization to complex, sensor-rich real-world settings. The Cityscapes~\cite{cordts2016cityscapes} dataset, with its high-resolution pixel-level annotations and complex urban street scene data, has become a crucial resource for urban semantic understanding. However, the dataset is limited to monocular RGB images captured from a fixed onboard camera, and lacks 3D information and sensor diversity, making it less suitable for tasks that require depth reasoning, multimodal fusion, or adaptation to dynamic urban scenes.

For indoor scene understanding, datasets such as Matterport3D\cite{chang2017matterport3d}, ScanNet~\cite{dai2017scannet}, NYU Depth V2~\cite{silberman2012indoor}, and Hypersim~\cite{roberts2021hypersim} have been pivotal. Matterport3D~\cite{chang2017matterport3d} provides 10,800 panoramic RGB-D views, along with accurate 3D reconstructions and semantic annotations, making it widely applicable for semantic segmentation, 3D reconstruction, and multi-task learning. ScanNet~\cite{dai2017scannet}, through its 2.5 million RGB-D frames across 1,513 scenes, utilizes crowdsourced annotations to provide a large-scale, high-quality dataset for semantic segmentation and 3D modeling. NYU Depth V2~\cite{silberman2012indoor}, as one of the earliest RGB-D datasets, offers 1,449 images from 464 diverse indoor scenes with detailed semantic segmentation and object relationship annotations, making it instrumental for research on physical support reasoning and cluttered environments. In addition, the Hypersim dataset~\cite{roberts2021hypersim}, utilizing high-fidelity synthetic techniques, generates 77,400 indoor scene images with detailed pixel-level annotations, light decomposition, and surface normals, serving as an important tool for simulation-to-reality (Sim-to-Real) transfer research.However, despite offering valuable resources for visual perception tasks, these datasets still exhibit limitations in generalization to real-world robotic systems. Hypersim, being synthetic, lacks real sensor noise and physical interactions, leading to domain gaps in deployment. ScanNet and Matterport3D, while captured from real environments, are largely static and unimodal, making them less suitable for dynamic and interactive tasks. NYU Depth V2, though historically influential, has a relatively small scale and limited scene diversity, which constrains its applicability to modern multimodal and high-complexity scenarios.

In recent years, datasets focusing on cross-modal and interactive scenarios have gained increasing attention, further driving advancements in multimodal semantic understanding. For instance, 360+x~\cite{chen2024360+}  combines 360° panoramic views, first-person, and third-person perspectives while integrating modalities such as audio and textual descriptions, providing a rich experimental foundation for multi-view scene understanding and multimodal fusion research. ScanQA~\cite{azuma2022scanqa}, on the other hand, takes a step forward in combining natural language question answering with 3D scene understanding. It explores object localization and scene reasoning guided by language through question-answering tasks that integrate linguistic and visual depth.

In summary, these multimodal datasets span a wide range of applications, from urban traffic to indoor scenes, and from single-task to multi-task learning, showcasing the importance of modality complementarity in semantic scene understanding. By integrating visual, depth, audio, and linguistic data, these datasets provide abundant resources for comprehensive model training, evaluation, and optimization, while laying a solid foundation for future research in multimodal fusion and cross-domain transfer learning.

\begin{table*}[ht]
    \centering
    \renewcommand{\arraystretch}{1.2}
    \small
    \caption{Representative datasets for multimodal semantic scene understanding.}
    \label{tab:scene_understanding_datasets}
    \begin{tabular}{p{3cm}p{2.5cm}p{2.5cm}p{5.5cm}p{1cm}p{1cm}}
        \hline
        \textbf{Datasets} & \textbf{Scene} & \textbf{Multimodal Data}& \textbf{Data Scale} & \textbf{Venue} & \textbf{Year} \\
        \hline
        360+x~\cite{chen2024360+} & Indoor, Outdoor & Video/Audio & 28 scenes, 1,688 videos & CVPR & 2024 \\
        \hline
        ScanQA~\cite{azuma2022scanqa} & Indoor & RGB/Text & 41,363 questions, 58,191 answers & CVPR & 2022 \\
        \hline
        Hypersim~\cite{roberts2021hypersim} & Indoor & RGB/Depth & 77,400 images & ICCV & 2021 \\
        \hline
        NuScenes~\cite{caesar2020nuscenes} & Urban street & RGB/Lidar/Radar & 1.4 million frames annotated & CVPR & 2020 \\
        \hline
        Waymo~\cite{sun2020scalability} & Outdoor & RGB/Lidar & 12 million camera box annotations & CVPR &  2020 \\
        \hline
        Semantickitti~\cite{behley2019semantickitti} & Urban street & RGB/Lidar & 43,000 scans & ICCV & 2019 \\
        \hline
        Matterport3D~\cite{chang2017matterport3d} & Indoor & RGB/Depth & 10,800 panoramic RGB-D views & arxiv & 2017 \\
        \hline
        ScanNet~\cite{dai2017scannet} & Indoor & RGB/Depth & 1,513 scenes , 2.5 million RGB-D frames & CVPR & 2017 \\
        \hline
        Cityscapes~\cite{cordts2016cityscapes} & Urban street & 25,000 images & RGB/Depth & CVPR & 2016 \\
        \hline
        NYUDv2~\cite{silberman2012indoor} & Indoor & RGB/Depth & 464 scenes , 1449 RGBD images & ECCV & 2012 \\
        \hline
    \end{tabular}
\end{table*}

\subsection{Robot Manipulation Datasets}
\begin{table*}[ht]
    \centering
    \renewcommand{\arraystretch}{1.2} 
    \small
    \caption{Summary of representative multimodal datasets for robot manipulation tasks.}
    \label{tab:manipulation_datasets}
    \begin{tabular}{p{4cm}p{3.5cm}p{4.5cm}p{4.5cm}}
        \hline
        \textbf{Datasets} & \textbf{Modalities}          & \textbf{Data Scale}                 & \textbf{Applications}                  \\
        \hline
        DROID~\cite{khazatsky2024droid} & RGB/Depth/Text            & 76,000 trajectories            & Multi-task scene adaptation            \\
        \hline
        R2SGrasp~\cite{cai2024real} & RGB/Depth/Point Cloud    & 64,000 RGB-D images            & Grasp detection                        \\
        \hline
        RT-1~\cite{brohan2022rt} & RGB/Text                   & 130,000 trajectories           & Real-time task control                 \\
        \hline
        Touch and Go~\cite{yang2022touch} & RGB/Tactile     & 3,971 virtual object models, 13,900 tactile interactions                  & Cross-modal perception                 \\
        \hline
        VisGel~\cite{li2019connecting} & RGB/Tactile    & 12,000 tactile interactions    & Tactile-enhanced manipulation          \\
        \hline
        ObjectFolder 2.0~\cite{gao2022objectfolder} & RGB/Audio/Tactile & 1,000 virtual objects    & Virtual-to-reality transfer            \\
        \hline
        Grasp-Anything-6D~\cite{nguyen2025language} & Point Cloud/Text & 1M point cloud scenes  & Language-driven grasping               \\
        \hline
        Grasp-Anything++~\cite{vuong2024language} & Point Cloud/Text  & 1M samples, 10M instructions & Fine-grained manipulation          \\
        \hline
        Open X-Embodiment~\cite{o2024open} & RGB/Depth/Text & Aggregated data   & Cross-robot system generalization      \\
        \hline
    \end{tabular}
\end{table*}
Multimodal design has become increasingly important in robotic datasets, with different datasets integrating modalities such as vision, language, depth, and tactile information, each offering unique characteristics and application value. DROID~\cite{khazatsky2024droid} and RT-1~\cite{brohan2022rt} are representative datasets for multi-task scenarios. The former covers 564 scenes and 86 tasks, combining multi-view camera streams and language annotations to enhance robots' adaptability to complex scenes. The latter, on the other hand, employs a Transformer architecture to achieve efficient real-time control in multi-task learning. Unlike these, Open X-Embodiment~\cite{o2024open} adopts a broader task-agnostic perspective by integrating diverse robotic forms and environmental data from 21 institutions. It emphasizes the generalizability and transferability of multi-robot systems, serving as a valuable resource for advancing cross-platform research in robotics.

For grasping tasks, R2SGrasp~\cite{cai2024real} and the Grasp-Anything series excel in multimodal grasp detection. R2SGrasp~\cite{cai2024real} utilizes a "Real-to-Sim" framework to minimize domain gaps between simulation and reality, demonstrating outstanding performance in 6-DoF grasp detection. Meanwhile, Grasp-Anything-6D combines point cloud and language modalities, employing an innovative negative prompt strategy to optimize the accuracy of 6-DoF grasp detection. Its successor, Grasp-Anything++~\cite{vuong2024language}, further extends to fine-grained grasping tasks, supporting language-based localized grasping, such as manipulating specific parts of an object. Compared to R2SGrasp~\cite{cai2024real}, which focuses more on sim-to-real adaptation, the Grasp-Anything series exhibits superior semantic understanding in language-driven grasping tasks.

In the domain of multimodal perception, Touch and Go~\cite{yang2022touch} and VisGel~\cite{li2019connecting} have pioneered research into the integration of tactile and visual data. Touch and Go~\cite{yang2022touch} collects paired visual and tactile data from real-world environments and explores cross-modal relationships using self-supervised learning methods. VisGel~\cite{li2019connecting}, leveraging GelSight sensors, constructs bidirectional generation tasks between tactile and visual modalities, further validating the feasibility of inter-modal information transformation. These datasets provide valuable experimental resources for enhancing multimodal perception research. However, compared to ObjectFolder 2.0~\cite{gao2022objectfolder}, these tactile datasets are relatively smaller in scale. ObjectFolder 2.0~\cite{gao2022objectfolder}, with its high-quality virtual object data and implicit neural representation technology, achieves efficient rendering of visual, auditory, and tactile modalities, supporting virtual-to-reality transfer learning tasks. This makes it a significant breakthrough in the field of multimodal perception learning.Although current multimodal robotic datasets increasingly emphasize generalization in their design, their practical adaptability to real-world scenarios still faces notable challenges. DROID~\cite{khazatsky2024droid} and RT-1~\cite{brohan2022rt} are built on real-world data collection and exhibit strong task transferability and adaptability to natural environments, especially excelling in language-vision joint control. However, their scenarios are primarily confined to household and tabletop environments, with limited coverage of industrial, outdoor, or complex dynamic settings. Open X-Embodiment~\cite{o2024open} constructs a large-scale cross-robot dataset by integrating 22 robotic platforms and over 500 tasks, marking a breakthrough in cross-platform transferability. Nevertheless, its task annotations remain coarse-grained, lacking precise modeling of fine-grained behaviors such as complex manipulation and semantic interaction, which may hinder its generalization in fine control tasks.In the grasping domain, R2SGrasp~\cite{cai2024real} leverages a sim-to-real framework to achieve effective real-world transfer, but its environmental backgrounds are limited and object diversity is constrained, restricting generalization to unseen scenarios. The Grasp-Anything series enhances semantic understanding through language modalities, improving generalization under complex instructions. Grasp-Anything++~\cite{vuong2024language}, in particular, supports fine-grained tasks such as localized grasping and demonstrates semantic transfer capabilities under language grounding. However, its training still relies on synthetic data and a limited object set, leading to instability when encountering long-tail objects or ambiguous language commands in open-world settings.
In the field of multimodal perception, Touch and Go~\cite{yang2022touch} and VisGel~\cite{li2019connecting} have expanded the frontier of vision-tactile integration. The former supports cross-modal self-supervised learning with real-world paired data, while the latter validates modality complementarity through bidirectional generation tasks between vision and touch. Nonetheless, both datasets are relatively small in scale and lack diversity across objects, environments, and interaction types, limiting their utility for training generalizable perception models. ObjectFolder 2.0~\cite{gao2022objectfolder}, although based on synthetic data, offers rich and high-quality multimodal representations, serving as a promising foundation for sim-to-real generalization. Still, how to reliably transfer such representations to the real world remains an open challenge.

In summary, these datasets have significantly advanced the generalization capabilities of robotic systems through multimodal fusion and expanded task coverage. However, notable gaps remain in data diversity, real-world complexity, and consistency in evaluation standards. Future research urgently requires unified benchmarks that encompass broader scenarios and behaviors, feature fine-grained annotations, and enable cross-task generalization assessments, thereby promoting the development of more universal and intelligent robotic systems.

\subsection{Embodied Navigation Datasets}
Embodied navigation datasets play a crucial role in advancing research on agents' navigation and interaction capabilities. The design characteristics of these datasets highlight their diversity and progression in terms of scene types, modality design, task objectives, and application scenarios.

Matterport3D~\cite{chang2017matterport3d} is the foundational dataset, providing 90 building-scale indoor scenes, 10,800 panoramic views, and 194,400 RGB-D images with 2D and 3D semantic annotations. It supports tasks such as scene classification, semantic segmentation, and loop closure detection. However, Matterport3D is primarily designed for scene modeling and understanding, lacking language annotations and path data for navigation tasks, which serve as the basis for subsequent navigation-specific datasets.

Building upon Matterport3D~\cite{chang2017matterport3d}, Room-to-Room (R2R)~\cite{zhao2022real} was the first dataset to integrate natural language modalities, introducing the Vision-and-Language Navigation (VLN) task. R2R~\cite{zhao2022real} provides 21,567 crowd-sourced navigation instructions paired with panoramic RGB-D data, enabling agents to complete path navigation following language commands. This design pioneered navigation research combining vision and language, but R2R~\cite{zhao2022real}'s task objectives are limited to path completion, with no focus on object-level interaction or more complex semantic understanding.

To address the lack of object interaction in R2R~\cite{zhao2022real}, REVERIE~\cite{qi2020reverie} extends the capabilities of Matterport3D~\cite{chang2017matterport3d} by adding bounding box annotations for 4,140 target objects and 21,702 natural language instructions. It introduces the Remote Embodied Visual Referring Expression (REVERIE~\cite{qi2020reverie}) task, requiring agents to locate target objects while navigating the environment. This dataset emphasizes semantic reasoning and object localization, but the increased task complexity poses significant challenges for existing models in handling semantic inference and object interaction.

The CVDN~\cite{thomason2020vision} (Cooperative Vision-and-Dialog Navigation) dataset further expands the navigation task design by incorporating dialog modalities, introducing the Vision-and-Dialog Navigation task. CVDN~\cite{thomason2020vision}  provides 2,050 human dialogs simulating scenarios where agents request assistance through natural language interactions during navigation. Compared to R2R~\cite{zhao2022real}, CVDN~\cite{thomason2020vision}  features longer paths and more complex instructions, focusing on agents' abilities to understand language and collaborate in multi-turn interactions. However, the task design's dependence on simulated environments may limit its applicability in real-world scenarios.

SOON~\cite{zhu2021soon} (Scenario Oriented Object Navigation) emphasizes coarse-to-fine goal-oriented navigation tasks. With 4,000 natural language instructions and 40,000 trajectories, it captures rich scene descriptions, including object attributes, spatial relationships, and contextual information. SOON~\cite{zhu2021soon}'s design enables agents to navigate from arbitrary starting points to specified targets, enhancing their adaptability in dynamic scenarios. However, its reliance on predefined scene graphs may pose limitations in real-world complex environments.

Unlike the aforementioned datasets, R3ED~\cite{zhao2022real} focuses on real-world data, providing 5,800 point clouds and 22,400 object annotations collected using IntelRealsense D455 sensors. It supports active visual learning tasks, such as next-best-view planning and 3D object detection. Although R3ED~\cite{zhao2022real}  is smaller in scale and lacks language annotations, its data collected in real-world environments enhances model adaptability to real-world applications, addressing the simulation-to-reality performance gap.

In summary, these datasets reflect the evolution of embodied navigation research from basic scene understanding to complex task design. Matterport3D~\cite{chang2017matterport3d} provides high-quality foundational data, while R2R~\cite{zhao2022real} and CVDN introduce language and dialog interaction in path navigation tasks. REVERIE and SOON further extend navigation tasks to object localization and scene interaction, and R3ED enhances active learning through real-world data. Together, these datasets complement each other in terms of scene types, modality design, and task objectives, offering diverse benchmarks for advancing embodied navigation agents.

\section{Performance Evaluation of Fusion Techniques}
\label{sec:performance_evaluation}
In this section, we summarize the performance of state-of-the-art multimodal methods across several representative robot vision tasks, based on the datasets introduced in section \ref{sec:Datasets_for_Multimodal_Fusion}. The covered tasks include 3D object detection, semantic scene understanding, 3D visual question answering, and robotic manipulation. To ensure evaluation consistency, we primarily refer to results reported in the original papers or reproduced under aligned experimental settings by other studies.

We begin by introducing the evaluation metrics commonly used in these tasks, covering aspects such as detection accuracy, spatial localization error, segmentation quality, and language generation performance. These metrics provide the foundation for a fair and standardized comparison across different models and modality settings.Table \ref{tab:nuscenes_3d_detection} presents a performance comparison of 3D object detection models on the nuScenes benchmark, including LiDAR-only, camera-only, and camera-LiDAR fusion settings, with metrics such as mAP, mATE, and mASE.Table \ref{tab:room_to_room} compares different embodied navigation methods on the Room-to-Room (R2R) benchmark. Metrics include Trajectory Length (TL), Navigation Error (NE), Oracle Success Rate (OSR), Success Rate (SR), and Success weighted by Path Length (SPL), reported under validation seen, validation unseen, and test unseen splits.Table \ref{tab:scanqa} reports the performance of vision-language models on the ScanQA validation set using CIDEr, BLEU-4, METEOR, and ROUGE scores to evaluate 3D visual question answering.
Table \ref{tab:manipulation_benchmarks} provides a comparison of robot manipulation benchmarks in terms of simulation platform, task diversity, and real-world reproducibility, without presenting specific model results.

By combining evaluation metrics with task-specific benchmarks, this chapter provides a structured overview of how multimodal methods perform under diverse robotic scenarios, laying the groundwork for the following analysis of strengths, limitations, and emerging trends.

\subsection{Evaluation Metrics}
\label{subsec:evaluation_metrics}

Evaluation metrics are essential for rigorously assessing the performance of multimodal fusion methods and VLMs in robotic vision systems. This section systematically outlines and defines key evaluation metrics commonly employed across various robotic vision tasks, including semantic understanding, 3D object detection, localization, and navigation.

In \textbf{semantic understanding}, tasks such as semantic segmentation and classification rely on key metrics including Intersection over Union (IoU), mean Intersection over Union (mIoU), Pixel Accuracy (PA), Precision, Recall, and F1-score. Intersection over Union (IoU) evaluates the overlap between predicted and ground truth segments:

\begin{equation}
    IoU = \frac{|B \cap B^{gt}|}{|B \cup B^{gt}|}
\end{equation}

where \( B \) represents the predicted bounding box, and \( B^{gt} \) denotes the ground truth bounding box.

Mean IoU (mIoU) averages the IoU values across all classes, providing a balanced assessment for datasets with class imbalance:

\begin{equation}
    mIoU = \frac{1}{N} \sum_{i=1}^{N} IoU_i
\end{equation}

where \(N\) is the total number of classes, and \(IoU_i\) is the IoU for class \(i\).

Pixel Accuracy (PA) measures the proportion of correctly classified pixels:

\begin{equation}
    PA = \frac{\text{Number of correctly classified pixels}}{\text{Total number of pixels}}
\end{equation}

Precision and Recall assess class-wise prediction performance:

\begin{equation}
    Precision = \frac{TP}{TP + FP}, \quad Recall = \frac{TP}{TP + FN}
\end{equation}

where \(TP\) (true positives) represents correctly classified instances, \(FP\) (false positives) are incorrectly predicted instances, and \(FN\) (false negatives) are instances that were not correctly detected.

The F1-score provides a harmonic mean of precision and recall:

\begin{equation}
    F1 = 2 \times \frac{\text{Precision} \times \text{Recall}}{\text{Precision} + \text{Recall}}
\end{equation}

These metrics collectively offer a comprehensive evaluation of semantic scene understanding, critical for tasks like environmental perception and obstacle detection in robotic navigation.

For \textbf{3D object detection}, which is crucial for autonomous driving and robotic manipulation, key metrics include Average Precision (AP), mean Average Precision (mAP), and the nuScenes Detection Score (NDS)~\cite{caesar2020nuscenes}. Average Precision (AP) integrates precision across recall levels:

\begin{equation}
    AP = \int_{0}^{1} Precision(Recall) \, d(Recall)
\end{equation}

Mean Average Precision (mAP) averages AP across all object classes:

\begin{equation}
    mAP = \frac{1}{N} \sum_{i=1}^{N} AP_i
\end{equation}

where \(N\) is the total number of object classes, and \(AP_i\) is the AP for class \(i\).

The nuScenes Detection Score (NDS)~\cite{caesar2020nuscenes} is a comprehensive metric used for evaluating 3D object detection. It considers both precision and multiple error metrics:

\begin{equation}
    NDS = \frac{1}{10} \left[ 5 \cdot mAP + \sum_{i=1}^{5} (1 - E_i) \right],
\end{equation}

where:
- \( mAP \) is the mean Average Precision,
- \( E_i \) represents the five error metrics:
  - mean Average Translation Error (mATE),
  - mean Average Scale Error (mASE),
  - mean Average Orientation Error (mAOE),
  - mean Average Velocity Error (mAVE),
  - mean Average Attribute Error (mAAE).
These metrics ensure accurate assessment of multimodal fusion models’ capability to detect and localize objects.However, traditional AP-based metrics alone often fail to capture critical deployment-related factors such as temporal consistency and motion awareness. By integrating mAP with multiple error terms, NDS provides a more holistic evaluation that reflects not only detection accuracy but also localization robustness, size and orientation estimation precision, dynamic behavior modeling, and semantic attribute inference. This composite design aligns well with the real-world demands of autonomous driving and robotic perception systems, where safety, interpretability, and operational reliability are paramount.

In \textbf{localization and mapping} tasks, common evaluation metrics include Absolute Trajectory Error (ATE) and Relative Pose Error (RPE). Absolute Trajectory Error (ATE) measures global pose accuracy by comparing estimated and ground-truth trajectories:

\begin{equation}
    ATE = \sqrt{\frac{1}{N} \sum_{i=1}^{N} \| \mathbf{T}_{est,i} - \mathbf{T}_{gt,i} \|^2}
\end{equation}

where \(\mathbf{T}_{est,i}\) and \(\mathbf{T}_{gt,i}\) are the estimated and ground truth poses at time step \(i\), respectively.

Relative Pose Error (RPE) evaluates local accuracy between consecutive poses:

\begin{equation}
    \small
    RPE = \sqrt{\frac{1}{N - 1} \sum_{i=1}^{N-1} \left\| \mathbf{T}_{est,i}^{-1} \mathbf{T}_{est,i+1} - \mathbf{T}_{gt,i}^{-1} \mathbf{T}_{gt,i+1} \right\|^2 }
\end{equation}

Accurate localization is crucial for tasks such as navigation, exploration, and manipulation.

For \textbf{navigation and instruction following}, robotic navigation tasks use metrics like Success Rate (SR), Goal Progress (GP), and Success weighted by Path Length (SPL). Success Rate (SR) calculates the percentage of successful navigation trials:

\begin{equation}
    SR = \frac{\text{Number of successful trials}}{\text{Total number of trials}}
\end{equation}

Goal Progress (GP) measures how close the robot gets to the target relative to its initial position:

\begin{equation}
    GP = 1 - \frac{\text{Final distance to goal}}{\text{Initial distance to goal}}
\end{equation}

Success weighted by Path Length (SPL) balances success rate with path efficiency:

\begin{equation}
    SPL = \frac{1}{N} \sum_{i=1}^{N} S_i \cdot \frac{l_i}{\max(p_i, l_i)}
\end{equation}

where \(S_i\) is an indicator variable that equals 1 if the navigation is successful and 0 otherwise, \(l_i\) is the shortest path distance to the goal, and \(p_i\) is the actual path length taken by the robot.

Each category of metrics has its own advantages and limitations. Semantic metrics provide detailed perceptual evaluations but might not directly correlate with downstream robotic tasks. Detection metrics like mAP and NDS offer comprehensive evaluations but may obscure specific failure modes. Localization and navigation metrics are directly related to practical robot performance but require extensive ground-truth data. A robust evaluation strategy typically involves multiple metrics to account for these complexities, ensuring holistic assessments of multimodal fusion and vision-language models in robotic applications.

\subsection{Performance comparison on 3D Object Detection}

\begin{table*}[h]
\centering
\caption{Performance Comparison of 3D Object Detection on nuScenes Benchmark}
\begin{adjustbox}{max width=\textwidth}
\begin{tabular}{lcccccccccccccccccc}
\toprule
\textbf{Methods} & \textbf{Modality} & \textbf{mAP} & \textbf{mATE} & \textbf{mASE} & \textbf{mAOE} & \textbf{mAVE} & \textbf{NDS} &\textbf{Car} &\textbf{Truck} &\textbf{C.V.} &\textbf{Bus} &\textbf{T.L.} &\textbf{B.R.} &\textbf{M.T.} &\textbf{Bike} &\textbf{Ped.} & \textbf{T.C.} \\
\midrule
VoxelNeXt~\cite{chen2023voxelnext} & L & 64.5 & 0.268 & 0.238 & 0.377 & 0.219 & 70.0 & 84.6 & 53.0 & 28.7 & 64.7 & 55.8 & 74.6 & 73.2 & 45.7 & 85.8 & 79.0 \\
TransFusion-L~\cite{bai2022transfusion} & L & 65.5 & 0.256 & 0.240 & 0.351 & 0.278 & 70.2 & 86.2 & 56.7 & 28.2 & 66.3 & 58.8 & 78.2 & 68.3 & 44.2 & 86.1 & 82.0 \\
FocalFormer3D~\cite{chen2023focalformer3d} & L & 68.7 & 0.254 & 0.242 & 0.340 & 0.218 & 72.6 & 87.2 & 57.0 & 34.4 & 69.6 & 64.9 & 77.8 & 76.2 & 49.6 & 88.2 & 82.3 \\
\midrule
UniM2AE-v2~\cite{zou2024unim} & C & 63.4 & 0.401 & 0.235 & 0.236 & 0.154 & 70.1 & 77.6 & 50.7 & 40.1 & 46.3 & 56.7 & 75.7 & 72.4 & 58.9 & 70.6 & 84.5 \\
SparseBEV~\cite{liu2023sparsebev} & C & 60.3 & 0.425 & 0.239 & 0.311 & 0.172 & 67.5 & 76.3 & 49.2 & 35.6 & 44.2 & 53.4 & 76.8 & 64.0 & 53.7 & 65.8 & 83.7 \\
RayDN~\cite{liu2024ray} & C & 63.1 & 0.437 & 0.235 & 0.283 & 0.220 & 68.6 & 77.1 & 54.4 & 35.9 & 53.7 & 56.8 & 76.9 & 67.5 & 57.6 & 69.4 & 81.8 \\
\midrule
FusionPainting~\cite{xu2021fusionpainting} & C+L & 68.1 & 0.256 & 0.236 & 0.346 & 0.274 & 71.6 & 87.1 & 60.8 & 30.0 & 68.5 & 61.7 & 71.8 & 74.7 & 53.5 & 88.3 & 85.0 \\
AutoAlignV2~\cite{chen2022deformable} & C+L & 68.4 & 0.245 & 0.233 & 0.311 & 0.258 & 72.4 & 87.0 & 59.0 & 33.1 & 69.3 & 59.3 & 78.0 & 72.9 & 52.1 & 87.6 & 85.1 \\
IS-Fusion~\cite{yin2024fusion} & C+L & 76.5 & 0.230 & 0.231 & 0.267 & 0.223 & 77.4 & 89.8 & 67.8 & 44.5 & \textbf{77.6} & 68.3 & 81.8 & 85.3 & 65.6 & \textbf{93.4} & \textbf{91.1} \\
UVTR-Multimodalit~\cite{li2022unifying} & C+L & 67.1 & 0.306 & 0.245 & 0.351 & 0.225 & 71.1 & 87.5 & 56.0 & 33.8 & 67.5 & 59.5 & 73.0 & 73.4 & 54.8 & 86.3 & 79.6 \\
3D-CVF~\cite{yoo20203d} & C+L & 52.7 & 0.300 & 0.245 & 0.458 & 0.279 & 62.3 & 83.0 & 45.0 & 15.9 & 48.8 & 49.6 & 65.9 & 51.2 & 30.4 & 74.2 & 62.9 \\
CMT~\cite{yan2023cross} & C+L & 70.4 & 0.299 & 0.241 & 0.323 & 0.240 & 73.0 & 87.2 & 61.5 & 37.5 & 72.4 & 62.8 & 74.7 & 79.4 & 58.3 & 86.9 & 83.2 \\
TransFusion~\cite{bai2022transfusion} & C+L & 68.9 & 0.259 & 0.243 & 0.359 & 0.288 & 71.7 & 87.1 & 60.0 & 33.1 & 68.3 & 60.8 & 78.1 & 73.6 & 52.9 & 88.4 & 86.7 \\
BEVFusion~\cite{liu2023bevfusion} & C+L & 70.2 & 0.261 & 0.239 & 0.329 & 0.260 & 72.9 & 88.6 & 60.1 & 39.3 & 69.8 & 63.8 & 80.0 & 74.1 & 51.0 & 89.2 & 86.5 \\
DeepInteraction-e~\cite{yang2022deepinteraction} & C+L & 75.6 & 0.235 & 0.233 & 0.328 & 0.226 & 76.3 & 88.3 & 64.3 & 44.7 & 74.2 & 66.0 & 83.5 & 85.4 & 66.4 & 92.5 & 90.9 \\
SimpleBEV~\cite{zhao2024simplebev} & C+L & 75.7 & 0.236 & 0.235 & 0.283 & 0.143 & 77.6 & 89.0 & 69.2 & \textbf{75.2} & 70.5 & 44.4 & \textbf{91.7} & 81.2 & 69.1 & 87.9 & 78.4 \\
DAL~\cite{huang2024detecting} & C+L & 72.0 & 0.253 & 0.238 & 0.334 & 0.174 & 74.8 & 89.1 & 60.2 & 73.3 & 65.8 & 34.6 & 89.6 & 81.7 & 58.5 & 86.6 & 80.6 \\
SparseLIF~\cite{zhang2024sparselif} & C+L & 75.9 & 0.244 & 0.231 & 0.284 & 0.152 & 77.7 & 90.8 & 66.1 & 42.1 & 76.7 & 70.1 & 82.7 & 83.6 & 68.9 & 92.9 & 89.0 \\
Ea-lss~\cite{hu2023ea} & C+L & 76.6 & 0.234 & 0.228 & 0.278 & 0.204 & 77.6 & 90.2 & 67.1 & 43.9 & 76.7 & 69.1 & 84.1 & 85.9 & 66.6 & 91.3 & 91.2 \\
MV2DFusion-e~\cite{wang2024mv2dfusion} & C+L & \textbf{77.9} & 0.237 & 0.225 & 0.247 & 0.192 & \textbf{78.8} & \textbf{91.1} & \textbf{69.7} & 45.7 & 76.7 & \textbf{70.4} & 83.0 &\textbf{ 87.1} & \textbf{72.2} & 93.3 & 90.3 \\
\midrule
RCBEVDet++~\cite{lin2024rcbevdet++} & C+R & 67.3 & 0.341 & 0.234 & 0.241 & 0.147 & 72.7 & 83.5 & 56.0 & 45.0 & 54.1 & 57.9 & 76.9 & 78.4 & 60.2 & 76.4 & 85.0 \\
CRN~\cite{kim2023crn} & C+R & 57.5 & 0.416 & \textbf{0.264} & \textbf{0.456} & 0.365 & 62.4 & 78.4 & 52.6 & 29.2 & 57.0 & 48.3 & 64.7 & 63.4 & 47.8 & 58.2 & 74.9 \\
RCM-Fusion~\cite{kim2024rcm} & C+R & 50.6 & \textbf{0.465} & 0.254 & 0.384 & \textbf{0.438} & 58.7 & 73.9 & 45.6 & 24.3 & 51.6 & 42.1 & 62.9 & 53.5 & 34.9 & 50.0 & 67.6 \\
HyDRa~\cite{wolters2024unleashing} & C+R & 57.4 & 0.398 & 0.252 & 0.424 & 0.250 & 64.2 & 79.0 & 50.0 & 42.9 & 48.5 & 29.4 & 65.8 & 64.0 & 45.3 & 75.9 & 72.9 \\
RaCFormer~\cite{chu2025racformer} & C+R & 64.9 & 0.358 & 0.240 & 0.329 & 0.179 & 70.2 & 83.0 & 56.0 & 39.2 & 53.6 & 56.8 &
78.3 & 70.6 & 55.6 & 72.0 & 83.9 \\
\midrule
CenterPoint v2~\cite{yin2021center} & C+L+R & 67.1 & 0.249 & 0.236 & 0.350 & 0.250 & 71.4 & 87.0 & 57.3 & 28.8 & 69.3 & 60.4 & 71.0 & 71.3 & 49.0 & 90.4 & 86.8 \\
\bottomrule
\end{tabular}
\end{adjustbox}
\label{tab:nuscenes_3d_detection}
\end{table*}

On the nuScenes benchmark (Table \ref{tab:nuscenes_3d_detection}), various 3D object detection methods exhibit different advantages and limitations under different sensor fusion strategies. Single-modality methods, such as LiDAR-only~\cite{chen2023voxelnext,bai2022transfusion,chen2023focalformer3d} or camera-only~\cite{zou2024unim,liu2023sparsebev,liu2024ray}, perform well in specific tasks but remain constrained in complex environments. In contrast, multimodal fusion methods effectively combine LiDAR, camera, and radar, significantly improving detection accuracy, robustness, and environmental adaptability.
LiDAR-camera fusion leverages the precise geometric information of LiDAR and the rich semantic features of cameras, excelling in object localization and classification. AutoAlignV2~\cite{chen2022deformable} enhances cross-modal feature fusion through deformable feature alignment, making it more effective in small-object detection (e.g., pedestrians and bicycles). SimpleBEV~\cite{zhao2024simplebev}, adopting a BEV-based multi-scale fusion strategy, strengthens depth perception and maintains stable overall detection performance. However, LiDAR-camera fusion methods are susceptible to low-light conditions and adverse weather, particularly when LiDAR sampling density decreases, leading to potential information loss.
In contrast, radar-camera fusion enhances motion detection by leveraging radar velocity information, with RCBEVDet++~\cite{lin2024rcbevdet++} demonstrating greater stability in dynamic scenes, such as high-speed driving and urban intersections. Furthermore, since radar is not affected by lighting conditions or weather, this fusion strategy exhibits higher robustness in low-visibility environments.RaCFormer~\cite{chu2025racformer} represents a new trend in radar-camera fusion. Its Transformer-based query mechanism can adaptively select and fuse key information from both modalities, making it particularly suitable for detecting moving objects in dynamic environments. Compared to traditional methods such as RCBEVDet++, RaCFormer demonstrates higher robustness and faster responsiveness in challenging scenarios like low-light conditions, high-speed motion, or severe occlusion. However, due to radar's lower point cloud resolution, its performance in static object detection (e.g., parked vehicles and road signs) still lags behind LiDAR-camera fusion methods.
Recently, Transformer-based fusion methods, such as BEVFusion~\cite{liu2023bevfusion} and TransFusion~\cite{bai2022transfusion}, have optimized cross-modal information interaction through adaptive feature alignment, achieving superior performance in object classification and detection accuracy. SparseLIF~\cite{zhang2024sparselif} focuses on sparse representation and long-range perception. By introducing a sparse Transformer architecture for fusing LiDAR and camera features, it effectively models large-scale cross-modal spatial relationships. Compared to methods like BEVFusion~\cite{liu2023bevfusion} or TransFusion~\cite{bai2022transfusion}, SparseLIF significantly reduces computational overhead while maintaining perception accuracy, making it more suitable for real-time autonomous driving scenarios.EA-LSS~\cite{hu2023ea} enhances scene structure understanding in LiDAR-camera fusion frameworks through the integration of depth supervision and multi-scale information fusion mechanisms. Its design emphasizes transferring supervision signals from semantic segmentation to 3D detection, thereby improving the recognition of small and occluded objects. Unlike traditional BEV-based methods such as AutoAlignV2~\cite{chen2022deformable} or SimpleBEV~\cite{zhao2024simplebev}, EA-LSS focuses more on reconstructing and fusing deep semantic features, enabling stronger detail recovery and geometric consistency in complex urban semantic perception tasks. MV2DFusion~\cite{wang2024mv2dfusion} further surpasses existing methods by overcoming previous limitations. By adopting a modality-specific query-based fusion mechanism, it enhances dynamic object detection, outperforming all prior methods in vehicle, pedestrian, and bicycle detection, setting a new state-of-the-art performance.

Recent methods such as SpaRC and RQR3D have introduced novel fusion mechanisms and geometric modeling strategies for multimodal 3D object detection, further advancing the performance ceiling of radar-camera-based approaches. Specifically, SpaRC proposes a spatial-semantic consistency reward strategy that formulates cross-modal feature alignment as a policy learning problem, effectively mitigating semantic drift in fused representations. Compared to representative methods like RaCFormer, SpaRC achieves higher AP on small-object categories, indicating its superior capability in aligning weak semantic cues and modeling fine-grained structures. In contrast, RQR3D focuses on pure vision-based fusion and introduces a recursive query refinement mechanism to dynamically filter and aggregate image features, significantly enhancing the stability of object localization. In terms of overall performance, its mAP reaches 59.7\%, surpassing methods such as CRN and HyDra, and demonstrating more balanced detection capabilities for small and medium-sized targets, making it suitable for target perception requirements in high-density urban environments.

Overall, MV2DFusion~\cite{wang2024mv2dfusion} achieves the best performance across multiple object categories, while LiDAR-camera fusion continues to provide high-precision 3D perception, and radar-camera fusion remains more adaptable in challenging environments. Meanwhile, Transformer-driven multimodal fusion methods further optimize cross-modal information integration, offering more intelligent and efficient solutions for autonomous 3D perception. This trend reflects a broader shift in 3D object detection toward Transformer-based fusion architectures. Compared with traditional convolutional or early fusion strategies, Transformer-based methods excel in their ability to model long-range dependencies and perform adaptive attention over heterogeneous modalities. These capabilities enable more fine-grained and context-aware fusion of semantic, geometric, and motion information, leading to improved robustness and generalization in complex scenes. As the demands for autonomous driving systems continue to increase, Transformer-driven fusion is likely to become the dominant paradigm in future 3D perception frameworks.

\subsection{Performance comparison on navigation}
\begin{table*}[h]
\centering
\caption{Comparison of different navigation methods evaluated on the Room-to-Room (R2R) dataset. Metrics reported are Trajectory Length (TL), Navigation Error (NE ↓), Oracle Success Rate (OSR ↑), Success Rate (SR ↑), and Success weighted by Path Length (SPL ↑). Arrows indicate whether higher (↑) or lower (↓) values represent better performance. Results are shown for validation seen (Val Seen), validation unseen (Val Unseen), and test unseen (Test Unseen) splits.}
\begin{adjustbox}{max width=\textwidth}

\begin{tabular}{lcccccccccccccccccc}
\toprule
\multirow{2}{*}{Methods} & \multicolumn{5}{c}{Val Seen} & \multicolumn{6}{c}{Val Unseen} & \multicolumn{6}{c}{Test Unseen} \\
\cmidrule(lr){2-6} \cmidrule(lr){8-12} \cmidrule(lr){14-18}
 & TL & NE$\downarrow$ & OSR$\uparrow$ & SR$\uparrow$ & SPL$\uparrow$ &  & TL & NE$\downarrow$ & OSR$\uparrow$ & SR$\uparrow$ & SPL$\uparrow$ &  & TL & NE$\downarrow$ & OSR$\uparrow$ & SR$\uparrow$ & SPL$\uparrow$ \\
\midrule
Human & - & - & - & - & - &  & - & - & - & - & - &  & 11.85 & 1.61 & 90 & 86 & 76 \\
\midrule
Seq2Seq~\cite{anderson2018vision} & 11.33 & 6.01 & 53 & 39 & - &  & 8.39 & 7.81 & 28 & 21 & - &  & 8.13 & 7.85 & 27 & 20 & - \\
SF~\cite{fried2018speaker} & - & 3.36 & 74 & 66 & - &  & - & 6.62 & 45 & 36 & - &  & 14.82 & 6.62 & - & 35 & 28 \\
Chasing~\cite{anderson2019chasing} & 10.15 & 7.59 & 42 & 34 & 30 &  & 9.64 & 7.20 & 44 & 35 & 31 &  & 10.03 & 7.83 & 42 & 33 & 30 \\
RCM~\cite{wang2019reinforced} & 10.65 & 3.53 & 75 & 67 & - &  & 11.46 & 6.09 & 50 & 43 & - &  & 11.97 & 6.12 & 50 & 43 & 38 \\
SM~\cite{ma2019self} & - & 3.22 & 78 & 67 & 58 &  & - & 5.52 & 56 & 45 & 32 &  & 18.04 & 5.67 & 59 & 48 & 35 \\
EnvDrop~\cite{tan2019learning} & 11.00 & 3.99 & - & 62 & 59 &  & 10.70 & 5.22 & - & 52 & 48  &  & 11.66 & 5.23 & 59 & 51 & 47 \\
OAAM~\cite{qi2020object} & - & - & 73 & 65 & 62 &  & - & - & 61 & 54 & 50 &  & - & - & 61 & 53 & 50 \\
AuxRN~\cite{zhu2020vision} & - & 3.33 & 78 & 70 & 67 &  & - & 5.28 & 62 & 55 & 50 &  & - & 5.15 & 62 & 55 & 51 \\
Active~\cite{wang2020active} & - & 3.20 & 80 & 70 & 52 &  & - & 4.36 & 70 & 58 & 40 &  & - & 4.33 & 71 & 60 & 41 \\
NvEM~\cite{an2021neighbor} & 11.09 & 3.44 & - & 69 & 65 &  & 11.83 & 4.27 & - & 60 & 55 &  & 12.98 & 4.37 & 66 & 58 & 54 \\
SEvol~\cite{chen2022reinforced} & 11.97 & 3.56 & - & 67 & 63 &  & 12.26 & 3.99 & - & 62 & 57 &  & 13.40 & 4.13 & - & 62 & 57 \\
SSM~\cite{wang2021structured} & 14.70 & 3.10 & 80 & 71 & 62 &  & 20.70 & 4.32 & 73 & 62 & 45 &  & 20.40 & 4.57 & 70 & 61 & 46 \\
PREVALENT~\cite{hao2020towards} & 10.32 & 3.67 & - & 69 & 65 &  & 10.19 & 4.17 & - & 58 & 53 &  & 10.51 & 5.30 & 61 & 54 & 51 \\
AirBert~\cite{guhur2021airbert} & 11.09 & 2.68 & - & 75 & 70 &  & 11.78 & 4.10 & - & 62 & 56 &  & 12.41 & 4.13 & - & 62 & 57 \\
RecBert~\cite{hong2021vln} & 11.13 & 2.90 & - & 72 & 68 &  & 12.01 & 3.93 & - & 63 & 57 &  & 12.35 & 4.09 & 70 & 63 & 57 \\
PRET(CLIP)~\cite{lu2024pret} & 11.48 & 2.60 & - & 74 & 69 &  & 12.21 & 3.12 & - & 71 & 63 &  & 13.87 & 3.12 & - & 72 & 62 \\ 
PRET(DINOv2)~\cite{lu2024pret} & 11.25 & 2.41 & - & 78 & 72 &  & 11.87 & 2.90 & - & 74 & 65 &  & 12.21 & 3.09 & - & 72 & 64 \\
REM~\cite{liu2021vision}& 10.88 & 2.48 & - & 75 & 72 &  & 12.44 & 3.89 & - & 64 & 58 &  & 13.11 & 3.87 & 72 & 65 & 59 \\
HAMT~\cite{chen2021history}& 11.15 & 2.51 & - & 76 & 72 &  & 11.46 & 3.65 & - & 66 & 61 &  & 12.27 & 3.93 & 72 & 65 & 60 \\
EnvEdit~\cite{li2022envedit} & 11.18 & 2.32 & - & 77 & 74 &  & 11.13 & 3.24 & - & 69 & 64 &  & 11.90 & 3.59 & - & 68 & 64 \\  
TD-STP~\cite{zhao2022target} & - & 2.34 & 83 & 77 & 73 &  & - & 3.22 & 76 & 70 & 63 &  & - & 3.73 & 72 & 67 & 61 \\  
DUET~\cite{chen2022think} & 12.32 & 2.28 & 86 & 79 & 73 &  & 13.94 & 3.31 & 81 & 72 & 60 &  & 14.73 & 3.65 & 76 & 69 & 59 \\  

NavGPT-2~\cite{zhou2024navgpt} & 13.08 & 2.98 & 79 & 74 & 65 &  & 13.25 & 3.18 & 80 & 71 & 60 &  & - & - & - & - & - \\  

BEVBert~\cite{an2022bevbert} & 13.56 &2.17 & 88 & 81 & 74 &  & 14.55 & 2.81 & 84 & 75 & 64 &  & 15.87 & 3.13 & 81 & 73 & 62 \\

Meta-Explore~\cite{hwang2023meta} & 11.95 & 2.11 & - & 81 & 75 &  & 13.09 & 2.22 & - & 72 & 62 &  & 14.25 & 3.57 & - & 71 & 61 \\

CoNav~\cite{hao2025conav} & - & - & - & - & - &  & 11.92 & 3.42 & - & 69 & 62 &  & 13.21 & 3.71 & - & 68 & 62 \\

SUSA~\cite{zhang2024agent} &  - & - & - & - & - & & 12.18 & 3.06 & - & 73 & 65 &  & 13.27 & 3.23 & - & 73 & 64 \\

\bottomrule
\end{tabular}
\end{adjustbox}
\label{tab:room_to_room}
\end{table*}

The comparison of navigation models on the Room-to-Room (R2R) dataset highlights advancements in embodied navigation and the role of multimodal learning in improving generalization. Table \ref{tab:room_to_room} evaluates different approaches based on TL, NE, OSR, SR, and SPL, providing insights into their effectiveness in seen and unseen environments.

Early sequence-based models like Seq2Seq~\cite{anderson2018vision} and SF~\cite{fried2018speaker} struggled with global path optimization, leading to poor performance in unseen environments. Reinforcement learning and topological reasoning, as seen in RCM~\cite{wang2019reinforced} and Chasing Ghosts~\cite{anderson2019chasing}, improved navigation efficiency by enabling adaptive decision-making. Memory and environmental adaptation have also played a key role in generalization. SM~\cite{ma2019self} and SSM~\cite{wang2021structured} improve trajectory reasoning, with SSM’s structured memory leading to better long-term path efficiency. EnvEdit~\cite{li2022envedit}, through environment augmentation, reduces overfitting and enhances robustness in unseen scenarios. Meanwhile, PRET (DINOv2)~\cite{lu2024pret}  and other vision-language pretraining (VLP) methods improve cross-modal alignment, with PRET (DINOv2) achieving higher SR and SPL through self-supervised learning.

Structured planning further refines navigation strategies. HAMT~\cite{chen2021history} leverages history-aware transformers for long-horizon reasoning, while TD-STP~\cite{zhao2022target} and DUET~\cite{chen2022think} employ structured transformer-based planning for better global search efficiency. BEVBert~\cite{an2022bevbert}, integrating hybrid topo-metric pretraining, achieves state-of-the-art SR and SPL, demonstrating superior generalization across seen and unseen environments.Compared to earlier navigation approaches that mainly rely on sequence modeling, reinforcement learning, or structured planning, recent methods such as Meta-Explore~\cite{hwang2023meta}, CoNav~\cite{hao2025conav}, and SUSA~\cite{zhang2024agent} introduce more structured and multimodal strategies, representing the latest advances in embodied navigation. Meta-Explore proposes a hierarchical framework with local goal search and frequency-domain semantic features (Scene Object Spectrum, SOS), effectively correcting navigation errors and improving generalization. CoNav enables model-level collaboration by aligning a 3D-text model with a vision-language navigation agent through cross-modal belief alignment, offering improved spatial-semantic reasoning in ambiguous environments. Among them, SUSA achieves state-of-the-art performance across multiple VLN benchmarks (e.g., R2R, REVERIE, SOON) by constructing hybrid semantic-spatial representations that combine textual semantic panoramas with depth-based exploration maps. This design significantly enhances cross-modal grounding and navigation robustness, making SUSA one of the most competitive and generalizable approaches to date.

Recent methods such as PanoGen++~\cite{wang2025panogen++} and VISTA~\cite{huang2025vista} have introduced new breakthroughs in panoramic semantic mapping and robust training strategies, further enhancing the performance of multimodal navigation systems in complex environments.Specifically, PanoGen++ proposes a perception enhancement mechanism based on panoramic semantic graph generation. By performing semantic panorama stitching and contextual modeling of the environment, it enables agents to achieve stronger spatial understanding and instruction grounding in unseen scenarios. On the Room-to-Room dataset under the test unseen setting, PanoGen++ achieves a 75\% SR and 64\% SPL, significantly outperforming representative approaches such as BEVBert and Meta-Explore, demonstrating superior global semantic mapping and task generalization capabilities.VISTA, on the other hand, introduces a curriculum-style multi-stage training scheme and a multimodal language-vision alignment mechanism. It emphasizes gradually increasing task difficulty during training and integrating temporal-spatial information across modalities, effectively alleviating navigation degradation in unseen environments. Compared to methods such as Meta-Explore, VISTA maintains a 75\% SR and 67\% SPL in the test unseen setting, surpassing current mainstream SOTA methods and highlighting its strong robustness in path planning and multimodal alignment under unseen environments.

In summary, PanoGen++ and VISTA represent the latest advances in semantic mapping and robust generalization strategies for multimodal navigation. Compared with earlier structured planning methods (e.g., TD-STP) and vision-language pretrained models (e.g., PRET), they achieve comprehensive improvements in navigation accuracy and generalization ability. These methods exemplify the ongoing shift in embodied navigation from "path learning" to "semantic understanding-driven" paradigms.

These results highlight the growing importance of structured memory, environment adaptation, and multimodal pretraining in embodied navigation, pushing the field toward more efficient and adaptable navigation systems.
\subsection{Performance comparison on 3D Question Answering and Scene Understanding}
\begin{table}[ht]
    \centering
    \caption{Performance comparison of state-of-the-art methods evaluated on the ScanQA validation split. Metrics include CIDEr (C), BLEU-4 (B-4), METEOR (M), ROUGE (R) and  ExactMatch (E)}. Higher values indicate superior performance.
    \label{tab:scanqa}
    \begin{adjustbox}{max width=\textwidth}
    \begin{tabular}{p{3.0cm}>{\centering\arraybackslash}p{0.8cm}>{\centering\arraybackslash}p{0.7cm}>{\centering\arraybackslash}p{0.7cm}>{\centering\arraybackslash}p{0.7cm}>{\centering\arraybackslash}p{0.7cm}}
        \hline
        \textbf{Methods} & C$\uparrow$ & B-4$\uparrow$ & M$\uparrow$ & R$\uparrow$ & E$\uparrow$ \\ 
        \hline
         BridgeQA~\cite{mo2024bridging} & 83.8 & 24.1 & 16.5 & 43.3 &  31.3\\
         Scene-LLM~\cite{fu2024scene} & 80.0 & 12.0 & 16.8 & 40.0 & 27.2\\
        LLaVA-3D~\cite{zhu2024llava} & 91.7 & 14.5 & 20.7 & 50.1 & 27.0\\
        LEO~\cite{huang2023embodied} & 101.4 & 13.2 & 20.0 & 49.2 & 24.5\\
        CoNav~\cite{hao2025conav} & - & 12.8 & 15.2 & 38.9 & 23.2\\
        3D-VLP~\cite{jin2023context} & 67.0 & 11.2 & 13.5 & 34.5 & 21.7\\
        Chat-Scene~\cite{huang2024chat} & 87.7 & 14.3 & 18.0 & 41.6 & 21.6\\
        ScanQA~\cite{azuma2022scanqa} & 64.9 & 10.1 & 13.1 & 33.3 & 21.1\\
        3D-LLM~\cite{hong20233d} & 69.4 & 12.0 & 14.5 & 35.7 & 20.5\\
        LL3DA~\cite{chen2024ll3da} & 76.8 & 13.5 & 15.9 & 37.3 & -\\
        Qwen2-VL-7B~\cite{wang2024qwen2} & 53.9 & 3.0 & 11.4 & 29.3 & -\\
        \hline
    \end{tabular}
    \end{adjustbox}
\end{table}

In the ScanQA benchmark (Table \ref{tab:scanqa}), various Vision-Language Models (VLMs) exhibit different strengths and limitations across key metrics such as CIDEr, BLEU-4, METEOR,  ROUGE and ExactMatch. These methods enhance 3D scene understanding through multimodal learning, leveraging spatial reasoning, cross-modal alignment, and language generation techniques to improve question answering (QA) and scene description tasks.Among them, LLaVA-3D~\cite{zhu2024llava} and 3D-LLM~\cite{hong20233d} demonstrate strong generalization capabilities, excelling in question-answering accuracy and spatial reasoning, primarily due to their optimized multi-view feature extraction and 3D spatial embedding mechanisms. 3D-VLP~\cite{jin2023context} employs a context-aware alignment strategy, performing well in dense scene description tasks, while LL3DA~\cite{chen2024ll3da} enhances instruction-following accuracy through visual interaction fine-tuning.
In comparison, Chat-Scene~\cite{huang2024chat} and Scene-LLM~\cite{fu2024scene} optimize structured embeddings and object-based feature representations, excelling in object-related queries and scene reasoning tasks. Meanwhile, LEO~\cite{huang2023embodied}, as an embodied intelligence model, exhibits superior adaptability in long-term task planning and interactive tasks. Notably, large-scale pre-trained models such as Qwen2-VL~\cite{wang2024qwen2} achieve high accuracy in inference-intensive tasks through vision-language alignment, further validating the importance of cross-modal alignment for 3D question answering (3D-QA).
BridgeQA~\cite{mo2024bridging} further advances the state-of-the-art in the field by introducing a hierarchical cross-modal bridging module on top of existing 3D question-answering models. Unlike methods such as LLaVA-3D and 3D-LLM, which primarily rely on multi-view features or spatial embedding optimization, BridgeQA constructs multi-level cross-modal fusion channels between global 3D scene context and fine-grained object features, achieving tighter visual-linguistic alignment and spatial reasoning capabilities. This architectural design effectively addresses the shortcomings of existing methods in complex scenes, where they struggle to integrate global semantic information with local detail information. On multiple core metrics of ScanQA (including ExactMatch, CIDEr, BLEU-4, etc.), BridgeQA achieves state-of-the-art performance, particularly demonstrating higher robustness when handling complex spatial relationships and ambiguous semantic queries, further validating the advantages of its hierarchical bridging strategy in 3D scene understanding.
Overall, LLaVA-3D and 3D-LLM lead in 3D scene reasoning, Chat-Scene and Scene-LLM perform best in object-based understanding tasks, and LEO excels in long-term task execution, reflecting the trade-offs between generalization capability, spatial alignment, and embodied interaction. Future research should further optimize multimodal fusion strategies to enhance the adaptability of 3D vision-language models in real-world 3D scenes.

\subsection{Comparison of Robot Manipulation Benchmarks}

Robot manipulation benchmarks play a crucial role in the field of robotic learning and intelligent control, providing a systematic approach to evaluating algorithms in terms of execution capability, generalization, and adaptability across different tasks. In recent years, multiple benchmarks~\cite{james2020rlbench,garcia2024towards,zheng2022vlmbench,jiang2022vima,xing2021kitchenshift,mees2022calvin,pumacay2024colosseum,hua2024gensim2} have emerged, covering a wide range of tasks from basic object grasping to complex long-horizon task planning while integrating multimodal perception, vision-language reasoning, and domain generalization. The primary differences among these benchmarks lie in the number of tasks, real-world reproducibility, the types of algorithms they support, and key evaluation metrics. Supporting these benchmarks, various simulators play a critical role in robotic manipulation research, each offering different strengths in physical accuracy, computational efficiency, task complexity support, and sim-to-real transfer capabilities.

RLBench~\cite{james2020rlbench} is built on PyRep (CoppeliaSim), a simulator that provides a diverse set of robotic manipulation tasks with high-dimensional visual inputs, making it well-suited for end-to-end policy learning. However, due to the limitations of CoppeliaSim’s physics engine, RLBench~\cite{james2020rlbench} remains restricted to purely simulated tasks, posing challenges for sim-to-real transfer. In contrast, Isaac Sim, developed by NVIDIA, is a GPU-accelerated physics simulation platform based on the PhysX engine, supporting large-scale parallel simulations and enabling direct deployment in sim-to-real tasks. It provides high-fidelity physics simulation, making it ideal for reinforcement learning (RL), imitation learning (IL), and robotic vision tasks, and is widely used in KitchenShift~\cite{xing2021kitchenshift} and GenSim2~\cite{hua2024gensim2} benchmarks. PyBullet, an open-source lightweight physics engine, is widely adopted for reinforcement learning training, offering high computational efficiency and seamless integration with OpenAI Gym. It powers benchmarks such as CALVIN~\cite{mees2022calvin} and Ravens, but its physics accuracy is lower than that of Isaac Sim and MuJoCo, particularly in complex multi-body interactions. Additionally, Ravens, developed by Google Robotics, is a robotic grasping and object manipulation simulation platform specializing in vision-based manipulation tasks. It employs Transporter Networks for efficient visuomotor policy learning and uses PyBullet as the underlying physics engine. While Ravens excels in vision-guided manipulation, its scope is relatively narrow, primarily focusing on grasping and placing tasks, with limited support for multi-step planning.

Different benchmarks choose simulators based on their research objectives. RLBench~\cite{james2020rlbench} relies on PyRep for multi-task manipulation simulation, while KitchenShift~\cite{xing2021kitchenshift} and GenSim2~\cite{hua2024gensim2} use Isaac Sim for higher physics fidelity and sim-to-real transfer. PyBullet provides high computational efficiency, making it suitable for RL-based tasks, whereas Ravens is specialized for vision-based manipulation policies. Future robotic benchmarks may further integrate high-fidelity physics simulation (e.g., Isaac Sim), high-performance computing (e.g., GPU acceleration), and multimodal learning (e.g., vision-based manipulation in Ravens) to drive robotic intelligence toward greater generalization and real-world adaptability.

RLBench~\cite{james2020rlbench} is one of the most widely used robotic manipulation benchmarks, offering 100+ tasks ranging from basic object grasping to complex tool use and assembly tasks. Its key strength lies in its task diversity and scalability, along with its extensive dataset of demonstrations, making it suitable for reinforcement learning (RL), imitation learning (IL), and traditional control methods. However, as RLBench is limited to simulation, it lacks real-world reproducibility, restricting its applicability in sim-to-real transfer. In contrast, GemBench~\cite{garcia2024towards}, built upon RLBench, focuses on task generalization, incorporating zero-shot task setups and vision-language extensions to assess robotic performance in previously unseen tasks and objects. While GemBench~\cite{garcia2024towards} presents unique advantages in multimodal fusion and generalization, it also remains constrained to simulation and lacks real-world validation.
\begin{table}[h!]
    \centering
    \renewcommand{\arraystretch}{1.2} 
    \small
    \caption{Comparison of widely used robot manipulation benchmarks in terms of simulation platform, task diversity, and real-world reproducibility.}
    \label{tab:manipulation_benchmarks}
    \begin{tabular}{p{2.5cm}p{1.4cm}p{1.1cm}p{2.2cm}}
        \hline
        \textbf{Benchmark} & \textbf{Simulator} & \textbf{\# Tasks} & \textbf{Real-World Reproducibility} \\
        \hline
        RLBench~\cite{james2020rlbench} & RLBench & 100+ & \ding{55}  \\
        \hline
        GemBench~\cite{garcia2024towards} & RLBench & 44 & \ding{55}  \\
        \hline
        VLMbench~\cite{zheng2022vlmbench} & RLBench & 8 & \ding{55} \\
        \hline
        KitchenShift~\cite{xing2021kitchenshift} & Isaac Sim & 7 & \ding{55}  \\
        \hline
        CALVIN~\cite{mees2022calvin} & PyBullet & 34 & \ding{55} \\
        \hline
        COLOSSEUM~\cite{pumacay2024colosseum} & RLBench & 20 & \checkmark \\
        \hline
        VIMA~\cite{jiang2022vima} & Ravens & 17 & \ding{55} \\
        \hline
    \end{tabular}
\end{table}

VLMbench~\cite{zheng2022vlmbench} places greater emphasis on language-conditioned robotic manipulation, similar to GemBench~\cite{garcia2024towards}. However, instead of using predefined tasks like RLBench~\cite{james2020rlbench}, VLMbench~\cite{zheng2022vlmbench} employs compositional task generation, requiring the robot to reason and generalize across multiple tasks. While VLMbench~\cite{zheng2022vlmbench} is well-suited for evaluating vision-language models (VLMs) such as CLIP and GPT-4V, its task set is relatively small (only 8 tasks), potentially limiting its comprehensiveness in benchmarking generalizable policies. In contrast, KitchenShift~\cite{xing2021kitchenshift} focuses on robot generalization under environmental domain shifts, simulating a kitchen environment where the robot interacts with various objects, such as opening an oven, moving utensils, and mixing liquids. It introduces domain shifts in object materials, lighting conditions, and camera angles to evaluate policy robustness across environmental variations. KitchenShift~\cite{xing2021kitchenshift} is particularly useful for benchmarking imitation learning (IL) and reinforcement learning (RL) strategies, but like RLBench~\cite{james2020rlbench}, it does not support real-world reproducibility, though its domain shift testing provides insight into real-world uncertainty modeling.

CALVIN~\cite{mees2022calvin}, based on PyBullet, includes 34 tasks and emphasizes long-horizon planning and multi-task adaptability. Unlike single-step manipulation tasks, CALVIN~\cite{mees2022calvin} requires robots to execute sequential subgoals, making it particularly relevant for multi-task RL and imitation learning (IL). However, it remains limited to simulation. COLOSSEUM~\cite{pumacay2024colosseum}, on the other hand, extends beyond simulated environments, supporting real-world reproducibility. It includes 20 tasks and introduces environmental perturbations, allowing robots to be tested under different physical conditions. COLOSSEUM~\cite{pumacay2024colosseum} primarily assesses robot robustness and multi-task learning performance, making it more aligned with real-world robotic applications compared to other simulation-only benchmarks, making it highly valuable for sim-to-real transfer research.

GenSim2~\cite{hua2024gensim2}, another real-world-capable benchmark, leverages MuJoCo and Isaac Sim to support variable task configurations, focusing on large-scale data generation and sim-to-real transfer. It allows researchers to dynamically adjust task complexity and test policies across different environments, making it an ideal platform for policy generalization and real-world deployment. Unlike COLOSSEUM~\cite{pumacay2024colosseum}, which focuses on environmental perturbation robustness, GenSim2~\cite{hua2024gensim2} prioritizes data scalability and adaptability, with additional support for traditional control methods such as dynamics modeling. Meanwhile, VIMA~\cite{jiang2022vima}, built on Ravens, is a multimodal manipulation benchmark designed for evaluating robot performance across multi-modal inputs (vision, language, and physics constraints). VIMA~\cite{jiang2022vima} includes 17 tasks that test a robot’s ability to understand complex task instructions and execute them accordingly. Unlike VLMbench~\cite{zheng2022vlmbench} and GemBench~\cite{garcia2024towards}, which primarily evaluate vision-language models (VLMs) in robotic control, VIMA~\cite{jiang2022vima} further integrates multi-modal task execution and precise instruction following, making it distinct in its benchmarking approach.

Overall, these benchmarks cater to different research objectives. RLBench~\cite{james2020rlbench} and CALVIN~\cite{mees2022calvin} are well-suited for large-scale multi-task RL research, while GemBench~\cite{garcia2024towards}, VLMbench~\cite{zheng2022vlmbench}, and VIMA~\cite{jiang2022vima} focus on vision-language models in robotic manipulation. KitchenShift~\cite{xing2021kitchenshift} and COLOSSEUM~\cite{pumacay2024colosseum} emphasize domain generalization and real-world adaptability, whereas GenSim2~\cite{hua2024gensim2} excels in data scalability and sim-to-real transfer. While most current benchmarks remain simulation-dependent, COLOSSEUM~\cite{pumacay2024colosseum} and GenSim2~\cite{hua2024gensim2} are gradually bridging the gap toward real-world robotic deployment, providing more valuable insights into real-world policy learning. Future research should focus on further standardizing evaluation metrics, improving sim-to-real transfer capabilities, and integrating more complex multimodal tasks to advance the development of robotic manipulation benchmarks.

\section{Challenges and Opportunities}
\label{sec:challenges_future_directions}
\begin{figure*}[ht]
    \centering
    \includegraphics[width=\linewidth]{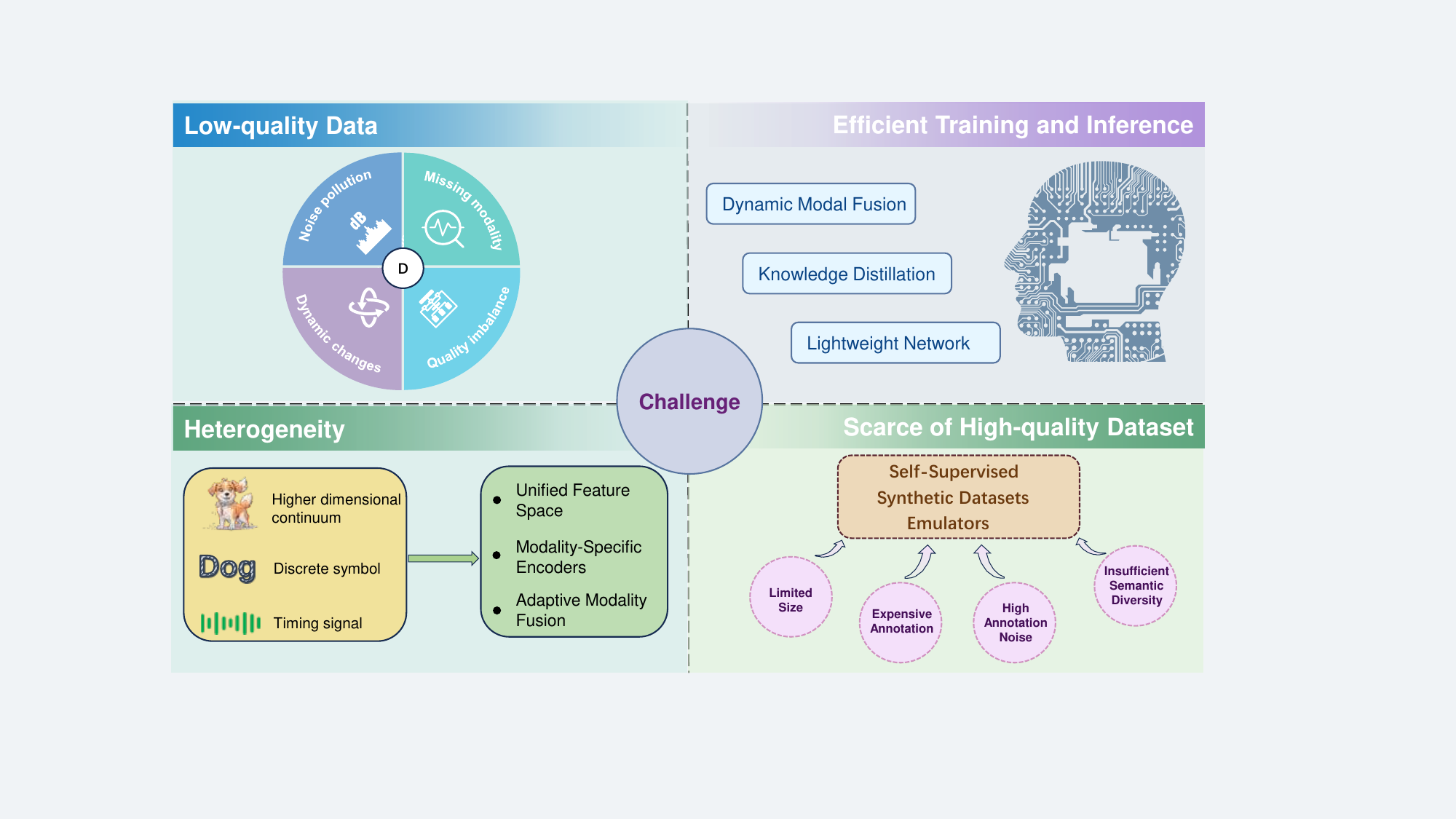}
    \caption{Major challenges faced by multimodal fusion and Vision-Language Models (VLMs). These challenges primarily encompass four aspects: (1) \textbf{Low-quality Data}, characterized by noise, missing modalities, dynamic changes, and modality quality imbalance; (2) \textbf{Efficient Training and Inference}, which involves the adoption of dynamic modality fusion strategies, knowledge distillation, and lightweight network architectures to reduce computational costs; (3) \textbf{Heterogeneity}, where significant differences in dimensions and structures across modalities necessitate unified feature spaces, modality-specific encoders, and adaptive fusion techniques; and (4) \textbf{Scarcity of high-quality datasets}, highlighting issues such as limited dataset sizes, high annotation costs, and insufficient semantic diversity, which motivate the use of self-supervised methods, synthetic datasets, and simulation-based data collection. Addressing these challenges is critical for advancing research and applications in robot vision and multimodal fusion.}
    \label{fig:Challenge}
\end{figure*}

This survey systematically reviews multimodal fusion methods and vision-language models in robotic vision systems. Despite significant progress and promising applications in multimodal fusion, several critical challenges and research gaps still remain. As illustrated in Figure \ref{fig:Challenge}, these challenges primarily include managing uncertainties caused by low-quality data, addressing the computational complexity inherent in multimodal data processing, effectively modeling and fusing cross-modal features, and dealing with dataset limitations such as small dataset sizes and insufficient semantic diversity. Successfully addressing these challenges is crucial for advancing research and applications in robotic vision and multimodal fusion.
\subsection{Low-quality Data}
The core goal of multimodal fusion is to improve the generalisation ability and robustness of models by integrating information from different modalities. However, in the real world, the presence of low-quality data seriously affects the performance of multimodal learning, making it difficult for models to work stably in complex environments. The challenges of low-quality data are mainly reflected in various aspects such as noise pollution, missing modalities, modal quality imbalance, and dynamic changes in modal quality, etc., which are widely found in many application scenarios such as autonomous driving~\cite{yoo20203d,liu2023bevfusion,liu2023sparsebev}, medical image analysis, speech recognition~\cite{oneațua2022improving}, etc., and pose great challenges to the validity and reliability of multimodal models.
Depending on the source, multimodal noise can be divided into two categories, modality-specific noise and cross-modal noise. The former refers to intra-modal noise caused by sensor errors, environmental disturbances, or data transmission losses, e.g., electronic sensor noise in visual modalities, environmental interference distortion in audio modalities. The latter, on the other hand, is high-level semantic noise introduced by weakly aligned or unaligned samples between modalities. Existing denoising methods mainly have two directions: modality-specific denoising~\cite{nie2021total,yang2021variational} and cross-modality denoising~\cite{changpinyo2021conceptual,radenovic2023filtering,gadre2023datacomp}. In terms of modality-specific denoising, common approaches include joint variation-based fusion denoising strategies, e.g., in multispectral image fusion tasks, noise is removed by jointly optimising different modal data through a variational model~\cite{yang2021variational}. As for cross-modal denoising, the main approaches are model-based rectifying and noise robustness regularization. Future research directions can further explore how to use the correlation between multimodal data to identify and reduce noise. For example, in hyperspectral images, noise patterns in similar bands are often correlated and can be used for cross-channel noise reduction. In addition, how to effectively deal with high-level noise at the semantic level, e.g., how to enhance cross-modal alignment and denoising capabilities in multimodal large models (MLLMs), is also a direction that deserves to be explored in depth.

Incomplete multimodal learning presents challenges in fusion tasks, as missing modalities due to device issues, data loss, or user preference affect performance. For instance, in medical diagnosis, key modalities like PET scans may be unavailable. Solutions fall into imputation-based~\cite{lin2021completer} and imputation-free methods~\cite{lian2023gcnet}. Imputation-based methods recover missing data via matrix decomposition, graph learning, or GANs, while imputation-free approaches leverage available modalities through latent representation learning and deep learning. Challenges remain in assessment, uneven modality information, and incomplete annotations. Modality imbalance also affects performance, including property and quality discrepancies, causing models to over-rely on certain modalities. To mitigate this, optimization techniques, gradient modulation, and adaptive architectures~\cite{ma2023calibrating,fan2023pmr} help balance training. Additionally, knowledge distillation~\cite{zhang2025dfmc}, contrastive learning, and dynamic data augmentation enhance lower-quality modalities, improving integration and model robustness.

\subsection{Heterogeneity}
Heterogeneity is a central challenge in multimodal fusion tasks, mainly due to the inherent differences in data structure, feature distribution and information representation. For example, image data are usually high-dimensional continuous values, audio data are time-series signals, and text data are discrete symbol sequences. This heterogeneity makes it difficult to directly fuse multimodal information, making uniform representation learning, feature alignment, and information interaction key issues.
To address the heterogeneity challenge, researchers have proposed three main types of approaches:(1) Unified Feature Space Learning, which projects data from all modalities into the same feature space by constructing a common representation space, e.g., CLIP~\cite{radford2021learning} employs Visual-Linguistic Contrast learning so that images and text share the same latent space. (2) Modality-Specific Encoders, which designs special encoders for different modalities, using Transformer for text, CNN for image, RNN for audio, etc., and then performs feature alignment through Cross-modal Attention~\cite{lu2019vilbert,chen2020uniter}. (3) Adaptive Modality Fusion, which dynamically adjusts the fusion method according to the data quality, information contribution and task requirements of different modalities. For example, in autonomous driving, the fusion weights of RGB images and LiDAR point clouds can be adaptively adjusted according to the ambient lighting conditions~\cite{huang2020epnet}.Although these methods alleviate the heterogeneity problem to some extent, there are still many challenges. For example, the distribution of different modalities in high-dimensional space may be skewed, leading to information loss or feature mismatch during direct fusion. In addition, how to effectively model the information complementarity between different modalities is still an open question. Therefore, future research can further explore methods based on graph neural networks and self-supervised learning to more effectively model the complex relationships between multimodalities and improve the generalisation ability of fusion.

\subsection{Efficient Training and Inference}
Efficient training and inference is an important challenge in the practical deployment of multimodal large models of visual language. Current mainstream VLMs \cite{dubey2024llama,bai2023qwen,driess2023palm,lin2024moe} are usually pre-trained based on large-scale Transformer architectures, which usually have billions or even tens of billions of parameters, and pre-training on large-scale datasets requires a large amount of computational resources and time costs, which limits the training efficiency and scale-up ability of the models, and makes the model deployment face serious computational bottlenecks in real-world applications. This computational cost not only limits the practical application scope of the model, but also increases energy consumption and cost, especially in edge devices or real-time application scenarios (e.g., autonomous driving). In addition, due to the obvious heterogeneity of visual and linguistic modalities in terms of data dimensions and feature space, the models may face low training efficiency, high inference latency, and high memory consumption during joint optimisation. Therefore, how to design a more lightweight model structure, reduce the computational overhead of cross-modal fusion, and effectively improve the training and inference efficiency are important challenges in current VLMs research.
In order to alleviate the above problems, current research mainly proposes solutions from lightweight network architecture design, model compression and knowledge distillation, and dynamic modal fusion. On the one hand, the computational complexity is effectively reduced by introducing lightweight visual language fusion architectures, such as CrossViT~\cite{chen2021crossvit}, MobileViT~\cite{mehta2021mobilevit},MoCoViT~\cite{ma2022mocovit}and other multi-scale Transformer architectures; on the other hand, the knowledge from the large-scale teacher model is migrated to the lightweight student model through knowledge distillation~\cite{guo2025deepseek}, which significantly improves the training efficiency while maintaining the performance of the while significantly improving the training efficiency. In addition, Modality Dropout or Adaptive Modality Selection methods can reduce redundant modal computations. Therefore, how to effectively combine structural design, optimisation strategies and modal fusion mechanisms to achieve efficient training and deployment of large-scale VLMs is an important research topic that needs to be broken through in the field of VLMs.
Despite the remarkable progress of VLMs in cross-modal understanding and reasoning, their application in robotic vision still faces several core challenges. One primary obstacle lies in their limited capability for real-time deployment. Leading VLMs such as CLIP~\cite{radford2021learning}, BLIP~\cite{li2023blip}, and GPT-4V~\cite{yang2023dawn} are generally constructed with large-scale Transformer architectures, featuring dense self-attention operations and extended token contexts. These result in high computational overhead and inference latency, making such models impractical for embedded systems or edge devices. Moreover, many of these models operate with frozen pretrained parameters, restricting their flexibility in adapting to online tasks.Another critical challenge is the insufficient domain adaptability of current VLMs. Since most models are trained on static, large-scale image-text corpora such as LAION-400M and Conceptual Captions, their alignment mechanisms often struggle to generalize to robotics-specific sensory inputs-such as sparse or low-light scenes, 3D geometry, and embodied interactions. This limitation becomes more pronounced under distribution shifts, including changes in lighting, task formats, or robot embodiment across platforms~\cite{shridhar2023perceiver}.Promising directions to address them include dynamic model pruning, adapter-based fine-tuning for efficient on-device learning, and reinforcement learning-based grounding mechanisms that enable real-time feedback and task adaptation.

\subsection{Scarce of High-quality Dataset}
The performance and generalisation capabilities of VLMs are highly dependent on the availability, diversity and quality of training datasets. However, existing large-scale visual language datasets (e.g., Visual Genome~\cite{krishna2017visual}, Conceptual Captions~\cite{sharma2018conceptual}, and LAION-400m~\cite{schuhmann2021laion}) generally suffer from limited size, high annotation noise, insufficient semantic diversity, and expensive annotation. These problems not only limit the models from learning rich and accurate cross-modal semantic alignment relations, but also easily lead to overfitting the models to specific contexts or tasks, making it difficult to effectively generalise them to wider and more realistic application scenarios. In addition, the reliability of the trained VLMs in real-world applications is significantly affected due to the annotation biases (e.g., semantic homogeneity, cultural bias, or linguistic representation bias) present in the dataset.

To alleviate this challenge, current research has explored three main avenues: first, to reduce the model's dependence on annotation quality through self-supervised or weakly supervised learning strategies, e.g., models such as CLIP~\cite{radford2021learning}, ALIGN~\cite{jia2021scaling}, etc., which use large-scale weakly annotated or natural language supervised learning of generalised visual language representations; second, to enhance the generalisation capability of the model by generating synthetic datasets models' generalisation capabilities, e.g., using rules or third-party tools (e.g., Stable Diffusion, DALL-E) to generate image-text pairs with diversity and semantic richness; and thirdly, constructing datasets closer to the physical world in specific domains (e.g., robotic manipulation, automated driving), using emulators or world models to interactively generate data to enhance the performance of VLM in the real physical world. Therefore, future research should further explore how to effectively fuse real and synthetic data, how to assess the diversity and quality of datasets, and how to minimise annotation costs while constructing more efficient and realistic VLM datasets to enhance the performance and reliability of VLMs in real-world environments.

In addition to limited scale and modality diversity, current multimodal datasets often exhibit cultural and environmental biases that constrain the generalization ability of VLMs in real-world robotic scenarios. Many widely used datasets are built upon Western-centric indoor layouts (e.g., ScanNet~\cite{dai2017scannet}, Matterport3D~\cite{chang2017matterport3d}), contain only English-language instructions (e.g., Room-to-Room~\cite{zhao2022real} for vision-language navigation), or are collected in urban driving environments (e.g., nuScenes~\cite{caesar2020nuscenes}, Waymo~\cite{sun2020scalability}). As a result, VLMs trained on such data may show reduced robustness when deployed in multilingual, cross-cultural, or rural environments, where object appearances, spatial structures, and human behaviors can differ significantly from those in the training data.

Environmental bias also poses structural challenges to model performance. Most vision-language corpora are collected under favorable conditions-clear weather, adequate lighting, and static scenes-lacking coverage of complex situations such as nighttime, adverse weather, or highly dynamic environments. This data collection bias can lead to severe performance degradation when models are exposed to fog, rain, occlusions, or high-motion scenarios in practical robotic tasks, thereby limiting their robustness and generalization capabilities.Such limitations in generalizability are particularly critical in robotic vision applications, especially when long-term deployment, task transfer, or operation in novel environments is required (e.g., navigation and manipulation). VLMs trained solely on static and homogeneous data sources are thus insufficient for the diverse needs of embodied agents operating in open-world settings.

To mitigate these issues, future research should place greater emphasis on diversity, coverage, and fairness during dataset construction. Promising directions include the development of multimodal datasets that encompass multiple cultures, languages, and environmental variables; the integration of cross-domain training and evaluation protocols to enhance model robustness under distribution shifts; and the adoption of domain adaptation, semantic augmentation, and sim-to-real transfer techniques to improve generalization in complex real-world deployments.

\subsection{Perspectives and future directions}

\textbf{Spatial intelligence:}
As robotic vision systems gradually evolve from static recognition and perception toward active exploration and embodied interaction, spatial intelligence has emerged as a key direction in the next phase of multimodal large model development~\cite{jeong2024survey}. Spatial intelligence requires models not only to possess geometric perception capabilities for 3D structures but also to perform memory, reasoning, and prediction over time-particularly maintaining persistent representations of spatial semantics, interaction history, and future actions in dynamic environments~\cite{yang2025thinking}.Currently, multimodal large models still exhibit limitations in spatial understanding. For example, they often lack the ability to explicitly model the spatial topology of real-world environments, which hinders their capability to perform high-quality causal reasoning across space-language-action domains. To address this, future research should introduce explicit spatial memory mechanisms and structured spatial representations, such as 3D graph structures and topological map embeddings, to better support long-term task context modeling and navigation history encoding. In addition, the spatial projection, path generation, and environmental adaptability of VLMs in 3D environments should be enhanced by incorporating components such as SLAM, egocentric modeling, and visual loop closure, thus forming a spatial cognition closed-loop system tailored for embodied agents.
Importantly, spatial intelligence also demands that models possess cross-modal and cross-scale generalization abilities, enabling them to handle a wide range of tasks from micro-level manipulation (e.g., object grasping) to macro-level planning (e.g., indoor pathfinding). Therefore, developing a unified framework that integrates spatial semantic mapping, action sequence prediction, and language-based interaction will be an essential direction for future research in embodied multimodal models. To enhance real-world deployment, further exploration is needed into sparse spatial representations, memory compression, and symbolic abstraction, which can improve task alignment and rapid adaptation in large-scale, unfamiliar environments.

\textbf{Ethical deployment:}
With the growing integration of vision-language models (VLMs) into robotic systems, concerns regarding model safety, robustness, and ethical deployment are becoming increasingly prominent. Recent studies have demonstrated that VLMs are susceptible to image perturbations and ambiguous linguistic inputs, potentially leading to incorrect interpretations and unsafe actions in real-world environments~\cite{chen2024cello}. To address these issues, future research should prioritize the development of standardized adversarial evaluation benchmarks that assess the robustness of VLMs under multimodal interference, complex semantics, and domain shifts. In addition, the incorporation of causal reasoning mechanisms and multimodal consistency regularization can significantly improve the explainability and accountability of robotic decision-making. For instance, causal chain-of-thought modeling, supported by datasets such as CELLO~\cite{chen2024cello}, may enable robots to generate interpretable and traceable justifications for their actions across different stages of task execution. These approaches can also be extended by designing intervention-aware reasoning modules that allow systems to anticipate and clarify potential failures before action is taken, thereby enhancing overall system transparency and reliability. Moreover, the ethical deployment of VLM-based robotic systems in safety-critical or cross-cultural contexts requires careful attention to human oversight and normative alignment. Recent advances have highlighted the potential of human-in-the-loop trust modeling~\cite{kumar2024applications} and ethical instruction tuning~\cite{kim2024openvla} to improve the controllability and societal acceptability of robot behavior. In this regard, future systems should include integrated ethical supervision modules capable of recording both attention distributions and decision trajectories, presented through interpretable visual-log interfaces to support auditing and post-hoc analysis. Furthermore, incorporating region-specific ethical guidelines-such as those defined by the IEEE Global Initiative on Ethics of Autonomous and Intelligent Systems or the European Union's AI Act-into model training and deployment workflows may enhance robots' cultural adaptability and regulatory compliance. Together, these strategies offer a promising direction toward building vision-language-enabled robotic systems that are not only capable and intelligent, but also trustworthy, transparent, and ethically aligned.

\textbf{Embodied Cognition and Brain-Inspired Design:}
Another important direction for the future development of VLM-based robotic systems lies in fostering interdisciplinary collaboration with fields such as cognitive science and neuroscience. Although current VLMs have achieved impressive performance in grounding visual and linguistic information, they still lack the embodied reasoning, memory consolidation, and value-driven decision-making observed in biological agents. Drawing inspiration from studies on the human brain's multimodal integration-particularly the role of the prefrontal cortex and hippocampus in semantic association and spatiotemporal memory-may provide valuable insights for building more generalizable and adaptive robotic systems. For example, recent advances in computational neuroscience have introduced biologically inspired attention and memory mechanisms that could inform the design of more interpretable and causally coherent vision-language architectures~\cite{liu2024cognitive}. Moreover, aligning robot learning objectives with models of human goal inference and reward prediction may help VLMs better handle ambiguous, underspecified, or ethically complex scenarios in real-world interactions. Through closer collaboration between robotics, machine learning, and neuroscience, future VLM-augmented robots may move beyond purely data-driven pattern recognition to achieve more grounded, cognitively plausible, and socially aware intelligence.

\section{Conclusions}

\label{sec:conclusions}
This paper systematically explores the critical role of VLMs in robot vision. Focusing on typical tasks such as semantic understanding, 3D Object detection, embodied navigation, and robotic manipulation, we summarize how different fusion paradigms (e.g., encoder-decoder frameworks, attention mechanisms, graph neural networks, etc.) integrate across modalities such as vision and language, depth, and point clouds. We also conduct a detailed comparison of traditional methods and the evolutionary paths and applicability of recent large-model-based approaches. Additionally, we systematically reviewed relevant datasets and evaluation benchmarks, revealing current imbalances in research across methodological, data, and task dimensions. We have identified the following three key research findings: First, cross-modal alignment strategies are critical to determining the upper limit of a robot's perception system performance, which is currently constrained by modal differences and mismatches in semantic granularity; Second, lightweight, multi-stage adaptation mechanisms are a practical necessity for deploying large-scale vision-language models on robot platforms, especially in scenarios with limited edge computing resources; Third, task-oriented and online adaptation fusion strategies are increasingly becoming a research focus, aiming to drive vision-language models to dynamically adjust their perception and decision-making paths through scene cognition, thereby supporting robot response capabilities under high-frequency, highly variable target conditions. Although current multimodal and VLM methods perform well on standardized benchmark tasks, they still face numerous challenges on actual robot platforms, including modal incompleteness caused by sensor heterogeneity, semantic bias caused by sparse or low-quality perception inputs, and requirements for latency, robustness, and interpretability during task execution. Future research should focus on introducing structured spatial modeling and memory mechanisms to enhance spatial reasoning, strengthening system interpretability and ethical adaptability, and incorporating brain-inspired modeling principles to develop cognitive VLM architectures with long-term learning capabilities, thereby driving the development of more autonomous, efficient, and explainable intelligent robotic systems.

\section*{Acknowledgement}

This work was supported by National Natural Science Foundation of China (Nos. 62376271, U21A20515, U22B2034, 62572059, 62172416, 62365014, 62271074 and 62262043),   Beijing Natural Science Foundation (JQ23014, L231013) and the Open Project Program of State Key Laboratory of Virtual Reality Technology and Systems, Beihang University (No. VRLAB2025B03).

\bibliography{IF}
\end{sloppypar}
\end{document}